%% file: main.tex
\def\NOTODO{}
\documentclass[10pt,twocolumn,letterpaper]{article}

\usepackage[pagenumbers]{cvpr}      %

\input{preamble}

\definecolor{cvprblue}{rgb}{0.21,0.49,0.74}
\usepackage[pagebackref,breaklinks,colorlinks,allcolors=cvprblue]{hyperref}

\newcommand{\danilo}[1]{\textcolor{teal}{{\bf [Danilo: }{\em #1}{\bf ]}}}
\ifdefined\NOTODO\renewcommand{\danilo}[1]{}\fi%

\def\confName{CVPR}
\def\confYear{2027}

\title{A Search Worth a Billion Pixels: Explaining Gigapixel Classifiers}

\author{Danilo Danese \qquad Angela Lombardi \qquad Tommaso Di Noia\\
Politecnico di Bari, Italy\\
{\tt\small \{name.surname\}@poliba.it}
}

\begin{document}
\maketitle
\input{sec/0_abstract}
\input{sec/1_intro}

\input{sec/2_related}

\input{sec/3_method}

\input{sec/4_setup}
\input{sec/5_results}
\input{sec/6_conclusion}

{
    \small
    \bibliographystyle{ieeenat_fullname}
    \bibliography{main}
}

\input{supplementary_list}

\clearpage
\end{document}

%% file: preamble.tex
\newcommand{\TODO}[1]{\textbf{\color{red}[TODO: #1]}}
\ifdefined\NOTODO\renewcommand{\TODO}[1]{}\fi

\newcommand{\nbold}[1]{\par\noindent\textbf{#1}\hspace{1em}\ignorespaces}

\usepackage{xspace}

\newcommand{\method}{WaFR\xspace}

\input{tables/style}

\newcommand{\nrm}{{\color{tabmute}n.r.$^{m}$}}%
\newcommand{\nrt}{{\color{tabmute}n.r.$^{t}$}}%
\newcommand{\nrd}{{\color{tabmute}n.r.$^{d}$}}%
\newcommand{\sa}{$^{\color{tabmute}a}$}%

\usepackage{amsthm}
\newtheorem{proposition}{Proposition}

\usepackage{algorithm}
\usepackage{algpseudocode}

\algrenewcommand\algorithmicrequire{\textbf{Require:}}
\algrenewcommand\algorithmicensure{\textbf{Return:}}

%% file: tables/style.tex
\usepackage{booktabs}
\usepackage{colortbl}
\usepackage{tikz}
\definecolor{tabmute}{HTML}{6B6F7A}   %
\definecolor{tabrule}{HTML}{BFC3CC}   %
\definecolor{tabbar}{HTML}{5B8DB8}    %
\definecolor{tabtree}{HTML}{AEB3BF}

\newif\iftabfirst
\newcommand{\tabstyle}{\arrayrulecolor{tabrule}\setlength{\heavyrulewidth}{.5pt}\setlength{\lightrulewidth}{.35pt}%
  \setlength{\cmidrulewidth}{.35pt}\renewcommand{\arraystretch}{1.08}\global\tabfirsttrue}
\newcommand{\tabheadfont}{\color{tabmute}\sffamily\fontsize{7.3}{8.6}\selectfont}
\newcommand{\tabhead}[1]{{\tabheadfont #1}}
\newcommand{\tabgroup}[2]{\iftabfirst\noalign{\global\tabfirstfalse}\else\midrule\fi\multicolumn{#1}{@{}l}{\tabheadfont #2}\\[-2pt]}
\newcommand{\ci}[1]{\kern1pt{\color{tabmute}\fontsize{5.6}{6}\selectfont\raisebox{-.25ex}{$\pm$#1}}}
\newcommand{\best}[1]{\textbf{#1}}
\newcommand{\ours}[1]{#1}%
\newcommand{\second}[1]{\underline{#1}}
\newcommand{\ctrl}[1]{{\color{tabmute}#1}}

\newcommand{\na}{{\color{tabmute}--}}
\newlength{\tbarlen}

\newlength{\tabtreeup}\newlength{\tabtreedown}
\makeatletter
\newcommand{\tabtree}[1]{\leavevmode\setlength{\tabtreeup}{\ht\@arstrutbox}\setlength{\tabtreedown}{\dp\@arstrutbox}%
  \rlap{\tikz[overlay]{\draw[tabtree,line width=.45pt] (2.5pt,\the\tabtreeup) -- (2.5pt,\ifnum#1=1 -\the\tabtreedown\else .33em\fi);
  \draw[tabtree,line width=.45pt] (2.5pt,.33em) -- (6.5pt,.33em);}}\hspace*{9pt}}
\makeatother
\newcommand{\treemid}{\tabtree{1}}
\newcommand{\treelast}{\tabtree{0}}
\newcommand{\dci}[2]{\kern4.6pt{\color{tabmute}\fontsize{5.6}{6}\selectfont\raisebox{-.25ex}{[#1,\,#2]}}}
\newcommand{\hurt}{\rlap{$^{\circ}$}}%
\newcommand{\ablid}[1]{{\color{tabmute}#1}\enspace}%

%% file: sec/0_abstract.tex
\begin{abstract}
Perturbation-based attribution methods explain model predictions without accessing their internal representations, but scaling them to high-resolution images requires balancing perturbation granularity and computational cost. Fine-grained perturbations require many model evaluations, whereas coarse masks may overlook small but relevant evidence. We introduce Wavelet Find-and-Refine (\method), a model-agnostic attribution method combining adaptive hierarchical search with sparse multiscale regression. By selectively refining regions that influence the prediction, \method maintains fine-grained localisation with a computational cost that depends on the spatial distribution of the evidence. As image resolution increases to 8,192 pixels per side, \method maintains a Pointing Game score of at least .995 in controlled localisation experiments, including at matched forward-pass budgets. On gigapixel whole-slide images, it requires a median of 176 forward passes and achieves the highest mean patch-level AUROC among the evaluated perturbation-based methods (.949).
\end{abstract}

%% file: sec/1_intro.tex
\section{Introduction}
\label{sec:intro}
Deep neural networks support decisions in medicine~\cite{esteva2017dermatologist,litjens2017survey,campanella2019clinical}, manufacturing~\cite{bergmann2019mvtec} and science~\cite{jumper2021alphafold}. A pathologist interpreting a model's tumour score, or an engineer examining a defect score, needs to understand which regions of the image contribute to the prediction~\cite{teneggi2022hshap}. Attribution methods address this problem by assigning importance to regions of the input~\cite{simonyan2014deep,zeiler2014visualizing,petsiuk2018rise}. Unlike gradient- and activation-based approaches~\cite{sundararajan2017axiomatic,selvaraju2017gradcam}, perturbation-based methods estimate regional importance from changes in model predictions under masking, without accessing internal representations~\cite{lundberg2017unified,novello2022hsic,ribeiro2016lime}.

As image resolution increases, the spatial granularity of perturbations becomes critical. Methods based on fixed-resolution masks progressively aggregate larger image regions, potentially overlooking small but discriminative evidence. Conversely, approaches that preserve a fixed spatial granularity by evaluating individual regions require an increasing number of forward passes as the image grows. This trade-off is particularly relevant for industrial inspection and digital pathology, where discriminative structures may occupy only a small fraction of images containing millions or billions of pixels~\cite{campanella2019clinical}. Explaining predictions at these scales therefore requires controlling computational cost without sacrificing the spatial resolution needed to identify relevant evidence.

Hierarchical perturbation methods organise regions into progressively finer partitions, concentrating forward passes where coarse perturbations indicate relevant evidence~\cite{cooper2022hipe,rashid2026shapbpt,teneggi2022hshap}. Their cost depends on how many regions require refinement. Haar wavelets provide a complementary, spatially localised multiscale representation~\cite{mallat1989theory} that can describe structured signals sparsely~\cite{kolek2022cartoonx}, without determining which image regions should be refined.

We introduce Wavelet Find-and-Refine (\method), a model-agnostic attribution method combining adaptive hierarchical search with sparse multiscale regression. At the reference input resolution, regression estimates regional contributions from random perturbations. On larger images, prediction changes guide selective refinement over a dyadic partition, while regression ranks regions left unresolved by the search. A Haar low-pass baseline preserves coarse spatial content during perturbation. When few regions are expanded per level, the search cost grows with the hierarchy's depth rather than with an exhaustive spatial partition.

Our contributions are threefold: 

\noindent(i) a hierarchical perturbation strategy with a cost bound conditional on the number of expanded regions; (ii) its integration with sparse multiscale regression to rank regions left unresolved by the search; and (iii) an evaluation of localisation, fidelity and computational cost across image scales, including matched forward-pass comparisons.

\begin{figure}[t]
  \centering
  \includegraphics[width=\linewidth]{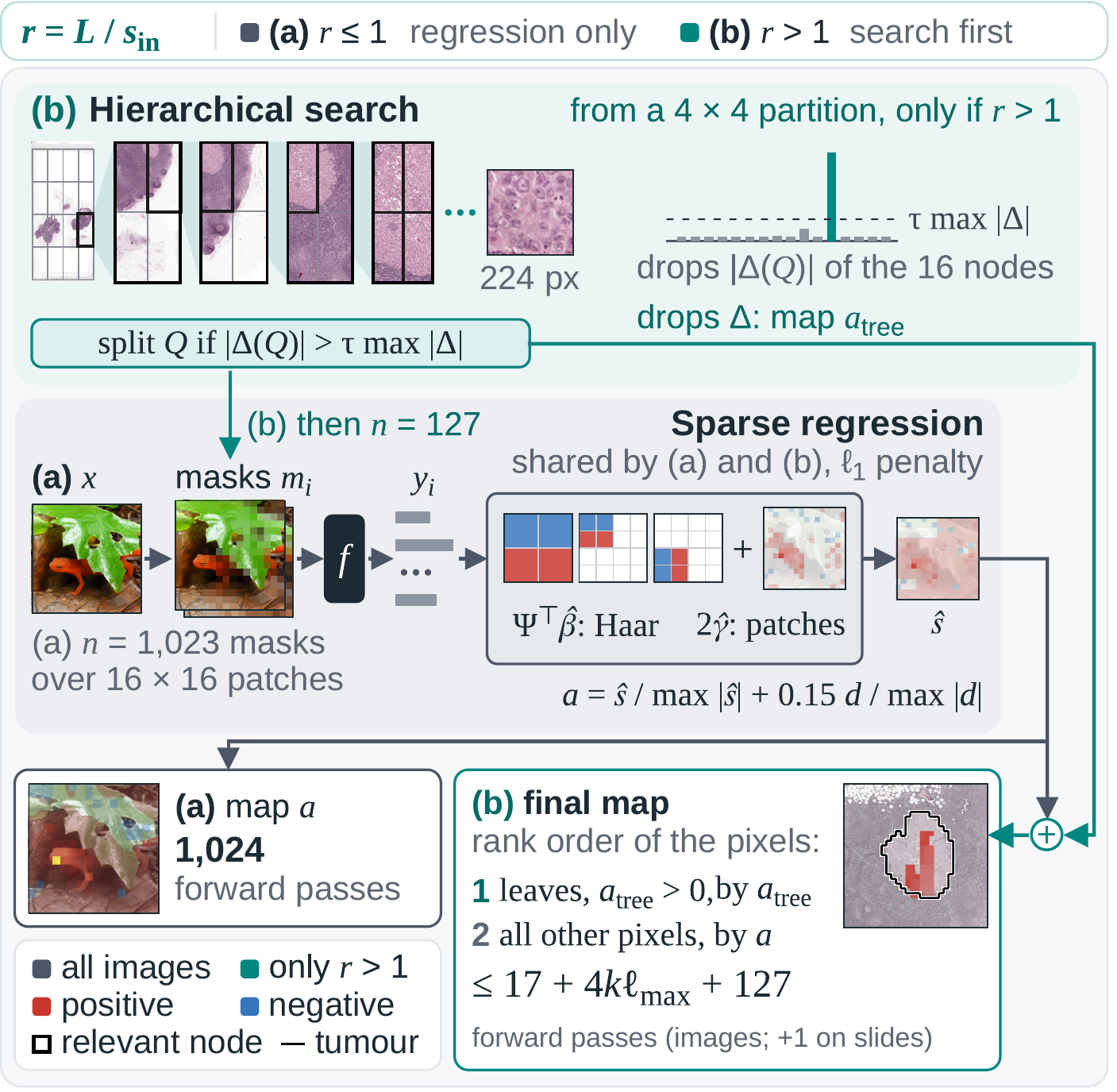}
\caption{\textbf{Overview of \method.} (a)~$r\leq1$: sparse regression on random masks over $16\times16$ patches. (b)~$r>1$: hierarchical search, then the same regression on 127 masks; the final map ranks leaves with $a_{\mathrm{tree}}>0$ first. Red: positive; blue: negative; thick frame: relevant node; outline: tumour.}
\label{fig:teaser}
\end{figure}

%% file: sec/2_related.tex
\section{Related work}
\label{sec:related}

\nbold{Gradient- and activation-based methods.}
These methods require access to the internal state of the model, so they are model-specific. Input gradients provide local attribution~\cite{simonyan2014deep}, extended through path integration in Integrated Gradients~\cite{sundararajan2017axiomatic} or noise averaging in SmoothGrad~\cite{smilkov2017smoothgrad}. Activation-based methods use intermediate feature maps, weighted by gradients in Grad-CAM~\cite{selvaraju2017gradcam} and Grad-CAM++~\cite{chattopadhay2018gradcampp}, or by prediction scores in Score-CAM~\cite{wang2020scorecam}. Even Score-CAM, which uses forward passes, requires internal activations; these maps also inherit the resolution of the chosen feature layer.

\nbold{Perturbation-based methods.}
These methods only probe the model with perturbed inputs, so they are model-agnostic. Occlusion~\cite{zeiler2014visualizing} estimates regional importance by masking one region at a time, while RISE~\cite{petsiuk2018rise} perturbs multiple regions simultaneously and averages their masks weighted by the prediction scores. LIME~\cite{ribeiro2016lime} and KernelSHAP~\cite{lundberg2017unified} fit additive surrogate models to the responses of perturbed inputs, whereas Sobol~\cite{fel2021sobol} and HSIC~\cite{novello2022hsic} estimate the influence of mask variables through variance decomposition and kernel dependence, respectively. Other approaches, including LIMA~\cite{chen2025lima}, MoXI~\cite{sumiyasu2024moxi} and PhaseWin~\cite{gu2026phasewin}, select informative subsets of image regions according to their effect on the prediction.

\nbold{Hierarchical methods.}
HiPe~\cite{cooper2022hipe} refines regions according to their attribution scores. h-Shap~\cite{teneggi2022hshap} recursively partitions images into quadrants and computes Shapley values using 16 forward passes per expanded node, with a cost bound under a multiple-instance assumption. In contrast, \method uses regional score differences to guide refinement and regression to rank unresolved regions. PartitionSHAP~\cite{lundberg2020partition} and ShapBPT~\cite{rashid2026shapbpt} estimate Owen values over hierarchical binary partitions, with costs depending on the chosen hierarchy and expanded nodes.

\nbold{Wavelets and explanations.}
CartoonX~\cite{kolek2022cartoonx}, ShearletX~\cite{kolek2023shearletx} and WAM~\cite{kasmi2025wam} optimise attributions in transformed image representations using gradients. WCAM~\cite{kasmi2023wcam} instead perturbs groups of wavelet coefficients and estimates Sobol indices~\cite{fel2021sobol}. Unlike adaptive spatial search, these approaches operate on a chosen multiscale representation rather than selecting regions for recursive refinement.

\nbold{Whole-slide images.}
WSI classifiers typically aggregate patch features through multiple-instance learning~\cite{ilse2018attention,li2021dsmil}. Explanations may use attention or internal relevance propagation, as in xMIL~\cite{hense2024xmil}; transformer features can also exhibit high-norm background tokens that distort attention maps~\cite{chen2022hipt,darcet2024registers,dosovitskiy2021image}. HiPe~\cite{cooper2022hipe} illustrates hierarchical perturbation on a whole slide.

%% file: sec/3_method.tex
\section{Method}
\label{sec:method}

\method operates on model predictions under spatial perturbations. Let $r=L/s_{\mathrm{in}}$ denote the ratio between the image side and the reference input side of the model. At $r\leq1$ it estimates patch attribution by sparse regression (\cref{sec:refine}); for $r>1$ it first performs adaptive hierarchical search and then ranks unrefined regions through regression (\cref{sec:find}).

\subsection{Problem setting}
\label{sec:notation}

Let $f$ be a classifier and $x\in\mathbb{R}^{L\times L\times C}$ an image of side $L$ and $C$ channels. We consider a target class and denote its logit by $f(x)\in\mathbb{R}$. We count the cost of an explanation in forward passes.

The contribution of a region $Q\subseteq\Omega$, where $\Omega$ denotes the image domain, is estimated by replacing its pixels with a reference image $b(x)$ and measuring the resulting change in prediction. Since the reference determines which information is removed by the perturbation, we use a Haar low-pass baseline that retains coarse spatial content. Specifically, $b(x)$ replaces each pixel with the mean of its cell in a $16\times16$ grid of square blocks. When $L/16$ is a power of two, this corresponds to the Haar low-pass approximation of $x$, so replacing a region made of whole squares with $b(x)$ removes its Haar detail coefficients finer than the squares while keeping their means. With $\mathbf{1}_Q$, the indicator, denoting the binary map of $Q$ and $\odot$ the Hadamard product, the perturbed image and its associated score difference are
\begin{equation}
\small
\begin{aligned}
x_{\setminus Q}
    &= \mathbf{1}_{\Omega\setminus Q}\odot x
       + \mathbf{1}_Q\odot b(x),\\
\Delta(Q)
    &= f(x)-f(x_{\setminus Q}).
\end{aligned}
\label{eq:drop}
\end{equation}
A positive $\Delta(Q)$ indicates that removing $Q$ reduces the target-class score.

Occlusion requires $1+(L/w)^2$ forward passes for square patches of side $w$, regardless of where relevant evidence occurs. Independent perturbations also fail to capture some interactions between regions.

\subsection{Sparse regression on random masks}
\label{sec:refine}

Rather than perturbing regions independently, \method estimates their contributions from masks that retain or remove multiple regions simultaneously, as in RISE~\cite{petsiuk2018rise}. The image is divided into $16\times16$ patches, yielding 256 regional effects independently of image resolution. This partition also supports four levels of a discrete Haar representation.

We draw $n$ binary masks $m_i\in\{0,1\}^{256}$, where $m_{i,j}=1$ retains patch $j$ and $m_{i,j}=0$ replaces it with the corresponding patch of $b(x)$. Each mask is sampled from a $7\times7$ matrix of independent Bernoulli $(0.5)$ variables, bilinearly interpolated with a random spatial shift and thresholded at $0.5$ after averaging over each patch. This construction introduces spatial correlation between masked regions, while each patch remains visible in approximately half of the perturbations. Let $y_i$ denote the target-class logit of the image produced by mask $m_i$.

Following additive surrogate approaches~\cite{lundberg2017unified}, we approximate the prediction score as
\begin{equation}
\small
y_i \approx \beta_0+\langle m_i,s\rangle,
\end{equation}
where $\beta_0\in\mathbb{R}$ is an intercept and $s\in\mathbb{R}^{256}$ contains the estimated patch effects. We regularise this estimation using complementary sparse representations: individual patches for localised evidence and Haar coefficients for spatially coherent effects. Let $\Psi\in\mathbb{R}^{256\times256}$ contain the orthonormal Haar basis of the $16\times16$ patch matrix, including its constant component. At each scale $j\in\{1,\ldots,4\}$, three wavelets describe horizontal, vertical and diagonal contrasts within squares of $2^j\times2^j$ patches, giving 255 wavelets in addition to the constant component. We express the patch effects as
\begin{equation}
\small
s=2\gamma+\Psi^\top\beta,
\label{eq:split}
\end{equation}
where $\gamma,\beta\in\mathbb{R}^{256}$ represent patch-wise and multiscale contributions, respectively.
The coefficients are estimated through an $\ell_1$-penalised least-squares objective,
\begin{equation}
\small
\begin{aligned}
(\hat{\beta}_0,\hat{\gamma},\hat{\beta})
=\arg\min_{\beta_0,\gamma,\beta}\;
&\frac{1}{2n}\sum_{i=1}^{n}
\Big(
y_i-\beta_0
-\langle 2m_i,\gamma\rangle\\
&
-\langle\Psi m_i,\beta\rangle
\Big)^2+\lambda\left(\|\gamma\|_1+\|\beta\|_1\right).
\end{aligned}
\label{eq:lasso}
\end{equation}
We select $\lambda$ per image by five-fold cross-validation over 30 log spaced values down to $10^{-3}$ of the all-zero threshold.

The factor of two in \cref{eq:split} controls the relative penalty assigned to the two representations. Under the common $\ell_1$ penalty, constant regions of up to $4\times4$ patches are less costly to represent through $\gamma$, whereas larger coherent regions can be represented more economically through $\beta$. Since both components describe the same patch effects, their decomposition is determined by the regularisation rather than being uniquely identified by the observations. The resulting estimate,
\begin{equation}
\small
\hat{s}=2\hat{\gamma}+\Psi^\top\hat{\beta},
\end{equation}
provides a sparse attribution map at patch resolution.

Sparsity, however, can assign identical values, often zero, to many patches, preventing a complete ranking of the image. To distinguish these patches, we complement $\hat{s}$ with a dense estimate $d$ derived from the same random masks. For each patch $j$, $d_j$ is the difference between the mean target-class softmax probability over masks retaining that patch and the mean probability over masks removing it. After upsampling both estimates to image resolution and smoothing $d$ with a Gaussian of standard deviation equal to half a patch, the attribution map is
\begin{equation}
\small
a=
\frac{\hat{s}}{\max|\hat{s}|}
+0.15\,\frac{d}{\max|d|}.
\label{eq:map}
\end{equation}
The weight $0.15$ is selected on the reference-resolution setting. At $r\leq1$, \method uses $n=1{,}023$ masks plus one forward pass of the original image (1,024 forward passes).

\subsection{Hierarchical search}
\label{sec:find}
\label{sec:rule}

For $r>1$, each patch of the $16\times16$ regression partition spans $L/16$ pixels per side, which may conceal smaller evidence. \method therefore refines selected regions hierarchically before applying regression to the remainder.

The search is based on the assumption that regions whose removal has little effect on the prediction do not need further subdivision. This assumption allows computational effort to be concentrated on selected regions, but does not guarantee that every relevant structure will be recovered. In particular, redundant evidence or interactions among spatially separated regions can make the removal of an individual region appear unimportant even when its contents contribute to the prediction.

We initialise the search with a $4\times4$ partition of the image and recursively divide selected regions into four quadrants, forming a quadtree. Let $\mathcal{Q}_\ell$ denote the regions evaluated at depth $\ell$, with $\mathcal{Q}_0$ containing the 16 initial squares. At each depth, regions are selected according to the magnitude of their score difference relative to the largest difference at that depth:
\begin{equation}
\small
\mathcal{O}_\ell=
\left\{
Q\in\mathcal{Q}_\ell:
|\Delta(Q)|>
\tau\max_{Q'\in\mathcal{Q}_\ell}|\Delta(Q')|
\right\},
\label{eq:open}
\end{equation}
where $\tau\in(0,1)$ is a relative threshold. The absolute value includes regions whose masking increases the prediction score. The children of the selected regions form $\mathcal{Q}_{\ell+1}$; regions that are not expanded remain leaves.

Since the node with the largest $|\Delta|$ satisfies \cref{eq:open} unless all differences vanish, refinement continues until the region side reaches $w_{\min}$ or $N_{\max}$ forward passes are reached. Of these $N_{\max}$ forward passes, 127 are reserved for the regression stage. If the search approaches this maximum, selected regions are expanded in descending order of $|\Delta(Q)|$.

The initial search requires one forward pass of the original image and 16 forward passes for its first partition. Each expanded node then requires four additional passes. With
\begin{equation}
\small
\ell_{\max}=
\left\lceil
\log_2\left(\frac{L}{4w_{\min}}\right)
\right\rceil,
\end{equation}
if at most $k$ nodes are expanded at each depth, the number of forward passes used by the search is bounded by
\begin{equation}
\small
N_{\mathrm{search}}\leq17+4k\ell_{\max}.
\label{eq:wafr_complexity}
\end{equation}
Doubling the image side adds one search level when $w_{\min}$ is fixed, requiring at most $4k$ additional forward passes if the number of expanded nodes per level remains bounded by $k$. For spatially diffuse evidence, more nodes may require refinement, increasing the search cost.

The score differences collected during the search define an initial attribution map. For a pixel $p$, let $\ell(p)$ denote the depth of the smallest evaluated node containing it, and let $|Q|$ be the area of node $Q$ in pixels. We define
\begin{equation}
a_{\mathrm{tree}}(p)=
\sum_{Q\ni p}
2^{-(\ell(p)-\ell(Q))}
\frac{\max(\Delta(Q),0)}{|Q|^2}.
\label{eq:pwmap}
\end{equation}
This weighting prioritises small regions with positive score differences, while reducing the influence of their coarser ancestors. The tree map therefore ranks regions identified through refinement, but does not provide a complete ordering of the remaining image.

To estimate attribution outside these regions, \method draws 127 additional random masks over the $16\times16$ image partition and applies the regression of \cref{sec:refine}. In this regime, the dense correction $d$ is computed using the target-class logit rather than its softmax probability. The final attribution ranks first the pixels belonging to leaves smaller than a regression patch and with $a_{\mathrm{tree}}>0$, ordered by their tree attribution, and subsequently all remaining pixels according to $a$ from \cref{eq:map}. If the search produces no such fine leaves, the final attribution is $a$.

For whole-slide models, the search operates on the grid of model patches rather than individual pixels. Perturbations are applied in pixel space using $b(x)$, with the tissue filter excluding fully masked tiles when they no longer satisfy its threshold. The tissue bounding box defines the search domain, excluding regions without tissue. The finest node corresponds to one 224px model patch. Unlike \cref{eq:open}, only nodes with positive score differences above the relative threshold are expanded. The procedure also evaluates the fully masked slide and respects $N_{\max}$. The 127 regression masks act on a $16\times16$ partition of the tissue bounding box, and the final map prioritises positively contributing leaves smaller than a regression region in both spatial dimensions.

When the search identifies fine leaves with positive tree attribution, they are ranked first. Remaining regions receive non-positive scores, except on whole slides, where residual positive attribution is limited to 1\% of that assigned to the selected leaves. Otherwise, we use the positive part of $a$.

%% file: sec/4_setup.tex
\section{Experimental setup}
\label{sec:setup}
\begin{figure*}[t]
  \centering
  \includegraphics[width=\linewidth]{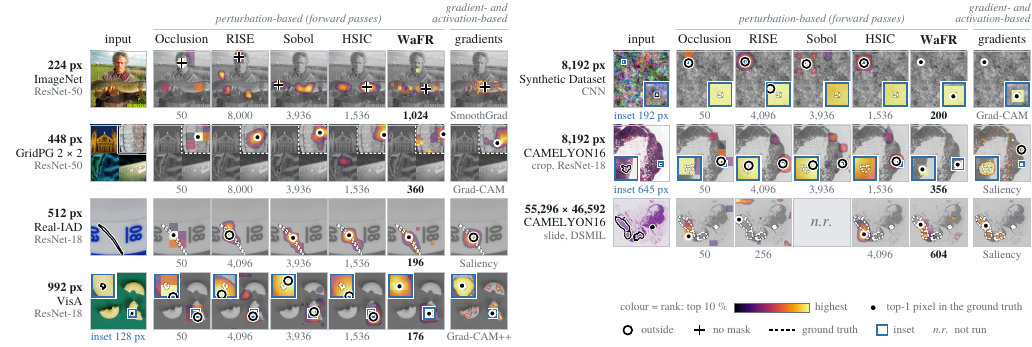}
\caption{\textbf{Attribution maps from 224\,px images to whole slides.} Rows: image side (px), dataset and model. Columns: the input with the ground truth outlined, five perturbation-based methods and the best gradient- or activation-based method. Numbers: forward passes.}
  \label{fig:qual}
\end{figure*}
\input{tables/T2_hr}
\begin{figure*}[t]
  \centering
  \includegraphics[width=\linewidth]{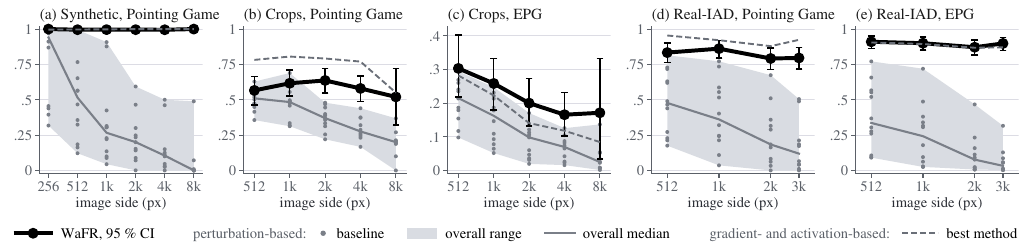}
\caption{\textbf{Localisation against image side.} (a)~Pointing Game ($\uparrow$) on the Synthetic Dataset; (b, c)~Pointing Game and EPG ($\uparrow$) on CAMELYON16 crops; (d, e)~the same on Real-IAD. Black: \method with its 95\,\% CI; grey dots: perturbation-based baselines; band: their range; grey line: their median; dashed: best gradient- or activation-based method.}
  \label{fig:side}
\end{figure*}

\nbold{Datasets.}
On ImageNet~\cite{deng2009imagenet}, we use 1,000 validation images, one per class. GridPG~\cite{rao2022towards} combines four ImageNet images of different classes into a $2\times2$ grid of 448\,px, providing a known region for each class; we use 500 grids per model. Following h-Shap~\cite{teneggi2022hshap}, we also construct a Synthetic Dataset in which the evidence is known exactly. Each scene contains a checkerboard target of side 16 to 256\,px, surrounded by squares with other periodic patterns on a random textured background, and the models predict whether the target is present.

VisA~\cite{zou2022spot} and Real-IAD~\cite{wang2024realiad} provide industrial images with pixel-level defect annotations. We use 100 defective VisA images (512-1,168\,px) and 92 Real-IAD images (512-2,992\,px). For digital pathology, CAMELYON16~\cite{bejnordi2017camelyon} provides lymph-node whole-slide images with annotated tumour regions. We evaluate 100 lesion crops from 49 slides, with sides ranging from 512 to 8,192\,px, and tumour-positive and normal CAMELYON16 test slides, with sides of 26,880 to 110,848\,px ($1.0$-$5.5\times10^9$ pixels).

\nbold{Models.}
We use publicly available ResNet-50~\cite{he2016deep} and VGG16~\cite{simonyan2015very} on ImageNet and GridPG, and DSMIL~\cite{li2021dsmil}, a dual-stream multiple-instance classifier, on whole slides. For DSMIL, perturbations modify image pixels: unchanged tiles reuse cached features, partially masked tiles are re-encoded, and fully masked tiles are typically rejected by the tissue filter. For VisA, Real-IAD and CAMELYON16 crops, we finetune ImageNet-pretrained ResNet-18 models. The industrial models aggregate spatial logits through log-sum-exp pooling, whereas the crop classifier uses an attention-based multiple-instance head~\cite{ilse2018attention}. Small CNNs are used for the Synthetic Dataset. 
We explain the predicted-class logit on ImageNet, each class logit on GridPG, and the positive-class logit on the remaining datasets. The reference input side is 224\,px: $r=1$ on ImageNet and $r>1$ elsewhere. DSMIL uses 224-px patches.

\nbold{Baselines.}
We compare \method with 12 perturbation-based attribution methods: Occlusion~\cite{zeiler2014visualizing}, RISE~\cite{petsiuk2018rise}, LIME~\cite{ribeiro2016lime}, KernelSHAP~\cite{lundberg2017unified}, PartitionSHAP~\cite{lundberg2020partition}, h-Shap~\cite{teneggi2022hshap}, ShapBPT~\cite{rashid2026shapbpt}, Sobol~\cite{fel2021sobol}, HSIC~\cite{novello2022hsic}, LIMA~\cite{chen2025lima}, MoXI~\cite{sumiyasu2024moxi} and WCAM~\cite{kasmi2023wcam}. We use their official implementations and reported configurations. A random attribution map and a centre-weighted Gaussian of standard deviation $L/4$ provide two controls.

We report Saliency~\cite{simonyan2014deep}, Integrated Gradients~\cite{sundararajan2017axiomatic}, SmoothGrad~\cite{smilkov2017smoothgrad}, Grad-CAM~\cite{selvaraju2017gradcam} and Grad-CAM++~\cite{chattopadhay2018gradcampp} as separately evaluated internal-access references. XRAI~\cite{kapishnikov2019xrai} is excluded because it relies on Integrated Gradients, while PhaseWin~\cite{gu2026phasewin} selects among a predefined number of regions and does not provide a corresponding image-size scaling configuration.

\nbold{Matched numbers of forward passes.}
We additionally compare methods at approximately \method's number of forward passes on the same synthetic images and CAMELYON16 crops.
This number sets the sample counts of RISE, HSIC, LIME and KernelSHAP, and caps PartitionSHAP, ShapBPT and HiPe~\cite{cooper2022hipe}. On whole slides, we also evaluate HiPe without a forward-pass cap. Occlusion matches it within 10\,\%; on synthetic images, Occlusion and RISE use the mean number of forward passes of \method per model and side. Sobol retains its power-of-two design on admissible grids ($3\times3$ to $7\times7$) and may stay below it. h-Shap, LIMA, MoXI and WCAM have no direct control of their forward passes and are excluded.

\nbold{Metrics.}
\emph{Localisation.}
When annotated evidence is available, we measure whether the attribution is concentrated within its spatial extent. The Pointing Game~\cite{zhang2016excitation} ($\uparrow$) counts a hit when the highest-ranked pixel lies within the target region $T$, defined by the target square on the Synthetic Dataset or the object mask on~\cite{gao2022imagenets}. Unlike the original metric, we do not use a 15-pixel tolerance and resolve ties by random selection. This avoids rewarding coarse maps merely because their maximum overlaps a tolerance region, whose relative size would also change with image resolution.

The energy-based Pointing Game (EPG)~\cite{wang2020scorecam} ($\uparrow$) measures the proportion of positive attribution within the bounding box of $T$, while the GridPG score~\cite{bohle2021coda} ($\uparrow$) measures this proportion within the cell associated with the target class. We additionally compute the AUROC ($\uparrow$), comparing attribution values inside $T$ with those in a local window obtained by enlarging its bounding box by $4\sqrt{|T|}$ pixels per side. Restricting the negative pixels to this neighbourhood prevents distant background regions from dominating the score on very large images. For whole slides, the same local AUROC is computed at the resolution of the model patches, and the Pointing Game counts a hit when the top-ranked model patch is at least half tumour, on the tumour slides that contain such a patch. These metrics assess correspondence with annotated regions, which need not coincide with all the evidence used by the classifier.

\emph{Fidelity.}
We measure the response of the model when image regions are removed or restored according to their attribution ranking. At 224\,px, we use Deletion ($\downarrow$) and Insertion ($\uparrow$)~\cite{petsiuk2018rise}, with steps of 1\,\% of the pixels (502 pixels) and 101 points per curve, together with $\mu$Fidelity ($\uparrow$)~\cite{bhatt2020evaluating}. The latter measures, as in~\cite{novello2022hsic}, the Spearman correlation between the attribution assigned to randomly selected patches and the corresponding prediction-score changes.

For high-resolution images, we use logarithmically spaced Deletion steps because relevant evidence may occupy less than $0.01\,\%$ of the image and be removed entirely by the first step of a linearly spaced curve. On whole slides, Deletion progressively removes the highest-ranked $0.1$-$10\,\%$ of tissue patches and averages the tumour-class probability along the resulting curve. Since deletion may introduce inputs outside the training distribution~\cite{rong2022road}, we report fidelity alongside localisation rather than treating either as sufficient on its own.

\emph{Computational cost.}
We report forward passes and execution time. On whole slides, each forward pass denotes a DSMIL bag evaluation, reusing cached features for unchanged tiles and re-encoding partially masked tiles.

\nbold{Statistical analysis.}
We report mean scores with 95\,\% bootstrap confidence intervals for high-resolution experiments. At standard resolution, confidence interval half-widths do not exceed .02 and are omitted for brevity. Comparisons with \method use paired two-sided sign-flip tests, with Holm correction~\cite{holm1979simple} across the baselines of each metric. On whole slides, paired differences are computed only on slides explained by both methods and reported with 95\,\% bootstrap confidence intervals. Failure to reject a difference is not interpreted as evidence of equivalence.

\input{tables/T23}

\nbold{Implementation details.}
We set $\tau=0.25$ and $N_{\max}=2{,}048$ at $r>1$, with $w_{\min}=8$\,px (one 224-px model patch on slides). The patch and dense weights were tuned once on ResNet-50 at 224\,px, on held-out ImageNet images, and then used unchanged across datasets. All experiments are implemented in PyTorch and executed on NVIDIA H100 NVL GPUs with 94\,GB of memory. Methods exceeding the available memory or execution-time limits are not run. Further implementation details and parameter sensitivity analyses are provided in the supplementary material.

%% file: tables/T2_hr.tex
\begin{table*}[t]
\centering
\tabstyle\renewcommand{\arraystretch}{1}%
\caption{\textbf{High-resolution images.} Mean over image sides; slides: patch-level AUROC on the tumour slides and execution time on one slide; log Del.: Deletion with log-spaced steps; Growth: forward passes for patches of a predefined size ($k$: relevant nodes per depth). $\pm$: 95\,\% CI; bold: best; underline: second; $^{*}$: not statistically different from \method (paired sign-flip test, Holm correction); --: not run.}
\label{tab:hr}
\footnotesize
\setlength{\tabcolsep}{2pt}
\begin{tabular}{@{}l ccc ccc ccc cc c@{}}
\toprule
 & \multicolumn{3}{c}{\tabhead{VisA}} & \multicolumn{3}{c}{\tabhead{Real-IAD}} & \multicolumn{3}{c}{\tabhead{CAMELYON16 crops}} & \multicolumn{2}{c}{\tabhead{CAMELYON16 slides}} & \\
\cmidrule(lr){2-4}\cmidrule(lr){5-7}\cmidrule(lr){8-10}\cmidrule(lr){11-12}
\tabhead{Method} & \tabhead{AUROC $\uparrow$} & \tabhead{EPG $\uparrow$} & \tabhead{log Del.\ $\downarrow$} & \tabhead{AUROC $\uparrow$} & \tabhead{EPG $\uparrow$} & \tabhead{log Del.\ $\downarrow$} & \tabhead{AUROC $\uparrow$} & \tabhead{EPG $\uparrow$} & \tabhead{log Del.\ $\downarrow$} & \tabhead{AUROC $\uparrow$} & \tabhead{Time (s) $\downarrow$} & \tabhead{Growth} \\
\midrule
\csname @@input\endcsname tables/T2_hr_body.tex %
\bottomrule
\end{tabular}
\end{table*}

%% file: tables/T23.tex
\begin{table*}[t]
\centering
\tabstyle\renewcommand{\arraystretch}{1}\renewcommand{\ci}[1]{}\renewcommand{\dci}[2]{}%
\caption{\textbf{(a) ImageNet (224\,px) and GridPG; (b) ablations.}
(a)~Bold/underline: best/second; $^{*}$: no significant
difference from \method. (b)~Differences relative to
\method ($+$ favours \method); VisA: mean over image--side pairs; A3 on slides: identical maps due to tissue filtering.}
\label{tab:std}\label{tab:ablation}
\footnotesize
\begin{minipage}[t]{.46\linewidth}\vspace{0pt}\raggedright
\setlength{\tabcolsep}{2.2pt}
\begin{tabular}{@{}l ccc ccc c@{}}
\toprule
\tabhead{(a)} & \multicolumn{3}{c}{\tabhead{ResNet-50}} & \multicolumn{3}{c}{\tabhead{VGG16}} & \tabhead{GridPG} \\
\cmidrule(lr){2-4}\cmidrule(lr){5-7}\cmidrule(lr){8-8}
\tabhead{Method} & \tabhead{Del.\ $\downarrow$} & \tabhead{Ins.\ $\uparrow$} & \tabhead{$\mu$Fid.\ $\uparrow$} & \tabhead{Del.\ $\downarrow$} & \tabhead{Ins.\ $\uparrow$} & \tabhead{$\mu$Fid.\ $\uparrow$} & \tabhead{share $\uparrow$} \\
\midrule
\csname @@input\endcsname tables/T3_std_body.tex %
\bottomrule
\end{tabular}\\[4pt]
\end{minipage}\hfill
\begin{minipage}[t]{.535\linewidth}\vspace{0pt}\raggedleft
\setlength{\tabcolsep}{2.5pt}\global\tabfirsttrue
\begin{tabular}{@{}l rrrrr@{}}
\toprule
\tabhead{(b)} & \multicolumn{3}{c}{\tabhead{VisA}} & \multicolumn{1}{c}{\tabhead{Synthetic}} & \multicolumn{1}{c}{\tabhead{Slides}} \\
\cmidrule(lr){2-4}\cmidrule(lr){5-5}\cmidrule(l){6-6}
\tabhead{$r>1$} & \tabhead{AUROC $\uparrow$} & \tabhead{EPG $\uparrow$} & \tabhead{log Del.\ $\downarrow$} & \tabhead{PG $\uparrow$} & \tabhead{AUROC $\uparrow$} \\
\midrule
\method & .916 & .668 & .785 & .997 & .949 \\
\treemid \ablid{A1}$-$ quadtree & +.015 & {+.422} & +.021 & \textbf{+.331} & +.009 \\
\treemid \ablid{A3}$-$ Haar baseline & +.001 & {+.093} & $-$.042 & +.000 & \na \\
\treemid \ablid{A4}$-$ Haar wavelets & +.001 & +.000 & $-$.001 & +.000 & +.001 \\
\treelast \ablid{A6}$-$ random masks & {+.120} & +.004 & $-$.033 & +.000 & {+.041} \\
\midrule
 & \multicolumn{2}{c}{\tabhead{ResNet-50}} & \multicolumn{2}{c}{\tabhead{VGG16}} & \\
\cmidrule(lr){2-3}\cmidrule(lr){4-5}
\tabhead{$r=1$} & \tabhead{Del.\ $\downarrow$} & \tabhead{Ins.\ $\uparrow$} & \tabhead{Del.\ $\downarrow$} & \tabhead{Ins.\ $\uparrow$} & \\
\midrule
\method & .069 & .717 & .049 & .640 & \\
\treemid \ablid{A3}$-$ Haar baseline & +.005 & {+.015} & +.006 & +.008 & \\
\treemid \ablid{A4}$-$ Haar wavelets & $-$.012 & {+.019} & $-$.005 & {+.026} & \\
\treemid \ablid{A3+A4}$-$ both & $-$.012 & {+.045} & $-$.004 & {+.051} & \\
\treemid \ablid{A5}$-$ sparse regression & {+.048} & $-$.008 & {+.048} & $-$.032 & \\
\treelast \ablid{A7}$-$ probability in $d$ & +.003 & {+.012} & +.002 & {+.010} & \\
\bottomrule
\end{tabular}
\end{minipage}
\end{table*}

%% file: sec/5_results.tex
\section{Results}
\label{sec:results}
\input{tables/T5_equal}

\subsection{Efficiency}
\label{sec:efficiency}

\nbold{Computational cost and image resolution.}
On the Synthetic Dataset, where each image contains a single target, \method requires an average of 4.7 additional forward passes per doubling of the image side (95\% CI: 4.4-5.1), from 168 at 256\,px to 188-222 at 8,192\,px, while refining to 8-px nodes and maintaining Pointing Game $\geq .995$. Baselines that enlarge their perturbation regions with $L$ fall to at most .07 at 8,192\,px (\cref{fig:side}a). For an additive classifier with fixed-size targets ranging from 1 to 16, forward passes increase with target count (Spearman's $\rho\geq .95$), from 178 to 585 at 1,024\,px and from 201 to 772 at 8,192\,px, while target recall remains $\geq .95$ and $N_{\max}$ is never reached. This is consistent with the dependence of \cref{eq:wafr_complexity} on the number of expanded regions. Keeping patches of a predefined size would require forward passes growing with $L^2$ (Occlusion, WCAM, KernelSHAP) or $L^4$ (LIMA, MoXI; \cref{tab:hr}).

\nbold{Whole-slide images.}
On whole slides, \method requires a median of 176 forward passes (range: 166-902). On the slide timed in \cref{tab:hr}, it uses 458 passes, against 27,968 for h-Shap and 4,096 for HSIC. Occlusion and PartitionSHAP are faster on this slide, with 50 and 500 passes, but at lower AUROC. At its official settings and without a forward-pass cap, HiPe completes 16 of 20 slides, with a median of 54,726 passes (range: 13,881-702,419); the remaining four do not complete within 870,000 passes. On the eight tumour slides that HiPe completes, its patch-level AUROC (.924) and Pointing Game (.875) are lower than those of \method on the same slides (.961 and 1.000), whose AUROC overall is .949. Other methods face limitations at this scale: ShapBPT exceeds the available memory, with a projected requirement of 1.8\,TB, while the official Sobol design becomes degenerate under the evaluated slide configuration.

\subsection{Localisation and fidelity}
\label{sec:localisation}

\nbold{Whole-slide localisation.}
On the tumour slides, \method achieves the highest mean patch-level AUROC among the evaluated perturbation-based methods, .949, compared with .922 for h-Shap and .797 for HSIC (\cref{tab:hr}). Its paired improvements are .32 over Occlusion, .34 over RISE, .20 to .26 over LIME, PartitionSHAP, LIMA and MoXI, and .15 over HSIC ($p\leq.001$); KernelSHAP and h-Shap are .17 and .03 below \method on them. On the tumour slides that the model classifies as tumour, \method also achieves better slide Deletion than Occlusion, RISE and h-Shap, with paired differences of .03 [95\% CI: .02, .05], .05 [.03, .07] and .06 [.04, .08], respectively. These results support its ability to identify tumour-associated regions with relatively few forward passes, without establishing superiority over every perturbation-based baseline. Grad-CAM, which accesses the model's gradients and intermediate representations, achieves better tumour localisation and lower slide Deletion than \method.

\nbold{High-resolution images.}
The relative performance of attribution methods depends on whether localisation or fidelity is measured. On VisA and Real-IAD, \method achieves the highest EPG among perturbation-based methods (.648 and .895, respectively; $p<.002$ against each baseline). This ranking persists when all methods retain only positive attributions within the top 0.1, 0.5, 1, 5 or 10\% of pixels; even at 0.1\% on VisA, \method exceeds RISE (.658 vs. .585; Holm-adjusted $p=.018$). Retaining regression values outside refined leaves also preserves this ranking (EPG .620 and .865). On CAMELYON16 crops, \method ranks second to PartitionSHAP (.220 vs. .244; not significant), but exceeds the remaining perturbation-based methods.
For log-scale Deletion, \method achieves the best score on Real-IAD and ranks second on VisA and CAMELYON16 crops, without significant differences from the respective best methods. On VisA, its Deletion is lower than that of five perturbation-based methods, including RISE and LIME ($p\leq.003$), and than HSIC in a separate test on stored maps. In AUROC, \method performs comparably to RISE on VisA (.919 vs. .923) and Real-IAD (.957 vs. .959), with no significant differences, while Sobol and ShapBPT obtain higher scores on CAMELYON16 crops (.752 and .739 vs. .684).

\nbold{Images at the reference resolution.}
At 224\,px, \method achieves the highest $\mu$Fidelity on both ResNet-50 and VGG16, .537 and .570, compared with .494 and .533 for PartitionSHAP, the next best method ($p<.002$ against every perturbation-based baseline; \cref{tab:std}). On VGG16, it also obtains the lowest Deletion (.049), comparable to PartitionSHAP (.051). On GridPG, it achieves the highest share, .905, compared with .846 for KernelSHAP, the next best method ($p<.002$ against every perturbation-based baseline). However, RISE and PartitionSHAP achieve higher Insertion on both models, and PartitionSHAP obtains lower Deletion on ResNet-50.

\nbold{Localisation across image sizes.}
\Cref{fig:side}b-e compares Pointing Game and EPG as image resolution increases on CAMELYON16 crops and Real-IAD. The Pointing Game of \method remains between .52 and .64 on the crops and between .79 and .86 on Real-IAD, whereas several perturbation-based methods deteriorate with resolution. For instance, HSIC falls from .63 to .07 on Real-IAD. RISE is the closest on Real-IAD: its Pointing Game is comparable to that of \method at 512\,px, and \method is higher by .090, .118 and .291 at 1,024, 2,048 and 2,992\,px ($p\leq.028$). \method also maintains the highest EPG among the perturbation-based methods at each evaluated resolution, although no difference is significant at 8,192\,px after Holm correction. This behaviour does not extend uniformly to local AUROC: on CAMELYON16 crops, Sobol and ShapBPT frequently obtain higher scores, while on Real-IAD RISE is level with \method at every resolution. \Cref{fig:qual} illustrates the attribution maps across image sizes.

\nbold{Matched numbers of forward passes.}
At matched numbers of forward passes (\cref{tab:equal}), on the Synthetic Dataset, \method maintains a Pointing Game of at least .995 up to 8,192\,px and exceeds every evaluated baseline by at least .17 from 1,024\,px onwards. Its EPG is also higher by at least .55 at every image size. At 512\,px, however, RISE achieves a comparable Pointing Game, and HiPe locates the target slightly more often (.999 against .996, $p<.05$ after Holm correction). As resolution increases, the Pointing Game of HiPe and RISE falls to .036 and .250 at 8,192\,px, respectively, while that of \method remains at least .995.

\noindent On CAMELYON16 crops, the differences are smaller, but \method has the higher Pointing Game and EPG in 79 of the 80 comparisons of \cref{tab:equal}; the exception is HiPe at 512\,px.

\subsection{Ablation tests}
\label{sec:ablations}\label{sec:res_ablation}

\Cref{tab:ablation} evaluates the contribution of each component of \method. Removing the quadtree (A1) reduces the Pointing Game on the Synthetic Dataset from .997 to .665 and the EPG on VisA from .668 to .246. On whole slides, A1 keeps a similar AUROC (.009 lower) with at least as many forward passes, but its Pointing Game on the tumour slides falls from 1.000 to .600. The hierarchical search is therefore responsible for much of the fine-scale localisation obtained at high resolution.
The Haar components contribute at different stages. Compared with constant grey, the Haar low-pass baseline improves EPG on VisA by .093 and reduces mean search forward passes from 143 to 98. The Haar dictionary improves Insertion at $r=1$: removing it (A4) lowers Insertion by .019 and .026, while removing both components (A3+A4) yields larger decreases of .045 and .051. At $r>1$, the dictionary primarily affects the ranking outside refined leaves, with negligible changes in Pointing Game and EPG.
The random-mask regression instead affects the ranking of regions not explicitly resolved by the search. Removing it (A6) leaves the Synthetic Dataset Pointing Game unchanged, but decreases AUROC on VisA by .120 and on whole slides by .041. Conversely, the search alone achieves better log-scale Deletion on VisA while requiring 127 fewer forward passes. At $r=1$, removing sparse regression (A5) increases Deletion by .048 on both evaluated models, while changes to the Haar representation produce opposing effects on Deletion and Insertion.

%% file: tables/T5_equal.tex
\begin{table}[t]
\centering
\tabstyle\renewcommand{\arraystretch}{1}%
\caption{\textbf{Attribution at matched forward-pass budgets.}
The first rows report \method's mean forward passes and scores.
Remaining methods use approximately the same forward-pass budgets; score differences are computed on the subset of images evaluated by both methods (positive values favour \method).}
\label{tab:equal}
\footnotesize
\setlength{\tabcolsep}{1.2pt}
\begin{tabular}{@{}l ccccc ccccc@{}}
\toprule
 & \multicolumn{5}{c}{\tabhead{Synthetic Dataset}} & \multicolumn{5}{c}{\tabhead{CAMELYON16 crops}} \\
\cmidrule(lr){2-6}\cmidrule(l){7-11}
\tabhead{Side (px)} & \tabhead{512} & \tabhead{1k} & \tabhead{2k} & \tabhead{4k} & \tabhead{8k} & \tabhead{512} & \tabhead{1k} & \tabhead{2k} & \tabhead{4k} & \tabhead{8k} \\
\midrule
\method{} passes & 175 & 191 & 220 & 224 & 195 & 316 & 369 & 407 & 487 & 559 \\
\method PG & .996 & .996 & .996 & .995 & 1.000 & .566 & .617 & .636 & .580 & .520 \\
\method EPG & .934 & .935 & .939 & .939 & .953 & .303 & .258 & .200 & .166 & .171 \\
\csname @@input\endcsname tables/T5_equal_body.tex %
\bottomrule
\end{tabular}
\end{table}

%% file: sec/6_conclusion.tex
\section{Conclusion}
\label{sec:conclusion}
We introduced \method, a model-agnostic attribution method combining adaptive hierarchical search with sparse multiscale regression. Controlled experiments show that its computational cost depends on the distribution of evidence, with limited growth across image resolutions when few regions require refinement. On gigapixel whole-slide images, \method substantially reduces forward passes while retaining competitive localisation and fidelity. Hierarchical search resolves fine-scale evidence, while regression ranks unresolved regions. These advantages remain conditional on the evidence distribution: redundant or interacting regions may escape refinement, and diffuse evidence increases cost. Future work will investigate best-first refinement and finer within-patch attribution.

%% file: supplementary_list.tex
\makeatletter
\newsavebox\supp@fitbox
\newenvironment{suppfit}{\begin{lrbox}{\supp@fitbox}}{\end{lrbox}%
  \ifdim\wd\supp@fitbox>\linewidth\resizebox{\linewidth}{!}{\usebox{\supp@fitbox}}\else\usebox{\supp@fitbox}\fi}
\@ifundefined{sidewaysfigure*}{%
  \newsavebox\suppsidebox
  \newenvironment{sidewaysfigure*}%
    {\begin{figure*}[p]\begin{lrbox}{\suppsidebox}\begin{minipage}{\textheight}}%
    {\end{minipage}\end{lrbox}\centering\rotatebox{90}{\usebox{\suppsidebox}}\end{figure*}}%
}{}
\@ifundefined{FloatBarrier}{\def\FloatBarrier{\par\begingroup\let\@elt\relax\edef\@tempa{\@botlist\@deferlist\@dbldeferlist}\ifx\@tempa\@empty\else\clearpage\fi\endgroup}}{}
\newdimen\supp@colmax\newif\ifsupp@keepcol
\def\suppneedspace#1{\par\penalty-100\begingroup\dimen@=#1\relax\dimen@ii=\pagegoal\advance\dimen@ii-\pagetotal
  \ifdim\dimen@>\dimen@ii\break\fi\endgroup}
\newsavebox\supp@cbox
\let\supp@colpre\@gobble
\newenvironment{suppcol}[1]{\par\supp@colpre{#1}\addvspace{\intextsep}\begin{lrbox}{\supp@cbox}\begin{minipage}{\linewidth}\def\@captype{#1}%
  \captionsetup{hypcap=false}\centering}%
  {\end{minipage}\end{lrbox}%
   \ifdim\dimexpr\ht\supp@cbox+\dp\supp@cbox\relax>\@colht
     \typeout{SUPPCOL: box taller than the column left on page \thepage: new page}\clearpage\fi
   \if@firstcolumn\else\ifdim\dimexpr\ht\supp@cbox+\dp\supp@cbox\relax>\dimexpr\pagegoal-\pagetotal\relax
     \global\supp@keepcoltrue\fi\fi
   \noindent\usebox{\supp@cbox}%
   \ifdim\dimexpr\ht\supp@cbox+\dp\supp@cbox\relax>\supp@colmax \global\supp@colmax=\dimexpr\ht\supp@cbox+\dp\supp@cbox\relax\fi
   \par\addvspace{\intextsep}}
  \newsavebox\supp@wbox\newdimen\supp@wneed\newdimen\supp@wfree
  \newcount\supp@qin \newcount\supp@qout \newcount\supp@qi \supp@qin=0 \supp@qout=0
  \newif\ifsupp@fits
  \def\supp@qbox#1{\csname supp@q@\number#1\endcsname}%
  \def\supp@fitw#1{%
    \supp@wneed=\dimexpr\ht#1+\dp#1+2\stripsep\relax
    \if@firstcolumn \supp@wfree=\dimexpr\@colht-.5\pagetotal\relax
    \else \supp@wfree=\dimexpr.5\@colht-.5\pagetotal\relax \fi
    \ifdim\pagetotal=\z@ \advance\supp@wneed-\stripsep \fi
    \ifdim\supp@wneed>\supp@wfree \supp@fitsfalse \else \supp@fitstrue \fi
    \ifdim\pagetotal>\z@ \ifdim\supp@colmax>\z@ \supp@fitsfalse \fi\fi}%
  \def\supp@fitc#1{%
    \supp@wneed=\dimexpr\ht#1+\dp#1+\intextsep\relax
    \ifdim\pagetotal=\z@ \supp@wfree=\@colht \else \supp@wfree=\dimexpr\pagegoal-\pagetotal\relax \fi
    \ifdim\supp@wneed>\supp@wfree \supp@fitsfalse \else \supp@fitstrue \fi}%
  \AddToHook{shipout/after}{\ifsupp@keepcol\global\supp@keepcolfalse\else\global\supp@colmax=\z@\fi}%
  \def\supp@clearbal{\par\ifdim\pagetotal>\z@\ifdim\supp@colmax=\z@\begin{strip}\end{strip}\fi\fi\clearpage}%
  \def\supp@emitw#1{\begin{strip}\centering\usebox{#1}\end{strip}\global\supp@colmax=\z@}%
  \def\supp@emitc#1{\par\addvspace{\intextsep}%
    \ifdim\dimexpr\ht#1+\dp#1\relax>\@colht \clearpage\fi
    \if@firstcolumn\else\ifdim\dimexpr\ht#1+\dp#1\relax>\dimexpr\pagegoal-\pagetotal\relax \global\supp@keepcoltrue\fi\fi
    \noindent\usebox{#1}%
    \ifdim\dimexpr\ht#1+\dp#1\relax>\supp@colmax \global\supp@colmax=\dimexpr\ht#1+\dp#1\relax\fi
    \par\addvspace{\intextsep}}%
  \def\supp@enqueue#1#2{%
    \typeout{SUPPWAIT: #1 \@captype\space\csname the\@captype\endcsname\space on page \thepage\space waits (needs \the\supp@wneed, free \the\supp@wfree)}%
    \expandafter\ifx\csname supp@q@\number\supp@qin\endcsname\relax
      \expandafter\newsavebox\csname supp@q@\number\supp@qin\endcsname\fi
    \global\setbox\supp@qbox\supp@qin=\box\supp@wbox
    \expandafter\xdef\csname supp@qt@\number\supp@qin\endcsname{#2}%
    \expandafter\xdef\csname supp@qk@\number\supp@qin\endcsname{#1}%
    \global\advance\supp@qin\@ne}%
  \def\supp@flush#1{\def\supp@blocked{,}\supp@qi=\supp@qout \supp@flushloop{#1}\supp@advance}%
  \def\supp@advance{\@tempswafalse %
    \ifnum\supp@qout<\supp@qin \expandafter\ifx\csname supp@qd@\number\supp@qout\endcsname\relax\else\@tempswatrue\fi\fi
    \if@tempswa\global\advance\supp@qout\@ne\expandafter\supp@advance\fi}%
  \def\supp@flushloop#1{%
    \ifnum\supp@qi<\supp@qin
      \expandafter\ifx\csname supp@qd@\number\supp@qi\endcsname\relax
        \edef\supp@t{\csname supp@qt@\number\supp@qi\endcsname}%
        \edef\supp@k{\csname supp@qk@\number\supp@qi\endcsname}%
        \edef\@tempa{\noexpand\in@{,\supp@t,}{\supp@blocked}}\@tempa
        \ifin@\else
          \if w\supp@k \supp@fitw{\supp@qbox\supp@qi}\else \supp@fitc{\supp@qbox\supp@qi}\fi
          \ifsupp@fits\else
            \ifnum#1=1\relax \if w\supp@k\supp@clearbal\else\newpage\fi \supp@fitstrue \fi
            \ifnum#1=2\relax \if w\supp@k
              \ifdim\dimexpr\ht\supp@qbox\supp@qi+\dp\supp@qbox\supp@qi\relax<.5\textheight \clearpage\supp@fitstrue \fi
              \else \newpage\supp@fitstrue \fi\fi
          \fi
          \ifsupp@fits
            \if w\supp@k \supp@emitw{\supp@qbox\supp@qi}\else \supp@emitc{\supp@qbox\supp@qi}\fi
            \expandafter\gdef\csname supp@qd@\number\supp@qi\endcsname{}%
          \else \edef\supp@blocked{\supp@blocked\supp@t,}\fi
        \fi
      \fi
      \advance\supp@qi\@ne \expandafter\supp@flushloop\expandafter{\number#1}%
    \fi}%
  \def\supp@place#1{%
    \supp@flush0%
    \@tempswafalse\supp@qi=\supp@qout
    \loop\ifnum\supp@qi<\supp@qin
      \expandafter\ifx\csname supp@qd@\number\supp@qi\endcsname\relax
        \edef\@tempa{\csname supp@qt@\number\supp@qi\endcsname}\edef\@tempb{\@captype}%
        \ifx\@tempa\@tempb\@tempswatrue\fi\fi
      \advance\supp@qi\@ne\repeat
    \if w#1\supp@fitw\supp@wbox\else\supp@fitc\supp@wbox\fi
    \if@tempswa\supp@fitsfalse\fi
    \ifsupp@fits \if w#1\supp@emitw\supp@wbox\else\supp@emitc\supp@wbox\fi
    \else \supp@enqueue{#1}{\@captype}\fi}%
  \renewenvironment{suppcol}[1]{\par\def\@captype{#1}\begin{lrbox}{\supp@wbox}\begin{minipage}{\linewidth}%
    \captionsetup{hypcap=false}\centering}{\end{minipage}\end{lrbox}\supp@place c}%
  \pretocmd{\paragraph}{\par\supp@flush0}{}{}%
  \pretocmd{\subsection}{\par\supp@flush0}{}{}%
  \pretocmd{\section}{\par\supp@flush0}{}{}%
  \def\suppflushall{\par\supp@flush1}%
\@ifundefined{strip}{%
  \newenvironment{suppwide}[1]{\def\supp@wtype{#1}\csname #1*\endcsname[!t]\centering}{\csname end\supp@wtype*\endcsname}%
  \AddToHook{env/suppwide/end}{\expandafter\xdef\csname supp@wn@\supp@wtype\endcsname{\the\value{\supp@wtype}}}%
  \pretocmd{\supp@place}{\xdef\supp@pt{\@captype}\xdef\supp@pn{\the\value{\@captype}}%
    \expandafter\xdef\csname supp@qn@\number\supp@qin\endcsname{\the\value{\@captype}}}{}{}%
  \newif\ifsupp@pend
  \def\supp@pendtest#1{\global\supp@pendfalse\begingroup\def\@elt##1{\@tempcnta\count##1\divide\@tempcnta32
    \ifnum\@tempcnta=\csname ftype@#1\endcsname\relax\global\supp@pendtrue\fi}\@deferlist\@dbldeferlist\endgroup}%
  \let\supp@emitcin\supp@emitc
  \def\supp@emitc#1{\begingroup\def\@tempa{#1}\def\@tempb{\supp@wbox}%
    \ifx\@tempa\@tempb \edef\supp@et{\supp@pt}\edef\supp@en{\supp@pn}%
    \else \edef\supp@et{\supp@t}\edef\supp@en{\csname supp@qn@\number\supp@qi\endcsname}\fi
    \supp@pendtest\supp@et
    \ifsupp@pend \ifnum0\csname supp@wn@\supp@et\endcsname<\supp@en\relax\else\global\supp@pendfalse\fi\fi
    \ifsupp@pend \csname\supp@et\endcsname[!htb]\centering\usebox{#1}\csname end\supp@et\endcsname
    \else \supp@emitcin{#1}\fi\endgroup}%
}{%
  \nolinenumbers
  \newenvironment{suppwide}[1]{\par\def\@captype{#1}\begin{lrbox}{\supp@wbox}\begin{minipage}{\textwidth}%
    \captionsetup{hypcap=false}\centering}{\end{minipage}\end{lrbox}\supp@place w}%
}
\makeatother
\input{supplementary/supp_0_intro}
\input{supplementary/supp_A_method}
\FloatBarrier
\input{supplementary/supp_B_roles}
\input{supplementary/supp_C_protocol}

\FloatBarrier
\input{supplementary/supp_D_transfer}

\input{supplementary/supp_E_limits}

\input{supplementary/supp_F_tables}
\FloatBarrier

\input{supplementary/supp_G_galleries}
\suppflushall

%% file: supplementary/supp_0_intro.tex
\clearpage
\maketitlesupplementary
\appendix
\setcounter{figure}{0}\setcounter{table}{0}\setcounter{equation}{0}\setcounter{algorithm}{0}
\renewcommand{\thetable}{S\arabic{table}}\renewcommand{\theHtable}{S\arabic{table}}
\renewcommand{\thefigure}{S\arabic{figure}}\renewcommand{\theHfigure}{S\arabic{figure}}
\renewcommand{\theequation}{S\arabic{equation}}\renewcommand{\theHequation}{S\arabic{equation}}
\renewcommand{\thealgorithm}{S\arabic{algorithm}}\renewcommand{\theHalgorithm}{S\arabic{algorithm}}

\begin{suppcol}{table}
\centering
\tabstyle\renewcommand{\arraystretch}{1.15}
\caption{\textbf{Reading guide.} Main paper: section or table; right: section of this supplement.}
\label{tab:supp_guide}
\footnotesize
\setlength{\tabcolsep}{4pt}
\begin{tabular}{>{\raggedright\arraybackslash}m{.73\linewidth} r}
\toprule
\multicolumn{1}{l}{\tabhead{Main paper}} & \multicolumn{1}{r}{\tabhead{Section}} \\
\midrule
Method (\cref{sec:method}): regimes, baseline, search, number of forward passes, final map & \cref{sec:supp_impl} \\
\rowcolor{black!5}
Ablations (\cref{tab:ablation}): role of each component, regime rule, parameters & \cref{sec:supp_ablation} \\
Setup (\cref{sec:setup}): datasets, models, baselines, metrics, statistics, cost & \cref{sec:supp_protocol} \\
\rowcolor{black!5}
Whole slides (\cref{sec:results}): generalisation to a second slide model and to other hospitals, cached features; object detection on DOTA & \cref{sec:supp_more} \\
Limitations (\cref{sec:conclusion}): redundant evidence & \cref{sec:supp_limits} \\
\rowcolor{black!5}
\Cref{tab:hr,tab:std,tab:equal}, \cref{fig:side}: complete tables & \cref{sec:supp_results} \\
\Cref{fig:qual}: attribution maps of every method & \cref{sec:supp_gallery} \\
\bottomrule
\end{tabular}
\end{suppcol}

\noindent \Cref{tab:supp_guide} maps each section of this supplement to the part of the main paper it completes. Beyond specifying the algorithm, component ablations and experimental protocol, the additional evaluations characterise \method across interfaces, domains and failure modes: it retains the highest patch-level AUROC and Pointing Game when all whole-slide methods share the same cached-feature interface (\cref{tab:supp_cached_wsi}), preserves its localisation without parameter retuning on a second slide model, on slides from five additional hospitals and with an object detector on large aerial images (\cref{sec:supp_more}), covers a larger fraction of annotated foci on multi-focus slides (\cref{sec:supp_ablation}), and remains consistent on ViT-B/16 and DeiT-B/16 (\cref{tab:supp_imagenet_vit}) as well as under a common top-$q$\,\% post-processing and signed map composition (\cref{tab:supp_topq,tab:supp_sign}). We also illustrate how single-region search behaves under redundant evidence and targets centred on quadtree boundaries (\cref{sec:supp_limits}).

%% file: supplementary/supp_A_method.tex
\section{Method details}
\label{sec:supp_impl}
\label{sec:supp_method}
\label{sec:supp_params}
This section specifies the two regimes and the baseline (\cref{sec:supp_regimes}), the hierarchical search (\cref{sec:supp_search_spec}) and the regression with the final map (\cref{sec:supp_regression}). The parameters of \method (\cref{tab:supp_params}) are the same on every image dataset; on whole slides, the nodes and the maximum of the search differ (last rows). \Cref{alg:supp_method} summarises both regimes.

\begin{suppcol}{table}
\centering
\tabstyle
\caption{\textbf{Parameters of \method.} Regr.: sparse regression.}
\label{tab:supp_params}
\footnotesize
\setlength{\tabcolsep}{2.5pt}
\begin{tabular}{@{}l l l@{}}
\toprule
\tabhead{Part} & \tabhead{Parameter} & \tabhead{Value} \\
\midrule
Rule & search if & $r=L/s_{\text{in}}>1$ \\
$b(x)$ & Haar low-pass side & $L/16$ \\
\addlinespace[2pt]
Tree & root partition & $4\times4$ nodes \\
 & relative threshold $\tau$ & 0.25 of the depth maximum $|\Delta|$ \\
 & finest node side & $\le w_{\min}=8$\,px (4 to 8\,px) \\
 & max.\ passes $N_{\max}$ & 2,048 (search $\le$ 1,921) \\
 & stop & no relevant node, finest side, $N_{\max}$ \\
\addlinespace[2pt]
Regr. & patch grid & $16\times16$ \\
 & mask sampling & $7\times7$ Bernoulli$(0.5)$, bilinear, random shift \\
 & patch retained if & mean interpolated value $\geq 0.5$ \\
 & masks $n$ & 1,023 ($r\leq1$); 127 ($r>1$) \\
 & total passes & 1,024 ($r\leq1$); $\le$ 2,048 ($r>1$) \\
 & regression & $\ell_1$, 5-fold CV over 30 penalties \\
 & columns & 256 patches (weight 2) + 256 Haar \\
 & dense term $d$ & probability ($r=1$), logit ($r>1$) \\
 & smoothing of $d$ & Gaussian, $\sigma$ = half a patch \\
 & weight of $d$ & 0.15 \\
\addlinespace[2pt]
Slides & nodes & groups of patches in the tissue box \\
 & search maximum & 2,048, plus 127 masks (total $\le$ 2,175) \\
\bottomrule
\end{tabular}
\end{suppcol}

\subsection{Regimes and baseline}
\label{sec:supp_regimes}
\paragraph{Two regimes.} Let $s_{\text{in}}$ be the reference input side of the model and $r=L/s_{\text{in}}$ (\cref{sec:method} of the main paper). At $r\le1$, \method fits the sparse regression of \cref{sec:refine} of the main paper on 1,023 random masks, which with the forward pass of $x$ gives 1,024 forward passes. At $r>1$, it runs the hierarchical search of \cref{sec:find} of the main paper and then fits the same regression on 127 random masks, with at most $N_{\max}=2{,}048$ forward passes in total.

\paragraph{Baseline.} The baseline $b(x)$ partitions the image into $16\times16$ squares and replaces every pixel with the mean of its square. When $L/16$ is a power of two, this is the Haar low-pass approximation of $x$ at side $L/16$~\cite{mallat1989theory}: replacing a region made of whole squares with $b(x)$ removes the Haar detail coefficients finer than the squares and keeps their means. Both the search and the regression mask with $b(x)$.

\paragraph{Image dimensions.}\label{sec:supp_quadtree} When the image dimensions do not align with the $16\times16$ squares or with the quadtree, boundaries are rounded to whole pixels, so nodes at the same depth differ in width or height by at most one pixel. Non-square images are split along each axis separately.

\subsection{Hierarchical search}
\label{sec:supp_search_spec}
\label{sec:supp_proof}
\paragraph{Quadtree.} The search starts from a $4\times4$ partition $\mathcal{Q}_0$ of the image. At depth $\ell$, each node has side $L/(4\cdot2^\ell)$, with boundaries rounded as above. Each split halves the side of a node, and the search stops splitting once the node side is at most $w_{\min}=8\,\mathrm{px}$. The maximum depth is therefore $\ell_{\max}=\lceil\log_2(L/(4w_{\min}))\rceil$, with nodes expanded only at depths $\ell=0,\ldots,\ell_{\max}-1$; depending on the image size, nodes at the maximum depth have sides of 4 to 8\,px. At each depth, the relevant nodes $\mathcal{O}_\ell$ are those of \cref{eq:open} of the main paper, with $\tau=0.25$. Before it expands nodes at a depth, the search counts how many groups of four children fit within the forward passes that remain for the search, $N_{\max}-127-N$, and splits at most that many relevant nodes, those with the largest $|\Delta|$; they form $\mathcal{S}_\ell$ (\cref{alg:supp_method}). \Cref{fig:supp_partition} shows the leaves on a Real-IAD~\cite{wang2024realiad} image, a CAMELYON16~\cite{bejnordi2017camelyon} crop and a whole slide.

\begin{suppwide}{figure}
\centering
\includegraphics[width=\textwidth]{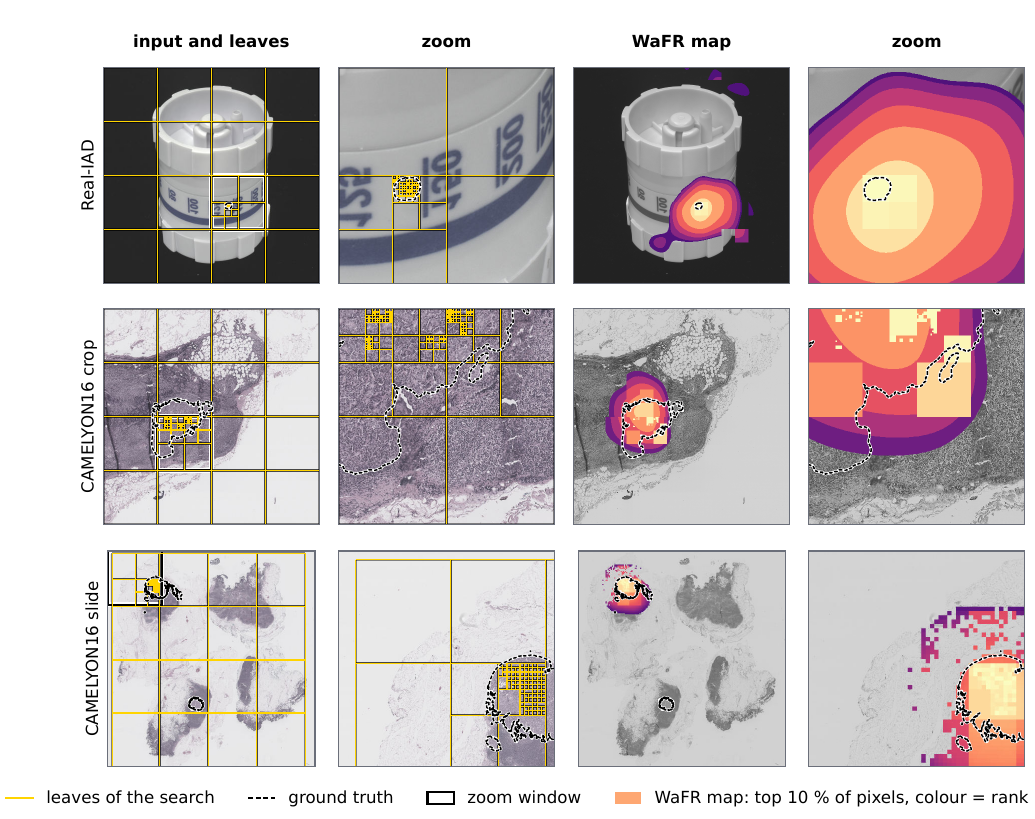}
\caption{\textbf{How \method partitions the image.}}
\label{fig:supp_partition}
\end{suppwide}

\suppneedspace{15\baselineskip}
\paragraph{Number of forward passes.} The following statement counts the forward passes of the search; it is conditional on the number of nodes expanded at each depth.

\begin{proposition}
\label{prop:cost}
Let $\mathcal{S}_\ell\subseteq\mathcal{O}_\ell$ be the relevant nodes split at depth $\ell$ and $k=\max_\ell|\mathcal{S}_\ell|$. The search uses
{\small
\begin{align*}
  N&=1+16+4\sum_{\ell=0}^{\ell_{\max}-1}|\mathcal{S}_\ell|\\
   &\le\min\big(N_{\max}-127,\ 17+4k\ell_{\max}\big)
\end{align*}}%
forward passes. The subsequent sparse regression requires 127 additional forward passes, giving a total of $N+127\leq N_{\max}=2{,}048$ for \method.
\end{proposition}

\begin{proof}
The search evaluates $f(x)$ once and reuses its output to compute each score difference $\Delta(Q)=f(x)-f(x_{\setminus Q})$. Evaluating the 16 nodes of $\mathcal{Q}_0$ requires 16 additional forward passes. Each split node introduces four children, each requiring one forward pass. Since every child has a unique parent, no node is evaluated twice. Summing over the depths gives the equality. Since $|\mathcal{S}_\ell|\le k$ at each of the $\ell_{\max}$ depths, $N\le 17+4k\ell_{\max}$. At each depth, the search splits at most $\lfloor(N_{\max}-127-N)/4\rfloor$ nodes; hence, $N\le N_{\max}-127$. The regression then evaluates 127 random masks, giving $N+127$ forward passes in total.
\end{proof}
The bound depends on $k$, the maximum number of nodes expanded at any depth. When few nodes are expanded per depth, the number of forward passes grows with the depth of the quadtree. For spatially diffuse evidence, more nodes may require refinement, increasing the search cost.

\paragraph{Relevant nodes per depth.}
The equality of \cref{prop:cost} holds on all 1,007 pairs of image and side of VisA~\cite{zou2022spot}, Real-IAD and the CAMELYON16 crops. Real defects and lesions span several nodes at finer depths, so the number $|\mathcal{S}_\ell|$ of nodes expanded at depth $\ell$ increases with depth (\cref{fig:supp_levels}): it stays below 17 on average on VisA and Real-IAD and rises from 3.3 to 27.4 on the 8,192\,px crops. No search reached its limit of $N_{\max}-127=1{,}921$ forward passes (means of 76 to 435; maximum 1,857).

\begin{suppcol}{figure}
\centering
\includegraphics[width=\linewidth]{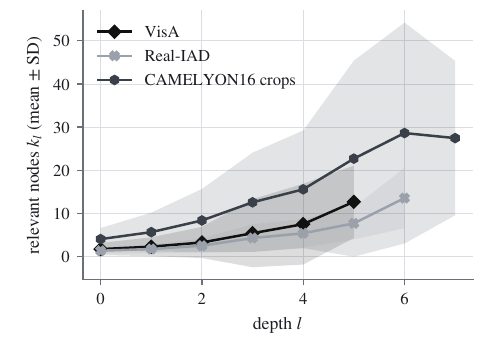}
\caption{\textbf{Relevant nodes per depth.} Shaded: $\pm$SD.}
\label{fig:supp_levels}
\end{suppcol}

\paragraph{Number of targets.}
\label{sec:supp_regime}
The count of \cref{prop:cost} grows with the number of nodes that contain evidence. On new Synthetic Dataset scenes at 1,024, 4,096 and 8,192\,px with $m$ targets, we explain the trained CNN (\cref{sec:supp_setup}) and two synthetic models that combine the scores of the unmasked targets by their mean (additive) or minimum (conjunctive); object recall is the fraction of targets found. For the additive model, the forward passes increase with $m$ (Spearman correlation .95 to .98, object recall $\ge.956$; \cref{sec:results} of the main paper); the conjunctive model behaves likewise, with 591 forward passes and object recall .988 at $m=16$.

\paragraph{Whole slides.}
On whole-slide images, \method operates on the $224$\,px patches used by DSMIL~\cite{li2021dsmil} at $20\times$ magnification. The search is restricted to the bounding box of the tissue mask, skips nodes without tissue, and stops when a node contains a single model patch; a node is split only when its score difference is positive and above the relative threshold. The search also evaluates the fully masked slide, the lower end of the score range of the slide, in addition to $f(x)$, giving two initial evaluations instead of one, so that $N\leq\min(N_{\max},\,18+4k\ell_{\max})$, where $\ell_{\max}$ is the depth needed to reach a single model patch. On whole slides, $N_{\max}=2{,}048$ limits only the search; the regression adds 127 forward passes, giving at most 2,175. Both the search nodes and the $16\times16$ regression regions are aligned with patch boundaries, so each mask either leaves a patch unchanged or replaces it entirely with its Haar low-pass version. The 127 regression masks cover the tissue bounding box, and the patches outside it remain visible. \Cref{sec:supp_cached} describes how a masked slide is evaluated.

\begin{suppcol}{algorithm}
\begin{algorithm}[H]
\footnotesize
\caption{\textbf{\method for one image $x$ and model $f$}, with $\tau=0.25$, $w_{\min}=8$\,px and $N_{\max}=2{,}048$. $N$: forward passes.}
\label{alg:supp_method}
\begin{algorithmic}[1]
\Require image $x$ of side $L$
\State $r\gets L/s_{\mathrm{in}}$; evaluate $f(x)$; $N\gets1$
\If{$r\le1$} \Comment{\cref{sec:refine}}
  \State $y_i\gets f(x_{m_i})$ for 1,023 random masks $m_i$
  \State fit \cref{eq:lasso}; $d$ from the probability of the class
  \State \Return $a$ \Comment{\cref{eq:map}}
\EndIf
\State $\mathcal{Q}_0\gets4\times4$ partition; evaluate $\Delta(Q)$, $Q\in\mathcal{Q}_0$; $N\gets17$; $\ell\gets0$ \\ \Comment{\cref{sec:find}}
\While{the nodes of $\mathcal{Q}_\ell$ are wider than $w_{\min}$}
  \State $\mathcal{O}_\ell\gets$ selected nodes of $\mathcal{Q}_\ell$ \Comment{\cref{eq:open}}
  \State $\mathcal{S}_\ell\gets$ the $\min\big(|\mathcal{O}_\ell|,\lfloor(N_{\max}-127-N)/4\rfloor\big)$ nodes of $\mathcal{O}_\ell$ with the largest $|\Delta|$
  \If{$\mathcal{S}_\ell=\emptyset$} \textbf{break} \EndIf
  \State $\mathcal{Q}_{\ell+1}\gets$ quadrants of $\mathcal{S}_\ell$; evaluate their $\Delta(Q)$; $N\gets N+4|\mathcal{S}_\ell|$; $\ell\gets\ell+1$
\EndWhile
\State $a_{\mathrm{tree}}\gets$ map of the visited nodes \Comment{\cref{eq:pwmap}}
\State $y_i\gets f(x_{m_i})$ for 127 random masks; fit \cref{eq:lasso}; $d$ from the logit; $a$ from \cref{eq:map}
\State \Return leaves smaller than a patch with $a_{\mathrm{tree}}>0$ first, by $a_{\mathrm{tree}}$; then all other pixels by $a$, below zero; $a$ if no such leaf
\end{algorithmic}
\end{algorithm}
\end{suppcol}

\subsection{Sparse regression and final map}
\label{sec:supp_regression}
\label{sec:supp_sign}
\paragraph{Masks and regression.} The random masks~\cite{petsiuk2018rise} are generated from a fixed seed, so the same masks are used for every model. At $r>1$, the 127 masks are the first 127 of the 1,023 masks generated with that seed. The $\ell_1$ regression~\cite{tibshirani1996regression} uses the logit $y_i$ of the explained class; the prediction on the original image is not included in the fit. The Haar basis $\Psi$ on the $16\times16$ patch grid consists of a constant function and 255 horizontal, vertical and diagonal wavelets, defined over dyadic squares of side 16, 8, 4 and 2 patches~\cite{mallat1989theory}. Both $\hat{s}$ and $d$ are upsampled to the image resolution by nearest-neighbour interpolation, after which only $d$ is smoothed. In \cref{eq:map} of the main paper, a term is divided by its maximum absolute value only when this maximum is positive; a term that is zero everywhere stays zero. If the cross-validated regression selects no column, $\hat{s}=0$ and the map is $0.15\,d/\max|d|$ (zero if $d$ is zero as well). With the 127 masks at $r>1$, this happens for 1.2\,\% of the pairs of image and side on VisA, 1.6\,\% on the CAMELYON16 crops and 0.1\,\% of the GridPG~\cite{rao2022towards} targets, and never on Real-IAD and the Synthetic Dataset.

\paragraph{Map composition.}
At $r>1$, the final map combines the leaves of the search with the map $a$ of the 127 regression masks. Pixels of leaves smaller than a regression patch ($L/16$) with $a_{\text{tree}}>0$ keep their tree attribution. All other pixels receive $c(0.999q-1)$, where $q\in(0,1]$ is their rank in $a$ divided by the number of pixels and $c$ is half the smallest retained tree attribution. These values keep the order of $a$ and are negative, below those of the selected leaves. If no leaf qualifies, the final map is $a$. The sign is a design decision: positive attribution is reserved for the regions that the search resolved. Metrics that use only the order of the pixels (AUROC, Pointing Game~\cite{zhang2016excitation}, Deletion, Insertion~\cite{petsiuk2018rise}) do not depend on it, whereas EPG~\cite{wang2020scorecam} and GridPG sum the positive attribution and do. \Cref{tab:supp_sign} gives the map of the paper (T) with two other compositions on the same searches: U1 has the same pixel order as T but keeps the signed regression values outside the selected leaves, and multiplies the tree attribution of the selected leaves by the smallest factor that keeps them above these values; U2 is the regression map alone, without the search. With U1, \method keeps its rank on every benchmark, with EPG .049 (VisA) and .030 (Real-IAD) below T (paired per image, sides averaged); without the search (U2), the scores and ranks fall on every benchmark of \cref{tab:supp_sign}.

\input{tables/S_sign}

\paragraph{Map on whole slides.}
On whole slides, leaves that were not split, have a positive score difference and $a_{\text{tree}}>0$, and are smaller than a regression region in both dimensions keep their tree attribution. All other tissue patches receive $q\epsilon$, where $q\in(0,1]$ is their normalised rank in $a$ and $\epsilon=\min\big(\tfrac12\min a_{\text{tree}},\,0.01\sum a_{\text{tree}}/\sum q\big)$. These values keep the order of $a$ and stay below the selected leaves.

%% file: tables/S_sign.tex
\begin{suppcol}{table}
\centering
\tabstyle
\caption{\textbf{Sign convention of \method maps.} Mean EPG and GridPG ($\uparrow$); in parentheses: rank among 13 perturbation-based methods (Crops 8k, 8,192\,px crops: among 10). T: map of the paper; U1: signed regression values outside the selected leaves; U2: regression map without the search.}
\label{tab:supp_sign}
\footnotesize
\setlength{\tabcolsep}{1.3pt}
\begin{tabular}{@{}l cccc cc@{}}
\toprule
 & \multicolumn{4}{c}{\tabhead{EPG}} & \multicolumn{2}{c}{\tabhead{GridPG}} \\
\cmidrule(lr){2-5}\cmidrule(l){6-7}
\tabhead{Map} & \tabhead{VisA} & \tabhead{Real-IAD} & \tabhead{Crops} & \tabhead{Crops 8k} & \tabhead{ResNet-50} & \tabhead{VGG16} \\
\midrule
T (paper) & .648\,(1) & .895\,(1) & .220\,(2) & .171\,(1) & .915\,(1) & .896\,(1) \\
U1 (signed) & .620\,(1) & .865\,(1) & .217\,(2) & .171\,(1) & .911\,(1) & .892\,(1) \\
U2 (regression) & .187\,(6) & .272\,(5) & .117\,(8) & .045\,(4) & .785\,(5) & .769\,(4) \\
\bottomrule
\end{tabular}
\end{suppcol}

%% file: supplementary/supp_B_roles.tex
\section{Role of each component}
\label{sec:supp_ablation}
\method uses the Haar wavelets in two places and the search in a third. The Haar low-pass baseline $b(x)$ defines what a perturbation removes: the detail finer than $L/16$ inside the masked region. The Haar wavelets of the regression are a prior on the patch effects. The search decides where the resolution is spent. This section follows the pipeline, from the baseline (\cref{sec:supp_lowpass}) to the parameters (\cref{sec:supp_sens}).

\Cref{tab:supp_ablation_full} adds to the ablations of \cref{tab:ablation} of the main paper all intervals and the joint removals of both Haar components (A2, named A3+A4 in \cref{tab:ablation} of the main paper) and of the low-pass baseline and the sparse regression (A10, which combines A3 and A5). It uses 254 to 255 VisA pairs of image and side, 2,680 Synthetic Dataset scenes, the 10 tumour slides of the 20-slide cohort (\method: .947, against .949 on the 45 slides of \cref{tab:supp_slides}) and 1,000 ImageNet~\cite{deng2009imagenet} images per model. \Cref{tab:supp_baseline} repeats the ablations under ROAD~\cite{rong2022road} (\cref{sec:supp_road}) and under log-scale Deletion with blurred replacement; on VisA, ROAD uses the initial segment of the curve and the bootstrap groups the image sides by image. Removing the sparse regression (A5) or the dense term $d$ (A9) raises ROAD on both ImageNet models, and the Deletion advantage of A4 at $r=1$ in \cref{tab:supp_ablation_full} is not reproduced under ROAD (difference within .001).%

\input{tables/S_ablation_full}

\subsection{Low-pass baseline}
\label{sec:supp_lowpass}
\paragraph{Search cost.}
Constant grey (A3) removes the colour of the masked region in addition to its fine detail, so more coarse nodes exceed the relative threshold and the search expands more nodes. Across the 100 VisA images, the Haar low-pass baseline reduces the mean number of search forward passes by 31.5\,\% relative to constant grey (97.7 against 142.7; fewer passes on 78 of 100 images, sign test $p<10^{-7}$). On Real-IAD (35 physical samples, all image sides), the reduction is 17\,\% relative to constant grey and 16\,\% relative to the image mean (100.8 against 121.5 and 120.3 passes; Holm~\cite{holm1979simple} $p<.001$), with a mean EPG and AUROC not lower than with either.%
 With the search limited to 77, 153 or 305 passes, reaching an EPG of .60 on VisA takes on average 270.4 forward passes (search plus 127 masks) with the low-pass baseline, against 305.7 with constant grey and 305.8 with the image mean (Holm $p<.001$; 432 is counted when .60 is not reached). At a search limit of 305 passes, its EPG is higher than with constant grey and with the image mean by .096 (Holm $p\le.002$).

\paragraph{Evidence removal and surrogate fit.}
With 1,023 masks, the Haar low-pass baseline raises the held-out $R^2$ of the regression over constant grey by .071 on ResNet-50~\cite{he2016deep} and .068 on VGG16~\cite{simonyan2015very}, so the model responses are closer to additive when the coarse content is kept. With the low-pass baseline, ROAD is also lower than with constant grey on ResNet-50, VGG16 and VisA (A3 in \cref{tab:supp_baseline}). Against a Gaussian blur tuned per image to the same mean squared difference to the input (mean $\sigma=9.04\,\mathrm{px}$), replacing the annotated object of the 247 ImageNet-S50~\cite{gao2022imagenets} images with the low-pass baseline lowers the target-class logit more, by 1.54 on ResNet-50 and 1.75 on VGG16 (Holm $p=.0004$). Under single-patch masking of $14\times14$\,px patches (244 images with patches inside and outside the object), fewer background patches change the logit more than the median object patch: 16.2\,\% with the low-pass baseline, against 18.5\,\% with the matched blur and 22.4\,\% with constant grey on ResNet-50 (Holm $p=.0004$).

\begin{suppwide}{table}
\centering
\tabstyle
\caption{\textbf{Ablations under ROAD and blur-based log-scale Deletion.} Paired difference ablated $-$ \method ($+$: \method is better), 95\,\% bootstrap CI. Blur log Del.: log-scale Deletion with blurred replacement. Bold: above .01, CI above 0 and Holm $p<.05$; --: not applicable.}
\label{tab:supp_baseline}
\footnotesize
\setlength{\tabcolsep}{3.5pt}
\begin{tabular}{@{}l *{4}{r@{}l}@{}}
\toprule
 & \multicolumn{2}{c}{\tabhead{ResNet-50}} & \multicolumn{2}{c}{\tabhead{VGG16}} & \multicolumn{4}{c}{\tabhead{VisA}} \\
\cmidrule(lr){2-3}\cmidrule(lr){4-5}\cmidrule(l){6-9}
\tabhead{Ablation} & \multicolumn{2}{c}{\tabhead{ROAD $\downarrow$}} & \multicolumn{2}{c}{\tabhead{ROAD $\downarrow$}} & \multicolumn{2}{c}{\tabhead{ROAD $\downarrow$}} & \multicolumn{2}{c}{\tabhead{blur log Del.\ $\downarrow$}} \\
\midrule
\ablid{A3}$-$ Haar low-pass baseline & \textbf{\boldmath$+.017$} & \dci{$+.014$}{$+.020$} & \textbf{\boldmath$+.014$} & \dci{$+.012$}{$+.017$} & \textbf{\boldmath$+.033$} & \dci{$+.014$}{$+.062$} & $+.024$ & \dci{$+.004$}{$+.044$} \\
\ablid{A4}$-$ Haar wavelets & $-.001$ & \dci{$-.003$}{$+.001$} & $-.001$ & \dci{$-.003$}{$+.001$} & $+.000$ & \dci{$-.001$}{$+.002$} & $+.001$ & \dci{$-.003$}{$+.005$} \\
\ablid{A5}$-$ sparse regression & \textbf{\boldmath$+.015$} & \dci{$+.012$}{$+.018$} & \textbf{\boldmath$+.017$} & \dci{$+.014$}{$+.020$} & $+.003$ & \dci{$-.000$}{$+.006$} & $+.010$ & \dci{$-.010$}{$+.031$} \\
\ablid{A6}$-$ random masks & \multicolumn{2}{c}{\na} & \multicolumn{2}{c}{\na} & $+.016$ & \dci{$+.001$}{$+.027$} & $-.014$ & \dci{$-.112$}{$+.066$} \\
\ablid{A7}$-$ probability in $d$ & $+.003$ & \dci{$+.002$}{$+.005$} & $+.002$ & \dci{$+.001$}{$+.003$} & \multicolumn{2}{c}{\na} & \multicolumn{2}{c}{\na} \\
\ablid{A8}$-$ single patches & $-.002$ & \dci{$-.005$}{$+.000$} & $+.000$ & \dci{$-.002$}{$+.002$} & $+.008$ & \dci{$+.004$}{$+.012$} & $+.006$ & \dci{$-.011$}{$+.030$} \\
\ablid{A9}$-$ term $d$ & \textbf{\boldmath$+.020$} & \dci{$+.017$}{$+.023$} & \textbf{\boldmath$+.011$} & \dci{$+.009$}{$+.013$} & $+.004$ & \dci{$+.001$}{$+.009$} & $-.004$ & \dci{$-.019$}{$+.012$} \\
\ablid{A2}$-$ both Haar parts & \textbf{\boldmath$+.019$} & \dci{$+.015$}{$+.022$} & \textbf{\boldmath$+.011$} & \dci{$+.008$}{$+.013$} & \textbf{\boldmath$+.034$} & \dci{$+.016$}{$+.063$} & $+.023$ & \dci{$+.003$}{$+.042$} \\
\ablid{A10}$-$ baseline, regression & \textbf{\boldmath$+.025$} & \dci{$+.021$}{$+.029$} & \textbf{\boldmath$+.028$} & \dci{$+.024$}{$+.031$} & \textbf{\boldmath$+.038$} & \dci{$+.018$}{$+.069$} & $+.027$ & \dci{$-.011$}{$+.061$} \\
\bottomrule
\end{tabular}
\end{suppwide}

\subsection{Hierarchical search}
\label{sec:supp_search}
\paragraph{Search and perturbation.}
\label{sec:supp_design}
Removing the quadtree (A1) lowers EPG on VisA by .422 and the Pointing Game on the Synthetic Dataset by .331 (\cref{tab:supp_ablation_full}). \Cref{tab:supp_factorial} runs Occlusion~\cite{zeiler2014visualizing}, KernelSHAP~\cite{lundberg2017unified}, RISE~\cite{petsiuk2018rise} and HiPe~\cite{cooper2022hipe} with their own baselines and with the Haar low-pass baseline of \method, so that the search and the reference image change separately. The statistical unit is the VisA image, the Real-IAD sample and the crops slide (RISE: 20, 18 and 24 units; HiPe: image sides up to 1,024\,px). The low-pass baseline changes the EPG of these methods by $-.04$ to $+.06$. With the same baseline, \method has the higher EPG in all 12 comparisons of dataset and method (Holm $p<.05$ in 11) and the higher Pointing Game in all 12 (Holm $p\leq.016$ in 9).

\input{tables/S_factorial}

\paragraph{Localisation and number of forward passes.}
\label{sec:supp_budget}
On the 2,048\,px CAMELYON16 crops and Real-IAD images, \cref{fig:supp_budget} runs RISE, HSIC~\cite{novello2022hsic} and KernelSHAP at the per-image number of forward passes of \method and at 512, 1,024 and 2,048 forward passes, and limits the forward passes of \method from 204 to 2,048. RISE runs on 24 crops from 24 slides and on 23 images of 18 samples. The means are grouped over slides or samples, so the EPG of \method on the crops is .236 there, against .200 per crop in \cref{tab:equal} of the main paper. The EPG and Pointing Game curves of the three methods stay nearly flat from 512 to 2,048 forward passes, whereas \method reaches its EPG at the smallest limit (204). By the area under each curve, \method has the higher Pointing Game against all three methods on both datasets (Holm $p\le.012$) and the higher EPG on Real-IAD (Holm $p\le.001$) and against HSIC and KernelSHAP on the crops (Holm $p\le.034$). Without a limit, HiPe uses on average 2,479 and 1,435 forward passes on the crops and Real-IAD, with lower localisation scores than \method (\cref{fig:supp_budget}).

\begin{suppwide}{figure}
\centering
\includegraphics[width=\linewidth]{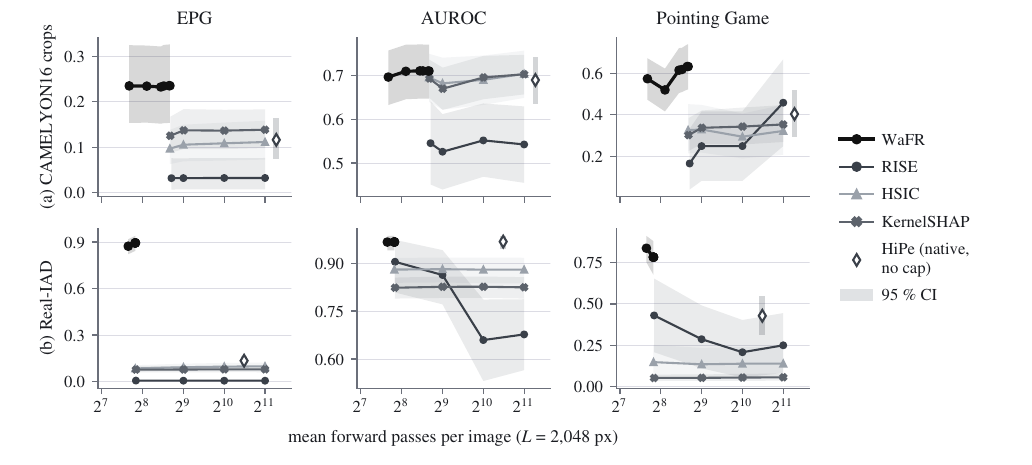}
\caption{\textbf{Localisation against the number of forward passes.} All metrics $\uparrow$; upper left: higher score with fewer forward passes.}
\label{fig:supp_budget}
\end{suppwide}

\paragraph{HiPe without a forward-pass limit.}
\label{sec:supp_hipe}
HiPe is also hierarchical, but splits a region when its attribution exceeds the midpoint between the smallest and largest attribution at its depth (\cref{sec:supp_baselines}). Without the limit of \cref{tab:equal} of the main paper, on all pairs of image and side of \cref{tab:hr}, HiPe uses 2.8 to 4.7 times the median forward passes of \method (per pair, HiPe\,/\,\method: VisA 586\,/\,208, Real-IAD 924.5\,/\,224, crops 1,660.5\,/\,356), has lower EPG and Pointing Game on all three datasets, and matches \method in AUROC (\cref{tab:supp_hipe_hr}).

\input{tables/S_hipe_hr}

On whole slides without the 870,000-pass limit of \cref{sec:results} of the main paper, HiPe finishes all 20 slides (\cref{sec:supp_setup}) with a median of 88,997 forward passes (range 13,881 to 2,397,918) and up to 50,744\,s on a single slide (wall time of the quality run; \cref{tab:supp_hipe_slides}), against a median of 176 for \method. On the 10 tumour slides, the patch-level AUROC is .947 for \method and .870 for HiPe (exact two-sided paired test, $p=.0625$), and the Pointing Game is 1.000 and .900.

\input{tables/S_hipe_slides}

\paragraph{Several tumour foci.}
41 of the 45 CAMELYON16 tumour slides of \cref{tab:supp_slides} contain several annotated tumour foci. On these slides, the 5\,\% highest-ranked tissue patches of \method reach .517 of the foci on average (fraction of foci containing at least one of them), more than for every perturbation-based method (.139 to .392; Holm $p\le.0093$).

\paragraph{Quadtree-boundary alignment.}
A target on the boundary between nodes is split among them, so each holds part of it. On Synthetic Dataset scenes at 2,048\,px for all 8 models, one target is shifted by 0, 1/8, 1/4 or 1/2 of its side from the corner of four nodes, a node boundary at every depth. At every offset, \method has a Pointing Game of 1.000 with 201 to 270 forward passes, and AUROC of 1.000, .951, .900 and .995, against .885, .782, .693 and .522 for h-Shap~\cite{teneggi2022hshap}.

\subsection{Haar prior in the regression}
\label{sec:supp_haar_role}
At $r=1$ on the 1,000 ImageNet images, the Haar wavelets of the regression act as a multiscale prior on the patch effects (\cref{tab:supp_haar_prior,fig:supp_masks}). At $r>1$, removing them (A4) only reorders the pixels outside the refined leaves, so EPG and the Pointing Game in \cref{tab:supp_ablation_full} are equal by construction.%

\input{tables/S_haar_prior}

\paragraph{Sample efficiency and stability.}
With 127 random masks, adding the Haar wavelets to the patch columns (\method against A4) raises Insertion by .023 on ResNet-50 and .030 on VGG16, and the mean Spearman correlation between the maps of five seeds by .090 and .079 (\cref{tab:supp_haar_prior}; Holm $p=.0006$ for every metric of the table). With 1,023 masks, the gain in correlation across seeds is .081 and .054. On the first 500 images, with mask counts on a logarithmic grid from 16 to 1,023, \method reaches the 127-mask Insertion of A4 with 51 masks on ResNet-50 and 46 on VGG16, 2.48 and 2.74 times fewer.

\begin{suppcol}{figure}
\centering
\includegraphics[width=\linewidth]{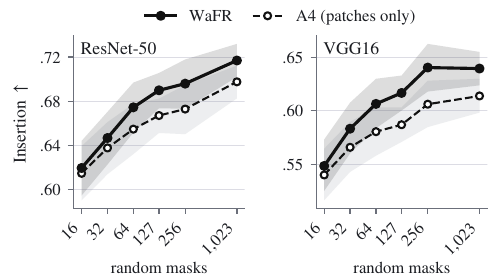}
\caption{\textbf{Insertion against the number of random masks ($r=1$).} Shaded: 95\,\% CI.}
\label{fig:supp_masks}
\end{suppcol}

\paragraph{Number of masks.}
\label{sec:supp_masks}
With the same number of masks, the Haar wavelets give the higher Insertion at every count from 32 to 1,023 on both models (\cref{fig:supp_masks}; Holm $p\leq.0006$). Refitting \method and the ablations A3, A4 and A3+A4 on the first 64 to 1,023 masks, \method with 64 masks reaches the mean Insertion of A3+A4 with 1,023 masks (A4 needs 256 masks to reach it).

\paragraph{Spatial layout of the wavelets.}
A fixed random permutation of the 256 patch indices inside the Haar basis (Perm.\ in \cref{tab:supp_haar_prior}) keeps the number and the norms of the columns and removes their spatial adjacency. With 127 masks, the true layout has the higher Insertion (by .007 on ResNet-50 and .009 on VGG16) and correlation across seeds (by .044 on both; Holm $p=.0006$). At 32, 64, 127 and 256 masks, the true layout is higher in all 16 combinations of model and count (Holm $p\leq.020$).

\paragraph{Both Haar parts.}
\label{sec:supp_two}
The low-pass baseline changes the model responses, and the Haar wavelets regularise the regression. At $r=1$, removing both (A3+A4) lowers Insertion by more than the sum of the two single removals: with $\Delta_k$ the paired advantage of \method over ablation $k$, $I=\Delta_{34}-(\Delta_{3}+\Delta_{4})$ is $+.011$ on ResNet-50 and $+.017$ on VGG16 (Holm $p<.001$). At high resolution, the contribution of the Haar wavelets therefore lies in the low-pass baseline, which lowers the cost of the search; in the regression, they act at $r=1$, where \method reaches the 127-mask Insertion of A4 with 2.5 to 2.7 times fewer masks.

\subsection{Regression after the search}
\label{sec:supp_a6}
\paragraph{Ranking outside the selected leaves.}
At $r>1$, the search resolves the fine evidence, and the 127 random masks order the pixels outside the selected leaves. Without the regression (A6), the search tree is the same and the other pixels are ranked by the per-pixel score difference $\Delta(Q)/|Q|$ of their leaf. The Pointing Game is then unchanged on all 36 pairs of model and side of the Synthetic Dataset, and EPG changes by at most .004 on VisA, Real-IAD and the CAMELYON16 crops. With the regression, AUROC is higher on every high-resolution dataset: on VisA and on whole slides (\cref{tab:supp_ablation_full}; patch-level AUROC from .906 to .947, higher on 9 of the 10 tumour slides), by .075 on Real-IAD, and by .055 to .103 across image sides on the CAMELYON16 crops. On VisA, the lower log-scale Deletion of A6 in \cref{tab:supp_ablation_full} is reversed under ROAD, where \method is lower by .016 (\cref{tab:supp_baseline}).%

\paragraph{Number of masks at $r>1$.}
Beyond 63 masks, the metrics are stable on all datasets (\cref{fig:supp_sens}); with no random masks, AUROC is lower by .083 on VisA, .025 on the Synthetic Dataset at 512\,px and .032 on whole slides.

\paragraph{$\mu$Fidelity at $r=1$.}
Among all ablations, $\mu$Fidelity~\cite{bhatt2020evaluating} decreases most without the sparse regression (A5), without the low-pass baseline and the regression (A3+A5), and without the dense term $d$ (A9) (\cref{tab:supp_ablation_mufid}).

\input{tables/S_ablation_mufid}

\subsection{Regime rule}
\label{sec:supp_vit}
\label{sec:supp_vit_switch}
The rule $r=L/s_{\text{in}}>1$ (\cref{sec:rule} of the main paper) runs the search when the image is larger than the input of the model. \Cref{tab:supp_switch} runs each regime where the rule does not select it: the sparse regression alone (1,024 forward passes) on GridPG ($r=2$, 100 grids per CNN) and on 43 VisA images at 512\,px ($r>1$), and the search followed by 127 masks on 100 ImageNet images at 224\,px ($r=1$, ResNet-50). At $r>1$, the search raises the GridPG share on both CNNs and the EPG on VisA with 2.7 and 5.2 times fewer forward passes. At 224\,px, the regression alone has the better Deletion, Insertion and $\mu$Fidelity.

\begin{suppcol}{table}
\centering
\tabstyle\renewcommand{\arraystretch}{1}
\caption{\textbf{Search and regression across regimes.} Search: hierarchical search plus 127 masks; Regr.: sparse regression alone, 1,024 forward passes; Fwd.: mean forward passes of the search. $\Delta$: Search $-$ Regr.\ ($+$ favours the search; sign inverted for Deletion) with 95\,\% bootstrap CI. Del.: Deletion; log Del.: log-scale Deletion; Ins.: Insertion; $\mu$Fid.: $\mu$Fidelity. ImageNet: 100 other validation images (\cref{tab:supp_cohorts}).}
\label{tab:supp_switch}
\footnotesize
\setlength{\tabcolsep}{2.4pt}
\begin{tabular}{@{}l l r rr r@{}l@{}}
\toprule
 & & & \multicolumn{2}{c}{\tabhead{Score}} & & \\
\cmidrule(lr){4-5}
\tabhead{Model} & \tabhead{Metric} & \tabhead{Fwd.} & \tabhead{Search} & \tabhead{Regr.} & \multicolumn{2}{c}{\tabhead{$\Delta$ [95\,\% CI]}} \\
\midrule
\multicolumn{7}{@{}l}{\tabheadfont GridPG, 448\,px ($r=2$)}\\[1pt]
ResNet-50 & GridPG $\uparrow$ & 373 & .889 & .802 & $+.087$ & \dci{$+.041$}{$+.128$} \\
 & AUROC $\uparrow$ & & .935 & .924 & $+.011$ & \dci{$+.002$}{$+.020$} \\
VGG16 & GridPG $\uparrow$ & 376 & .905 & .814 & $+.091$ & \dci{$+.050$}{$+.130$} \\
 & AUROC $\uparrow$ & & .914 & .916 & $-.002$ & \dci{$-.012$}{$+.008$} \\
\midrule
\multicolumn{7}{@{}l}{\tabheadfont VisA, 512\,px ($r>1$)}\\[1pt]
ResNet-18 & EPG $\uparrow$ & 198 & .664 & .408 & $+.255$ & \dci{$+.187$}{$+.325$} \\
 & AUROC $\uparrow$ & & .897 & .903 & $-.006$ & \dci{$-.026$}{$+.020$} \\
 & log Del.\,$\downarrow$ & & .811 & .837 & $+.026$ & \dci{$-.017$}{$+.062$} \\
\midrule
\multicolumn{7}{@{}l}{\tabheadfont ImageNet, 224\,px ($r=1$)}\\[1pt]
ResNet-50 & Del.\,$\downarrow$ & 306 & .056 & .040 & $-.017$ & \dci{$-.025$}{$-.009$} \\
 & Ins.\,$\uparrow$ & & .554 & .604 & $-.050$ & \dci{$-.068$}{$-.033$} \\
 & $\mu$Fid.\,$\uparrow$ & & .295 & .532 & $-.237$ & \dci{$-.279$}{$-.198$} \\
\bottomrule
\end{tabular}
\end{suppcol}

\subsection{Parameters}
\label{sec:supp_sens}
We change one parameter at a time around the values of \cref{tab:supp_params} (\cref{fig:supp_sens}), on the Synthetic Dataset (8 pairs of model and target size, 20 scenes per side up to 2,048\,px; 5 scenes of two models at 8,192\,px), 30 VisA images, 5 CAMELYON16 training tumour slides ($\tau$ and number of masks) and 100 ImageNet images ($r=1$; dense and patch weights). The maximal number of forward passes $N_{\max}$ (512 to 4,096) and mask counts above 63 leave all metrics nearly unchanged. The relative threshold $\tau$ trades forward passes against log-scale Deletion without changing the Pointing Game on the Synthetic Dataset: $\tau=.50$ and $.75$ use slightly fewer forward passes with higher log-scale Deletion ($+.013$ and $+.024$ at 2,048\,px), and $\tau=.10$ uses more (278 against 220 on the Synthetic Dataset at 2,048\,px, 329 against 236 on VisA) without lower Deletion. The finest node side $w_{\min}$ sets the spatial granularity: nodes of 16 and 32\,px lower EPG and Pointing Game, and nodes of 4\,px raise EPG by .045 on the Synthetic Dataset and .059 on VisA with about 25\,\% more forward passes. On ImageNet, removing the dense term lowers Insertion by .046 and $\mu$Fidelity by .030, weights above 0.15 trade higher Insertion for higher Deletion, and a patch weight of 4 instead of 2 lowers Insertion by .012.

\begin{suppwide}{figure}
\centering
\includegraphics[width=\linewidth,height=.86\textheight,keepaspectratio]{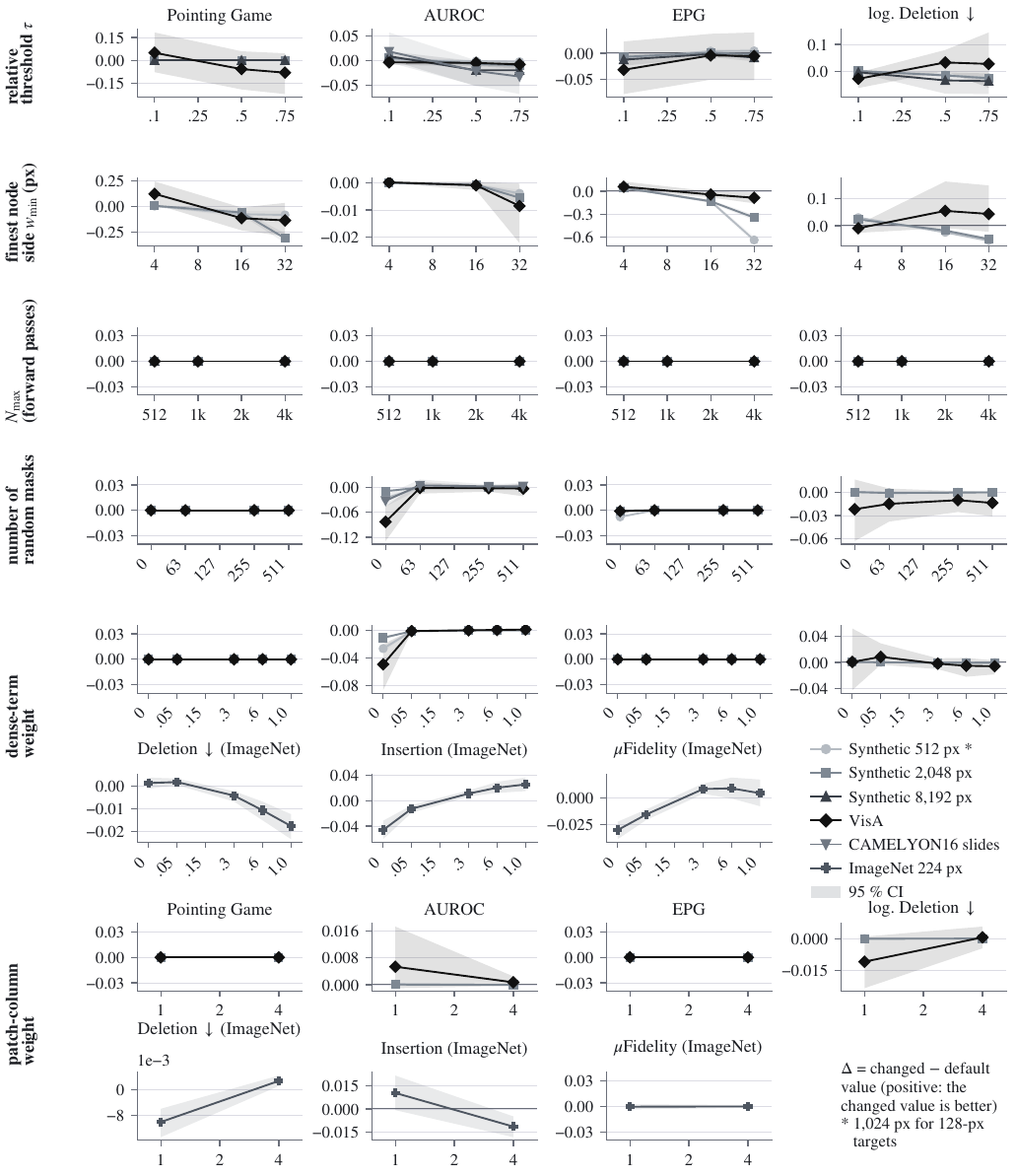}
\caption{\textbf{Sensitivity of \method to its parameters.} One row per parameter, one panel per metric; horizontal axis: changed value (the value of \cref{tab:supp_params} is at zero). Shaded: 95\,\% bootstrap CI.}
\label{fig:supp_sens}
\end{suppwide}

%% file: tables/S_ablation_full.tex
\begin{suppwide}{table}
\centering
\tabstyle
\caption{\textbf{Ablations, with all intervals.} \method: mean. Other rows: paired difference, \method $-$ ablated for $\uparrow$ and ablated $-$ \method for $\downarrow$ ($+$: \method is better), 95\,\% bootstrap CI. Bold: above the margin (.02 for AUROC, EPG and Pointing Game; .01 otherwise) with the CI above 0; $^{\circ}$: the ablated method is better by more than the margin, CI below 0. --: A3 on slides, identical maps due to tissue filtering.}
\label{tab:supp_ablation_full}
\footnotesize
\setlength{\tabcolsep}{3.5pt}
\begin{suppfit}
\begin{tabular}{@{}l *{5}{r@{}l}@{}}
\toprule
 & \multicolumn{6}{c}{\tabhead{VisA}} & \multicolumn{2}{c}{\tabhead{Synthetic}} & \multicolumn{2}{c}{\tabhead{CAMELYON16 slides}} \\
\cmidrule(lr){2-7}\cmidrule(lr){8-9}\cmidrule(lr){10-11}
\tabhead{$r>1$} & \multicolumn{2}{c}{\tabhead{AUROC $\uparrow$}} & \multicolumn{2}{c}{\tabhead{EPG $\uparrow$}} & \multicolumn{2}{c}{\tabhead{log-scale Deletion $\downarrow$}} & \multicolumn{2}{c}{\tabhead{Pointing Game $\uparrow$}} & \multicolumn{2}{c}{\tabhead{AUROC $\uparrow$}} \\
\midrule
\method & .916 &  & .668 &  & .785 &  & .997 &  & .947 &  \\
\treemid \ablid{A1}$-$ quadtree & $+.015$ & \dci{$+.000$}{$+.031$} & \textbf{\boldmath$+.422$} & \dci{$+.367$}{$+.475$} & $+.021$ & \dci{$-.026$}{$+.053$} & \textbf{\boldmath$+.331$} & \dci{$+.313$}{$+.349$} & $+.009$ & \dci{$-.008$}{$+.028$} \\
\treemid \ablid{A3}$-$ Haar low-pass baseline & $+.001$ & \dci{$-.014$}{$+.014$} & \textbf{\boldmath$+.093$} & \dci{$+.046$}{$+.146$} & $-.042$\hurt & \dci{$-.097$}{$-.009$} & $+.000$ & \dci{$-.000$}{$+.001$} & \multicolumn{2}{c}{--} \\
\treemid \ablid{A4}$-$ Haar wavelets & $+.001$ & \dci{$-.001$}{$+.003$} & $+.000$ & \dci{$+.000$}{$+.000$} & $-.001$ & \dci{$-.002$}{$+.001$} & $+.000$ & \dci{$+.000$}{$+.000$} & $+.001$ & \dci{$-.001$}{$+.003$} \\
\treelast \ablid{A6}$-$ random masks & \textbf{\boldmath$+.120$} & \dci{$+.087$}{$+.149$} & $+.004$ & \dci{$-.001$}{$+.008$} & $-.033$\hurt & \dci{$-.062$}{$-.008$} & $+.000$ & \dci{$+.000$}{$+.000$} & \textbf{\boldmath$+.041$} & \dci{$+.009$}{$+.082$} \\
\midrule
 & \multicolumn{4}{c}{\tabhead{ImageNet, ResNet-50}} & \multicolumn{4}{c}{\tabhead{ImageNet, VGG16}} & & \\
\cmidrule(lr){2-5}\cmidrule(lr){6-9}
\tabhead{$r=1$} & \multicolumn{2}{c}{\tabhead{Deletion $\downarrow$}} & \multicolumn{2}{c}{\tabhead{Insertion $\uparrow$}} & \multicolumn{2}{c}{\tabhead{Deletion $\downarrow$}} & \multicolumn{2}{c}{\tabhead{Insertion $\uparrow$}} & & \\
\midrule
\method & .069 &  & .717 &  & .049 &  & .640 &  & & \\
\treemid \ablid{A3}$-$ Haar low-pass baseline & $+.005$ & \dci{$+.003$}{$+.007$} & \textbf{\boldmath$+.015$} & \dci{$+.011$}{$+.019$} & $+.006$ & \dci{$+.004$}{$+.008$} & $+.008$ & \dci{$+.004$}{$+.012$} & & \\
\treemid \ablid{A4}$-$ Haar wavelets & $-.012$\hurt & \dci{$-.015$}{$-.010$} & \textbf{\boldmath$+.019$} & \dci{$+.016$}{$+.022$} & $-.005$ & \dci{$-.006$}{$-.004$} & \textbf{\boldmath$+.026$} & \dci{$+.023$}{$+.029$} & & \\
\treemid \ablid{A2}$-$ both Haar parts & $-.012$\hurt & \dci{$-.015$}{$-.009$} & \textbf{\boldmath$+.045$} & \dci{$+.040$}{$+.049$} & $-.004$ & \dci{$-.006$}{$-.003$} & \textbf{\boldmath$+.051$} & \dci{$+.046$}{$+.055$} & & \\
\treemid \ablid{A10}$-$ baseline, regression & \textbf{\boldmath$+.049$} & \dci{$+.044$}{$+.054$} & $-.002$ & \dci{$-.007$}{$+.004$} & \textbf{\boldmath$+.053$} & \dci{$+.049$}{$+.057$} & $-.017$\hurt & \dci{$-.024$}{$-.011$} & & \\
\treemid \ablid{A5}$-$ sparse regression & \textbf{\boldmath$+.048$} & \dci{$+.043$}{$+.053$} & $-.008$ & \dci{$-.013$}{$-.003$} & \textbf{\boldmath$+.048$} & \dci{$+.044$}{$+.052$} & $-.032$\hurt & \dci{$-.038$}{$-.026$} & & \\
\treemid \ablid{A7}$-$ probability in $d$ & $+.003$ & \dci{$+.002$}{$+.004$} & \textbf{\boldmath$+.012$} & \dci{$+.010$}{$+.014$} & $+.002$ & \dci{$+.001$}{$+.003$} & \textbf{\boldmath$+.010$} & \dci{$+.008$}{$+.012$} & & \\
\treemid \ablid{A8}$-$ single patches & \textbf{\boldmath$+.010$} & \dci{$+.008$}{$+.013$} & $-.007$ & \dci{$-.011$}{$-.003$} & \textbf{\boldmath$+.012$} & \dci{$+.010$}{$+.013$} & $-.005$ & \dci{$-.008$}{$-.001$} & & \\
\treelast \ablid{A9}$-$ term $d$ & $-.003$ & \dci{$-.005$}{$-.001$} & \textbf{\boldmath$+.028$} & \dci{$+.025$}{$+.032$} & $-.003$ & \dci{$-.004$}{$-.002$} & \textbf{\boldmath$+.044$} & \dci{$+.040$}{$+.047$} & & \\
\bottomrule
\end{tabular}
\end{suppfit}
\end{suppwide}

%% file: tables/S_factorial.tex
\begin{suppwide}{table}
\centering
\tabstyle
\caption{\textbf{Search and perturbation reference.} B: baseline method. Search: \method $-$ B with the Haar low-pass baseline; Value: B with the low-pass baseline $-$ B with its own; $+$ favours \method or the low-pass baseline. \method: mean over the same units. Bold: Holm $p<.05$ within each dataset and metric.}
\label{tab:supp_factorial}
\footnotesize
\setlength{\tabcolsep}{4pt}
\begin{tabular}{@{}l ccc ccc ccc@{}}
\toprule
 & \multicolumn{3}{c}{\tabhead{VisA}} & \multicolumn{3}{c}{\tabhead{Real-IAD}} & \multicolumn{3}{c}{\tabhead{Crops}} \\
\cmidrule(lr){2-4}\cmidrule(lr){5-7}\cmidrule(l){8-10}
\tabhead{Baseline B} & \tabhead{\method} & \tabhead{Search} & \tabhead{Value} & \tabhead{\method} & \tabhead{Search} & \tabhead{Value} & \tabhead{\method} & \tabhead{Search} & \tabhead{Value} \\
\midrule
\tabgroup{10}{EPG $\uparrow$}
Occlusion & .665 & \textbf{\boldmath$+.374$} & $-.001$ & .910 & \textbf{\boldmath$+.562$} & \textbf{\boldmath$+.013$} & .297 & \textbf{\boldmath$+.110$} & $+.003$ \\
KernelSHAP & .692 & \textbf{\boldmath$+.502$} & \textbf{\boldmath$+.056$} & .910 & \textbf{\boldmath$+.661$} & \textbf{\boldmath$-.039$} & .297 & \textbf{\boldmath$+.059$} & $-.009$ \\
RISE & .617 & \textbf{\boldmath$+.579$} & .000 & .873 & \textbf{\boldmath$+.835$} & .000 & .236 & $+.164$ & $-.001$ \\
HiPe & .669 & \textbf{\boldmath$+.411$} & \textbf{\boldmath$+.010$} & .919 & \textbf{\boldmath$+.537$} & \textbf{\boldmath$+.005$} & .313 & \textbf{\boldmath$+.083$} & $+.001$ \\
\tabgroup{10}{AUROC $\uparrow$}
Occlusion & .908 & \textbf{\boldmath$+.184$} & \textbf{\boldmath$-.035$} & .966 & \textbf{\boldmath$+.152$} & $-.020$ & .708 & \textbf{\boldmath$+.162$} & $+.012$ \\
KernelSHAP & .910 & \textbf{\boldmath$+.116$} & \textbf{\boldmath$+.104$} & .966 & \textbf{\boldmath$+.089$} & $+.004$ & .709 & $+.008$ & $+.003$ \\
RISE & .923 & $+.088$ & $-.064$ & .942 & $+.001$ & $-.021$ & .688 & $+.063$ & $+.004$ \\
HiPe & .908 & $+.010$ & $-.004$ & .965 & $+.012$ & $-.004$ & .700 & $+.011$ & $-.007$ \\
\tabgroup{10}{Pointing Game $\uparrow$}
Occlusion & .676 & \textbf{\boldmath$+.493$} & $-.009$ & .837 & \textbf{\boldmath$+.595$} & $+.012$ & .614 & \textbf{\boldmath$+.249$} & $+.017$ \\
KernelSHAP & .676 & \textbf{\boldmath$+.492$} & \textbf{\boldmath$+.048$} & .837 & \textbf{\boldmath$+.621$} & $-.028$ & .615 & \textbf{\boldmath$+.181$} & $-.032$ \\
RISE & .618 & $+.176$ & $+.025$ & .770 & $+.003$ & $+.059$ & .614 & \textbf{\boldmath$+.295$} & $-.118$ \\
HiPe & .676 & \textbf{\boldmath$+.144$} & $+.024$ & .864 & \textbf{\boldmath$+.155$} & $-.024$ & .608 & $+.058$ & $-.043$ \\
\bottomrule
\end{tabular}
\end{suppwide}

%% file: tables/S_hipe_hr.tex
\begin{suppwide}{table}
\centering
\tabstyle
\caption{\textbf{HiPe without a forward-pass limit on the high-resolution images.} \method, HiPe: mean over image sides. Difference: \method $-$ HiPe, paired over units ($+$: \method is better; log-scale Deletion: HiPe $-$ \method) with 95\,\% bootstrap CI; $p$: two-sided paired test. Grouped unit: VisA image, Real-IAD sample, crops slide.}
\label{tab:supp_hipe_hr}
\footnotesize
\setlength{\tabcolsep}{4pt}
\begin{tabular}{@{}l cc r@{}l c r@{}l c@{}}
\toprule
 & & & \multicolumn{3}{c}{\tabhead{Image as unit}} & \multicolumn{3}{c}{\tabhead{Grouped unit}} \\
\cmidrule(lr){4-6}\cmidrule(l){7-9}
\tabhead{Metric} & \tabhead{\method} & \tabhead{HiPe} & \multicolumn{2}{c}{\tabhead{Difference [95\,\% CI]}} & \tabhead{$p$} & \multicolumn{2}{c}{\tabhead{Difference [95\,\% CI]}} & \tabhead{$p$} \\
\midrule
\tabgroup{9}{VisA}
EPG $\uparrow$ & .648 & .212 & $+.421$ & \dci{$+.364$}{$+.476$} & $<$.001 & $+.421$ & \dci{$+.364$}{$+.476$} & $<$.001 \\
Pointing Game $\uparrow$ & .660 & .456 & $+.168$ & \dci{$+.111$}{$+.227$} & $<$.001 & $+.168$ & \dci{$+.111$}{$+.227$} & $<$.001 \\
AUROC $\uparrow$ & .919 & .904 & $+.006$ & \dci{$-.010$}{$+.021$} & .44 & $+.006$ & \dci{$-.010$}{$+.021$} & .44 \\
log-scale Deletion $\downarrow$ & .766 & .747 & $-.027$ & \dci{$-.071$}{$+.009$} & .21 & $-.027$ & \dci{$-.071$}{$+.009$} & .21 \\
\tabgroup{9}{Real-IAD}
EPG $\uparrow$ & .895 & .273 & $+.622$ & \dci{$+.579$}{$+.662$} & $<$.001 & $+.665$ & \dci{$+.614$}{$+.713$} & $<$.001 \\
Pointing Game $\uparrow$ & .821 & .619 & $+.201$ & \dci{$+.141$}{$+.262$} & $<$.001 & $+.205$ & \dci{$+.110$}{$+.295$} & $<$.001 \\
AUROC $\uparrow$ & .957 & .953 & $+.004$ & \dci{$-.003$}{$+.010$} & .29 & $+.002$ & \dci{$-.004$}{$+.008$} & .52 \\
log-scale Deletion $\downarrow$ & .737 & .737 & .000 & \dci{$-.017$}{$+.016$} & .98 & $+.001$ & \dci{$-.022$}{$+.022$} & .93 \\
\tabgroup{9}{Crops}
EPG $\uparrow$ & .220 & .129 & $+.094$ & \dci{$+.061$}{$+.129$} & $<$.001 & $+.105$ & \dci{$+.069$}{$+.145$} & $<$.001 \\
Pointing Game $\uparrow$ & .584 & .442 & $+.135$ & \dci{$+.074$}{$+.198$} & $<$.001 & $+.116$ & \dci{$+.048$}{$+.186$} & .002 \\
AUROC $\uparrow$ & .684 & .688 & $-.009$ & \dci{$-.036$}{$+.018$} & .51 & $-.003$ & \dci{$-.032$}{$+.026$} & .87 \\
log-scale Deletion $\downarrow$ & .774 & .806 & $+.029$ & \dci{$+.015$}{$+.042$} & $<$.001 & $+.033$ & \dci{$+.019$}{$+.046$} & $<$.001 \\
\bottomrule
\end{tabular}
\end{suppwide}

%% file: tables/S_hipe_slides.tex
\begin{suppcol}{table}
\centering
\tabstyle
\caption{\textbf{HiPe on whole slides, per slide.} PG: Pointing Game; Deletion: slide Deletion over 0 to 100\,\% of the tissue removed. --: slide without tumour (no ground truth).}
\label{tab:supp_hipe_slides}
\footnotesize
\setlength{\tabcolsep}{2pt}
\begin{tabular}{@{}l c r r ccc@{}}
\toprule
\tabhead{Slide} & \tabhead{Tumour} & \tabhead{\shortstack{Forward\\passes}} & \tabhead{\shortstack{Wall\\time (s)}} & \tabhead{\shortstack{AUROC\\$\uparrow$}} & \tabhead{PG $\uparrow$} & \tabhead{\shortstack{Deletion\\$\downarrow$}} \\
\midrule
test\_122 & yes & 62,615 & 3,159 & .929 & 1.000 & .930 \\
test\_121 & yes & 110,959 & 3,039 & .754 & 1.000 & .960 \\
test\_094 & yes & 265,729 & 6,022 & .987 & 1.000 & .946 \\
test\_008 & yes & 242,599 & 9,603 & .998 & 1.000 & .913 \\
test\_074 & yes & 67,035 & 4,396 & .963 & .000 & .958 \\
test\_048 & yes & 46,837 & 1,581 & .885 & 1.000 & .950 \\
test\_066 & yes & 23,323 & 1,669 & .878 & 1.000 & .938 \\
test\_016 & yes & 2,397,918 & 50,744 & .758 & 1.000 & .855 \\
test\_092 & yes & 702,419 & 20,368 & .996 & 1.000 & .921 \\
test\_090 & yes & 1,653,395 & 40,319 & .547 & 1.000 & .965 \\
test\_119 & no & 13,881 & 1,821 & \na & \na & .956 \\
test\_124 & no & 16,538 & 2,038 & \na & \na & .957 \\
test\_115 & no & 25,453 & 2,219 & \na & \na & .958 \\
test\_123 & no & 1,721,159 & 40,524 & \na & \na & .961 \\
test\_070 & no & 20,585 & 2,123 & \na & \na & .961 \\
test\_098 & no & 19,521 & 1,378 & \na & \na & .960 \\
test\_109 & no & 1,426,547 & 27,792 & \na & \na & .964 \\
test\_076 & no & 45,275 & 2,890 & \na & \na & .944 \\
test\_080 & no & 183,076 & 6,499 & \na & \na & .852 \\
test\_037 & no & 492,679 & 17,018 & \na & \na & .965 \\
\bottomrule
\end{tabular}
\end{suppcol}

%% file: tables/S_haar_prior.tex
\begin{suppcol}{table}
\centering
\tabstyle
\caption{\textbf{Haar wavelets with 127 random masks at $r=1$.} A4: \method without the Haar wavelets; Perm.: \method with the patches permuted inside the Haar wavelets. Seeds: mean Spearman correlation between the maps of two seeds. Masks to reach A4: random masks needed to reach the Insertion of A4 with 127 masks.}
\label{tab:supp_haar_prior}
\footnotesize
\setlength{\tabcolsep}{3pt}
\begin{tabular}{@{}l ccc ccc@{}}
\toprule
 & \multicolumn{3}{c}{\tabhead{ResNet-50}} & \multicolumn{3}{c}{\tabhead{VGG16}} \\
\cmidrule(lr){2-4}\cmidrule(l){5-7}
 & \tabhead{\method} & \tabhead{A4} & \tabhead{Perm.} & \tabhead{\method} & \tabhead{A4} & \tabhead{Perm.} \\
\midrule
Insertion $\uparrow$ & .690 & .667 & .683 & .617 & .587 & .607 \\
$\mu$Fidelity $\uparrow$ & .468 & .458 & .464 & .497 & .491 & .494 \\
Held-out $R^2$ $\uparrow$ & .612 & .580 & .604 & .627 & .602 & .620 \\
Seeds $\uparrow$ & .530 & .440 & .486 & .524 & .445 & .480 \\
\addlinespace[2pt]
Masks to reach A4 $\downarrow$ & 51 & 127 & 62 & 46 & 127 & 64 \\
\bottomrule
\end{tabular}
\end{suppcol}

%% file: tables/S_ablation_mufid.tex
\begin{suppcol}{table}
\centering
\tabstyle
\caption{\textbf{Ablations at $r=1$: $\mu$Fidelity.} \method: mean. Other rows: paired difference \method $-$ ablated ($+$: \method is better) with 95\,\% bootstrap CI. Bold: difference above .02 and CI above 0.}
\label{tab:supp_ablation_mufid}
\footnotesize
\setlength{\tabcolsep}{0.4pt}\renewcommand{\dci}[2]{\kern2pt{\color{tabmute}\fontsize{5.6}{6}\selectfont\raisebox{-.25ex}{[#1,\,#2]}}}%
\begin{suppfit}
\begin{tabular}{@{}l@{\hspace{4pt}} r@{}l@{\hspace{3pt}} r@{}l@{}}
\toprule
\tabhead{$\mu$Fidelity $\uparrow$, $r=1$} & \multicolumn{2}{c}{\tabhead{ResNet-50}} & \multicolumn{2}{c}{\tabhead{VGG16}} \\
\midrule
\method & .537 &  & .570 &  \\
\treemid \ablid{A3}$-$ Haar baseline & $-.004$ & \dci{$-.008$}{$-.000$} & $+.000$ & \dci{$-.003$}{$+.003$} \\
\treemid \ablid{A4}$-$ Haar wavelets & $+.003$ & \dci{$+.002$}{$+.004$} & $+.001$ & \dci{$+.001$}{$+.002$} \\
\treemid \ablid{A2}$-$ both Haar parts & $-.001$ & \dci{$-.004$}{$+.003$} & $+.003$ & \dci{$-.000$}{$+.007$} \\
\treemid \ablid{A10}$-$ baseline, regression & \textbf{\boldmath$+.024$} & \dci{$+.018$}{$+.030$} & \textbf{\boldmath$+.031$} & \dci{$+.025$}{$+.036$} \\
\treemid \ablid{A5}$-$ sparse regression & \textbf{\boldmath$+.026$} & \dci{$+.021$}{$+.031$} & \textbf{\boldmath$+.032$} & \dci{$+.027$}{$+.037$} \\
\treemid \ablid{A7}$-$ probability in $d$ & $+.014$ & \dci{$+.012$}{$+.016$} & $+.014$ & \dci{$+.013$}{$+.016$} \\
\treemid \ablid{A8}$-$ single patches & $+.007$ & \dci{$+.006$}{$+.009$} & $+.005$ & \dci{$+.004$}{$+.007$} \\
\treelast \ablid{A9}$-$ term $d$ & \textbf{\boldmath$+.024$} & \dci{$+.022$}{$+.026$} & \textbf{\boldmath$+.024$} & \dci{$+.023$}{$+.026$} \\
\bottomrule
\end{tabular}
\end{suppfit}
\end{suppcol}

%% file: supplementary/supp_C_protocol.tex
\section{Evaluation protocol}
\label{sec:supp_protocol}
This section gives the datasets and models (\cref{sec:supp_setup}), the baselines (\cref{sec:supp_baselines}), the metrics (\cref{sec:supp_metrics}) and the statistics and cost (\cref{sec:supp_stats}).

\subsection{Datasets and models}
\label{sec:supp_setup}

\begin{suppwide}{table}
  \centering
  \tabstyle
\caption{\textbf{Datasets and models.} $L$: image side; $s$: target side. Source: public weights or trained by us (ours). MIL: multiple-instance learning; DSMIL: dual-stream MIL network.}
  \label{tab:supp_datasets}
  \footnotesize
  \setlength{\tabcolsep}{4pt}
  \begin{tabular}{@{}llllll@{}}
    \toprule
    \tabhead{Dataset} & \tabhead{Instances} & \tabhead{$L$ (px)} & \tabhead{Ground truth} & \tabhead{Model explained} & \tabhead{Source} \\
    \midrule
    ImageNet val~\cite{deng2009imagenet} & 1,000 images & 224 & -- & ResNet-50, VGG16, ViT-B/16, DeiT-B/16 & public \\
    ImageNet-S50~\cite{gao2022imagenets} & 250 images & 224 & object mask & ResNet-50, VGG16, ViT-B/16, DeiT-B/16  & public \\
    GridPG~\cite{rao2022towards} & 500 grids per model & 448 & cell of the class & ResNet-50, VGG16 & public \\
    Synthetic Dataset & 2,680 scene--side pairs & 256--8,192 & square, $s$ 16--256 & 5 CNNs (8 pairs of model and $s$) & ours \\
    VisA~\cite{zou2022spot} & 100 defective images & 512--1,168 & defect mask & ResNet-18 per category & ours \\
    Real-IAD~\cite{wang2024realiad} & 92 defective images & 512--2,992 & defect mask & ResNet-18 per category & ours \\
    CAMELYON16 crops~\cite{bejnordi2017camelyon} & 100 crops, 49 slides & 512--8,192 & lesion mask & ResNet-18 + attention MIL~\cite{ilse2018attention} & ours \\
    CAMELYON16 slides~\cite{bejnordi2017camelyon} & 20 slides, 10 with tumour; & whole slide & tumour patches & DSMIL~\cite{li2021dsmil} & public \\
     & 39 more tumour slides & & & & \\
    DOTA~\cite{xia2018dota} & 20 images & 4,654--7,297 & object boxes & YOLO11x-OBB~\cite{khanam2024yolov11} & public \\
    \bottomrule
  \end{tabular}
\end{suppwide}

\paragraph{ImageNet.}
\label{sec:supp_datasets}
For Deletion and Insertion, we used 1,000 ImageNet validation images~\cite{deng2009imagenet}, one per class (\cref{tab:supp_datasets}). For the Pointing Games, we used 250 images of ImageNet-S50~\cite{gao2022imagenets}, five for each of its 50 classes, with their masks.

\paragraph{GridPG.} 
Each grid holds four images of different classes in a $2\times2$ grid of $448\times448$\,px, and each of the four classes is explained. As in \citet{rao2022towards}, an image enters a grid only if the model classifies it with a confidence of at least 0.99. We used 500 grids per model, each with 4 targets.

\paragraph{Synthetic Dataset.} 
Following the synthetic evaluation of h-Shap~\cite{teneggi2022hshap}, we constructed a localisation benchmark with known ground truth. Each scene contains a target pattern of side $s$, distractors drawn from four other textures, and a textured multi-scale background. The distractors have sides of 16, 32 or 64\,px, with an average of 1.5 per $512\times512$\,px. Each scene is rendered at the largest image size and centre-cropped to obtain smaller images, preserving the target across resolutions. The model predicts whether the target is present, and the ground truth is its bounding square. We include a scene only if the model predicts correctly both with and without the target, with a logit gap of at least 1 at every image size. The dataset contains 8 model–target-size combinations, with $s$ ranging from 16 to 256\,px and image side $L$ from 256 to 8,192\,px, yielding 2,680 scene–size pairs. For each model, we use 100 to 250 scenes per image size up to 1,024\,px, 30 at 2,048 and 4,096\,px, and 10 to 30 at 8,192\,px. The Synthetic Dataset can be reproduced using the code provided with our implementation.

\paragraph{Industrial inspection.}
From VisA~\cite{zou2022spot}, we used 100 defective test images, 8 or 9 for each of its 12 categories. From Real-IAD~\cite{wang2024realiad}, we used 92 defective images, 46 each of \emph{regulator} and \emph{terminal block}. These are the only two Real-IAD categories whose model reaches a sensitivity and a specificity of at least 0.9 at every image side $L$. We explain each image at native resolution on nested square crops of side 512, 1,024 and 2,048\,px where they fit, plus the largest square. The ground truth is the largest connected defect region, and the smallest crop contains all of it. The largest side is 1,168\,px for VisA and 2,992\,px for Real-IAD.

\paragraph{Histopathology.} We use the 129 official test slides of CAMELYON16~\cite{bejnordi2017camelyon}. For the crop experiments, we extracted 100 regions containing annotated lesions from 49 tumour slides, with 1 to 3 regions per slide. Each region is represented by nested square crops, ranging from 512 to 8,192\,px at approximately 1\,\textmu m per pixel. A crop is included at a given size only if the entire lesion fits within it. This gives 78, 90, 96 and 100 crops at 512, 1,024, 2,048 and 4,096\,px, respectively. At 8,192\,px, we use the first 20 crops. The annotated lesion defines the ground truth; in 7 crops, two lesions touch and their union is used as the target.

\paragraph{Slide cohorts.} On complete CAMELYON16 slides, a cohort of 20 of the 129 test slides, 10 with tumour and 10 without, gives the forward passes, the slide Deletion and HiPe without a forward-pass limit (\cref{sec:supp_hipe}); its 10 tumour slides give the patch-level AUROC and Pointing Game and the second slide model (\cref{sec:supp_abmil}). For localisation, the 39 other tumour slides from which the crops were extracted are added; of these 49 tumour slides, 45 contain at least one model patch with at least 50\,\% tumour and give the patch-level AUROC. ShapBPT~\cite{rashid2026shapbpt}, Sobol~\cite{fel2021sobol} and WCAM~\cite{kasmi2023wcam} were not evaluated on whole slides (\cref{sec:supp_census}). 50 annotated CAMELYON17~\cite{bandi2019camelyon17} slides from five hospitals are explained with the same DSMIL, without retraining (\cref{sec:supp_c17}).

\paragraph{Models.}
\label{sec:supp_models}
On ImageNet, we explain four public models as published: ResNet-50~\cite{he2016deep}, VGG16~\cite{simonyan2015very} and ViT-B/16~\cite{dosovitskiy2021image} with the torchvision\footnote{https://pypi.org/project/torchvision/} weights, and DeiT-B/16~\cite{touvron2021training} with the timm\footnote{https://pypi.org/project/timm/} weights; on GridPG, we explain ResNet-50 and VGG16. We explain the predicted-class logit on ImageNet and each target-class logit on GridPG, using the corresponding softmax probabilities for evaluation.

For whole slides, we use DSMIL~\cite{li2021dsmil} with the authors' pretrained weights on CAMELYON16. Every method perturbs the slide pixels before the feature extraction, not the stored features. The slides are divided into $224$\,px patches at $20\times$ magnification, read directly at the slide level of approximately 0.5\,\textmu m per pixel. %
 We use the authors' tissue filter to exclude patches without sufficient tissue.

For VisA and Real-IAD, we trained a separate defect classifier for each category. Each classifier uses an ImageNet-pretrained ResNet-18~\cite{he2016deep}, truncated after its third stage. It processes images at their native resolution and produces a logit map, which is aggregated by log-sum-exp pooling~\cite{pinheiro2015image}. On full test images, the classifiers achieve a mean balanced accuracy of 0.944 across the VisA categories. For CAMELYON16 crops, we use a ResNet-18 to extract patch features, which are aggregated by an attention-based multiple-instance head~\cite{ilse2018attention}. On patches excluded from the evaluation set, this model achieves a sensitivity and a specificity of at least 0.9 at every image side $L$. For the Synthetic Dataset, we trained convolutional neural networks (CNNs) with max, log-sum-exp or attention pooling to detect the target pattern. Each model achieves a sensitivity and a specificity of at least 0.9 at every image side $L$ in its evaluated range. For models with a single output, we use a sigmoid instead of a softmax to obtain probabilities. \Cref{tab:supp_hparams} shows training hyperparameters.

\input{tables/S_hparams}

\paragraph{Reduced cohorts.}
\label{sec:supp_cohorts}
\Cref{tab:supp_cohorts} lists every subset used in this supplement. The draws marked as seeded are random with a fixed seed, balanced over the classes or categories, and made from file names and labels only.

\begin{suppwide}{table}
\centering
\tabstyle\renewcommand{\arraystretch}{1.05}
\caption{\textbf{Reduced cohorts.} Size, selection rule, reason and use. Indented rows: further subsets of the same dataset.}
\label{tab:supp_cohorts}
\footnotesize
\setlength{\tabcolsep}{3.5pt}
\begin{tabular}{@{}l l >{\raggedright\arraybackslash}p{.31\textwidth} l l@{}}
\toprule
\tabhead{Cohort} & \tabhead{Size} & \tabhead{Rule} & \tabhead{Reason} & \tabhead{Used in} \\
\midrule
ImageNet & 1,000 & seeded, one per class, validation set & cost & \cref{tab:supp_imagenet_cnn,tab:supp_imagenet_vit} \\
\quad slower methods (\sa) & 250 & first 250 of the 1,000 & cost & \cref{tab:supp_imagenet_cnn,tab:supp_imagenet_vit} \\
\quad mask counts & 500 & first 500 of the 1,000 & cost & \cref{sec:supp_haar_role} \\
\quad regime rule, parameters & 100 & seeded, validation images not in the 1,000 & cost & \cref{sec:supp_vit,sec:supp_sens} \\
ImageNet-S50 & 250 & seeded, five per class & cost & \cref{tab:supp_imagenet_cnn,tab:supp_imagenet_vit} \\
\quad with a usable mask & 247 & mask neither the full image nor empty after the crop & ground truth & \cref{tab:supp_imagenet_cnn,sec:supp_lowpass} \\
\qquad single-patch masking & 244 & a $14\times14$\,px patch inside and one outside the object & ground truth & \cref{sec:supp_lowpass} \\
\quad slower methods (\sa) & 61 & first 62 of the 250, all 50 classes; 61 with a usable mask & cost & \cref{tab:supp_imagenet_cnn,tab:supp_imagenet_vit} \\
GridPG, per model & 500 grids & images with confidence $\ge.99$~\cite{rao2022towards} & protocol & \cref{tab:supp_gridpg} \\
\quad slower methods (\sa), regime rule & 100 grids & first 100 grids per model & cost & \cref{tab:supp_gridpg,sec:supp_vit} \\
VisA & 100 & seeded, 8 or 9 per category & cost & \cref{sec:supp_ablation,sec:supp_results} \\
\quad regime rule; parameters & 43; 30 & of 46 images not in the 100: whole defect within the 512\,px window; first 30 & geometry; cost & \cref{sec:supp_vit,sec:supp_sens} \\
Real-IAD & 92 & seeded, 46 per category; categories whose model reaches sensitivity and specificity $\ge.9$ at every $L$ & model accuracy & \cref{sec:supp_results} \\
CAMELYON16 crops & 100 & 1 to 3 lesion regions per slide, 49 slides; a side only if the lesion fits & ground truth & \cref{sec:supp_results} \\
\quad 8,192\,px & 20 & first 20 crops & cost & \cref{sec:supp_results} \\
Synthetic Dataset & 2,680 & 100 to 250 scenes per side up to 1,024\,px, 30 at 2,048 and 4,096\,px, 10 to 30 at 8,192\,px & cost & \cref{sec:supp_ablation,tab:supp_synth} \\
\quad parameters & 20 per side; 5 & seeded scenes not in the 2,680, up to 2,048\,px; 8,192\,px, 2 models & cost & \cref{sec:supp_sens} \\
\quad redundancy; node corner & 64; 32 & seeded scenes not in the 2,680, 2,048\,px, 1, 2 or 4 copies; target on a node corner & protocol & \cref{sec:supp_limits}; \cref{sec:supp_search} \\
CAMELYON16 slides & 20 & seeded, 10 with tumour and 10 without, of the 129 test slides & cost & \cref{sec:supp_hipe,sec:supp_abmil} \\
CAMELYON16 tumour slides & 49 & the 10 tumour slides above and the 39 slides of the crops & ground truth & \cref{tab:supp_slides} \\
\quad AUROC, Pointing Game & 45 & a model patch with at least 50\,\% tumour & ground truth & \cref{tab:supp_slides,tab:supp_cached_wsi} \\
\quad slide Deletion & 40 & classified as tumour by DSMIL & protocol & \cref{tab:supp_slides} \\
CAMELYON17 & 50 & all lesion-annotated slides & -- & \cref{sec:supp_c17} \\
\quad AUROC; slide Deletion & 35; 30 & a patch with at least 50\,\% tumour; classified as tumour & ground truth & \cref{tab:supp_c17} \\
CAMELYON16 training slides & 5; 5 & tumour slides with exhaustive annotation; 3 with tumour and 2 without & ground truth; cost & \cref{sec:supp_sens}; \cref{tab:supp_wsi_nocache} \\
2,048\,px crops, VisA, Real-IAD (RISE) & 24; 20; 23 & first 24 crops (24 slides); first 20 of the 100; first 23 images (18 samples) & cost & \cref{sec:supp_budget,sec:supp_design} \\
\bottomrule
\end{tabular}
\end{suppwide}

\subsection{Baselines}
\label{sec:supp_baselines}
We use the authors' implementations and recommended settings, with adaptations described below when necessary.

\paragraph{Occlusion and Monte Carlo sampling methods.} For Occlusion~\cite{zeiler2014visualizing}, we use the Xplique implementation~\cite{fel2022xplique} with patch and stride $L/7$, as~\citet{novello2022hsic}. RISE~\cite{petsiuk2018rise} uses random $7\times7$ masks with keep probability 0.5; following the original paper, we use 8,000 masks for ResNet-50 and 4,000 for VGG16. With no configuration given for transformers, we also use 4,000 masks for ViT-B/16 and DeiT-B/16. At high resolution, RISE uses 4,096 masks. On whole slides, the localisation runs of RISE and LIME~\cite{ribeiro2016lime} use 256 masks and 256 samples, whereas the execution times of the paper are projected for 4,096 masks and 1,000 samples (\cref{sec:supp_timing}). For Sobol~\cite{fel2021sobol}, we use 32 Quasi-Monte Carlo designs of $11\times11$ masks (3,936 forward passes); for HSIC~\cite{novello2022hsic}, $7\times7$ masks with 1,536 forward passes.

\paragraph{Linear and Shapley methods.}
LIME~\cite{ribeiro2016lime} and KernelSHAP~\cite{lundberg2017unified} use 50 superpixels from simple linear iterative clustering~\cite{achanta2012slic}; KernelSHAP and PartitionSHAP~\cite{lundberg2020partition} use the official shap package~\cite{lundberg2017unified}, PartitionSHAP with a blurred baseline. For h-Shap~\cite{teneggi2022hshap}, we use the authors' relative threshold (70th percentile of the scores at each depth) and a mean reference image. ShapBPT~\cite{rashid2026shapbpt} uses 1,000 forward passes and averages the logits over five backgrounds.

\paragraph{Subset selection and wavelets.}
LIMA~\cite{chen2025lima} and MoXI~\cite{sumiyasu2024moxi} use the settings recommended by their authors. For WCAM~\cite{kasmi2023wcam}, we use three wavelet levels, a grid size of $28\times28$ and eight designs.

\paragraph{HiPe.} 
HiPe~\cite{cooper2022hipe} replaces a masked region by its mean pixel value and splits a region when its attribution exceeds the midpoint between the minimum and maximum values at that depth. The attribution map is resized using nearest-neighbour interpolation, with a batch size of 32. On images, refinement continues until the regions have a side of 4 to 8\,px, the finest node side of \method (\cref{tab:supp_params}), without a fixed maximum depth. On whole slides, HiPe uses the same DSMIL model, slides, patches and ground truth as h-Shap.

\paragraph{References and controls.} 
On high-resolution images and the Synthetic Dataset, RISE, Sobol, HSIC and Occlusion replace masked pixels with zero as model input. We also report Saliency~\cite{simonyan2014deep}, Integrated Gradients~\cite{sundararajan2017axiomatic} (50 steps, black baseline), SmoothGrad~\cite{smilkov2017smoothgrad} (25 samples, noise 0.15), Grad-CAM~\cite{selvaraju2017gradcam} and Grad-CAM++~\cite{chattopadhay2018gradcampp}. Random and centre-weighted maps serve as independent controls.

\paragraph{Adaptations.}
Where official implementations do not support the evaluated input format, we adapt them as follows. On whole slides, LIMA and MoXI select regions on a thumbnail while evaluations use the full-resolution slide classifier; the region selections are mapped back to the slide. For ViT-B/16 and DeiT-B/16, Grad-CAM and Grad-CAM++ use the last transformer block with the token-to-image reshape of the official \texttt{pytorch-grad-cam}\footnote{https://pypi.org/project/grad-cam/} example. For MoXI, we retain the ordering and sign returned by its implementation, which ranks all patches; the source paper describes an early-stopping variant. For WCAM, we retain the approximation coefficients in accordance with the paper, rather than following the alternative choice in its notebook. These adaptations are applied consistently within each comparison and are stated to delimit, rather than extend, claims about official implementations.

\paragraph{Granularity.}
Occlusion, RISE, HSIC and Sobol use the patch and mask sizes of their official protocols, which grow with the image side; with a predefined patch side in pixels, their number of forward passes would grow with the number of patches (\cref{sec:supp_complexity}). Four experiments compare \method with baselines in settings closer to its own: HiPe, whose regions reach the finest node side of \method, without a cap on its forward passes (\cref{sec:supp_hipe}); all whole-slide baselines evaluated on the same patch-aligned cached-feature interface as \method (\cref{sec:supp_cached}); Occlusion, KernelSHAP, RISE and HiPe with the Haar low-pass baseline of \method (\cref{sec:supp_design}); and RISE, HSIC and KernelSHAP with the forward passes of \method and with up to 2,048 (\cref{sec:supp_budget}).

\subsection{Evaluation metrics}
\label{sec:supp_metrics}
\paragraph{Fidelity metrics.} Deletion~\cite{petsiuk2018rise} removes pixels in order of decreasing attribution and measures how quickly the predicted probability decreases. Insertion follows the reverse procedure, adding pixels to a blurred image and measuring how quickly the probability increases. Both metrics are computed as the area under the corresponding curve, with lower values preferred for Deletion and higher values for Insertion. At $224$\,px, we use the official RISE implementation, removing or restoring 1\,\% of the pixels (502 pixels) at each step. $\mu$Fidelity~\cite{bhatt2020evaluating} measures the Spearman correlation between the attribution of randomly selected patches and the score decrease when they are removed. We use the Xplique implementation~\cite{fel2022xplique}, as in HSIC~\cite{novello2022hsic}, with higher values indicating better fidelity.

\paragraph{Log-scale Deletion.}
At high resolution, the evidence can cover less than 0.01\,\% of the image. A Deletion curve with linear steps removes it entirely in its first step, so the curve cannot show whether the map ranked it first. On high-resolution data, we therefore remove pixels in log-spaced fractions from $2^{-18}$ to 1 of the image and set them to grey. We normalise the prediction score between the full and the fully removed image; it exceeds 1 when masking raises the score above $f(x)$. The score is the area under that curve over the log fraction (lower is better).

\paragraph{Localisation.} 
Localisation measures how well attribution maps identify annotated regions. %
The Pointing Game~\cite{zhang2016excitation} counts a hit when the top-ranked pixel of the map, at the native image size, lies on an annotated region of the image (the target square on the Synthetic Dataset; any annotated object, defect or lesion on the other datasets). When several pixels share the maximum, the top-ranked pixel is drawn at random among them, and the score of the image is the expected hit: on ImageNet-S50, the share of the tied pixels that lie on the region; on the other datasets, the mean over 20 random orders of the pixels, drawn from a fixed seed and shared by all methods. On whole slides, a hit is a top-ranked model patch that is at least half tumour. This is the Pointing Game of every table and figure. The Pointing Game breaks ties at random because counting a hit whenever any tied pixel lies on the region would reward block-wise or uniform maps: a uniform map would score 1.000. For the energy-based Pointing Game, images whose target bounding box covers more than half the image are excluded, following the original protocol~\cite{wang2020scorecam}. AUROC is computed separately for each image and then averaged, giving each image equal weight rather than pooling pixels across images as in anomaly-detection benchmarks~\cite{zou2022spot}. On whole slides, AUROC is evaluated only on tumour slides, with patches containing at least 50\% tumour considered positive.

\paragraph{EPG with a common post-processing.}
\label{sec:supp_topq}
The EPG counts the positive attribution inside the box of the evidence, so it depends on how a method scales its map and where it puts zero, not only on the order of the pixels. Every stored map is therefore also scored after one common post-processing. For a fraction $q$ of the pixels, $q\in\{0.1, 0.5, 1, 5, 10\}\,\%$, we keep the $q$ highest pixels of the map, set all other pixels to zero, and compute the official EPG on the result; pixels tied at the threshold share the remaining weight. We also report the mean over the five values of $q$. The maps of Sobol, HSIC and ShapBPT were stored at a side of 256 to 330\,px, and we score them at that size; for these maps (marked $^{d}$ in \cref{tab:supp_topq}), the full-map values may differ from \cref{tab:hr} of the main paper by up to .014. On VisA and Real-IAD, \method has the highest EPG at every $q$ and on the mean over $q$ (Holm $p=.001$ for the mean on both datasets); at $q=0.1\,\%$ on VisA, it exceeds RISE by .078 over images (Holm $p=.018$). On the CAMELYON16 crops, it has the highest EPG among the methods run on every image side; with PartitionSHAP and ShapBPT, which were not run on the largest sides, it ranks second or third, with no significant difference from the first method (Holm $p\geq.56$), as in \cref{tab:hr} of the main paper. %
The rank-normalised EPG replaces every pixel by its normalised average rank in the map and computes the EPG on these ranks; we compare each method with \method on the pairs of image and side that both have; there, no method exceeds \method by more than .001 on VisA, and \method has the highest mean on Real-IAD.

\input{tables/S_topq}

\paragraph{ROAD.}
\label{sec:supp_road}
Deletion can reward the shape of the removed region rather than the removed content~\cite{rong2022road}. ROAD~\cite{rong2022road} replaces the removed pixels by noisy linear imputation from their neighbours. On the 1,000 ImageNet images, \method has a lower ROAD score than each of the seven perturbation-based methods evaluated with ResNet-50 (.139, against .148 for PartitionSHAP to .266 for Occlusion; Holm $p\le.0007$), and than each of them except PartitionSHAP with VGG16 (.114 against .116, $p=.23$). With ViT-B/16 and DeiT-B/16, its ROAD score is .190 and .215, lower than that of each of the seven methods (Holm $p\le.0029$) except ShapBPT with DeiT-B/16 (.208, Holm $p=.049$).
On VisA and Real-IAD, ROAD is computed on the initial segment of the log-spaced removal fractions above, from $2^{-18}$ to $2^{-4}$ of the image. On this segment, \method also scores lower than each of the six perturbation-based methods evaluated on Real-IAD (Holm $p\le.0099$; PartitionSHAP and ShapBPT on 184 and 276 of the 368 pairs of image and side) and than each of them except PartitionSHAP on VisA (Holm $p\le.0006$).

\subsection{Statistics and timing}
\label{sec:supp_stats}
\paragraph{Statistics.}
Each table cell gives the mean and a 95\,\% percentile bootstrap confidence interval~\cite{efron1993introduction} (1,000 resamples of images, grids, scenes or slides). We test each method against the random map with a paired sign-flip test. We correct the $p$-values with Holm's method~\cite{holm1979simple} over the roster. We also compare methods in pairs with the same test; on the slides, the test is exact when a pair has at most 12 slides and uses 10,000 random sign flips otherwise.

\paragraph{Statistical unit.}
The sides of one image share their content, and so do the crops of one slide. In \cref{tab:hr,tab:equal} of the main paper, the unit of the bootstrap and of the test is the pair of image and side. We also repeat both with a unit that groups the dependent pairs: the image on VisA (100 images, all sides pooled), the physical sample on Real-IAD (category and sample id; 35 samples, 92 images, each sample photographed from several cameras), and the slide on the CAMELYON16 crops (100 crops from 49 slides). For each unit, we average the paired differences over its pairs, then flip the signs of the unit means and resample the units with all their pairs; on the Synthetic Dataset, each scene is already one unit.
\label{sec:supp_unit}
With the grouped units, 94 of the 99 comparisons of \cref{tab:hr} and 349 of the 354 paired tests behind \cref{tab:equal} of the main paper keep their significance outcome, and no significant effect reverses direction.

\paragraph{Timing protocol.}
\label{sec:supp_timing}
For each method and image side, the timing pass makes one timed call per image, after one untimed warm-up call, and reports the median over 4 images at 224 to 2,048\,px (3 for h-Shap at 224\,px) and the time of 1 image at 4,096\,px, at 8,192\,px and on the slide. \Cref{sec:supp_cached} also reports the times of all whole-slide methods on the same cached-feature interface. A method that would need hours on the slide (RISE, LIME, Sobol, WCAM) runs only the start of its sequence of forward passes (120; 10 for WCAM), and we project its time as the measured time plus the remaining forward passes times the time per forward pass. The timer covers the whole method call, including the baseline, setup and the CPU-side estimation.

\paragraph{Cost of every method.}
\label{sec:supp_complexity}
\label{sec:supp_census}
Under the official protocols of \cref{sec:supp_baselines}, the perturbation-based methods keep their forward passes nearly constant across image sides (\cref{tab:supp_cost}; whole-slide times with and without the feature cache in \cref{tab:supp_cached_wsi,tab:supp_wsi_nocache}) by enlarging their perturbation regions with $L$ (LIME keeps 1,000 samples over a predefined number of superpixels). With patches of a predefined side $c$ in pixels, the number of patches is $(L/c)^2$: Occlusion needs $1+(L/c)^2$ forward passes, Sobol $32((L/c)^2+2)$~\cite{fel2021sobol}, WCAM $8((L/c)^2+2)$, and KernelSHAP at least $(L/c)^2$ samples. Pairwise methods scale as $L^4$, with $d(d+1)/2$ forward passes for MoXI~\cite{sumiyasu2024moxi} ($d=(L/c)^2$) and a similar number for LIMA~\cite{chen2025lima}, whereas RISE, HSIC and LIME choose their number of samples independently of the number of patches. For \method, \cref{prop:cost} gives the forward passes of the search; ShapBPT needs memory that grows with $L^2$ to build its partition tree~\cite{rashid2026shapbpt}. From 4,096 to 8,192\,px, the execution time grows 4.7 times for \method (1.10 to 5.20\,s on two timed scenes with 208 and 176 forward passes), against 1.8 (Occlusion) to 6.3 (KernelSHAP and h-Shap).

\input{tables/T1_cost}

%% file: tables/S_hparams.tex
\begin{suppwide}{table}
\centering
\tabstyle
\caption{\textbf{Summary of model training.} $s$: target side; $L$: image side.}
\label{tab:supp_hparams}
\footnotesize
\setlength{\tabcolsep}{4pt}
\renewcommand{\arraystretch}{1.15}
\begin{tabular}{@{}>{\raggedright\arraybackslash}p{2.1cm} >{\raggedright\arraybackslash}p{4.6cm} >{\raggedright\arraybackslash}p{4.3cm} >{\raggedright\arraybackslash}p{4.7cm}@{}}
\toprule
 & \tabhead{Synthetic Dataset} & \tabhead{VisA, Real-IAD} & \tabhead{CAMELYON16 crops} \\
\midrule
Architecture & Custom fully convolutional network (stride 16, receptive field 128 or 1,120\,px) with max, log-sum-exp, or gated attention pooling~\cite{ilse2018attention}. & ResNet-18 (up to stage 3, stride 16) with log-sum-exp pooling. & ResNet-18 (stage 4, stride 32), gated attention head~\cite{ilse2018attention} (hidden size 128). \\
Parameters & 61k--165k & 2.8M & 11.2M (patch), 132k (head) \\
Initialisation & Random  & ImageNet (frozen batch norm) & ImageNet (patch), random (head) \\
Training data & 32k--56k target size $s\in\{16..256\}$. & Official splits. & 29 slides (61k patches; 3k tiles for head). \\
Input \& augmentation & Crops $L\in\{256..8192\}$; flips and $90^\circ$ rotations. & Two-phase: patches (256 or 384\,px) then crops or full image; flips, rotations, colour. & 128\,px patches (512--4,096\,px tiles for head); geometric, stain, and blur. \\
Optimisation & Binary cross-entropy, AdamW~\cite{loshchilov2019decoupled} (learning rate $10^{-3}$ to $3{\times}10^{-4}$), cosine decay~\cite{loshchilov2017sgdr}, batch size 48--64. & Binary cross-entropy, AdamW (learning rate $3{\times}10^{-4}$ to $5{\times}10^{-5}$), one-cycle~\cite{smith2017superconvergence}, batch size 32 then 4. & Binary cross-entropy, AdamW (learning rate $3{\times}10^{-4}$ patch, $2{\times}10^{-4}$ head), batch size 256 / 8. \\
Duration & 4,000--7,000 steps & 2,100 steps (VisA), 3,800 steps (Real-IAD) & 12 epochs (patch), 40 epochs (head) \\
Selection \& criteria & Minimum validation loss. Target: sensitivity and specificity $\ge 0.9$ for all $L\in[256, 8192]$. & Maximum development balanced accuracy. Target: sensitivity and specificity $\ge 0.9$ across $L$ and full image. & Maximum development AUROC (patch), last epoch (head). Target: sensitivity and specificity $\ge 0.9$. \\
\bottomrule
\end{tabular}
\end{suppwide}

%% file: tables/S_topq.tex
\begin{suppcol}{table}
\centering
\tabstyle\renewcommand{\arraystretch}{.92}
\caption{\textbf{EPG under common top-$q$ post-processing.} EPG ($\uparrow$), mean over image sides, of the full map and of the top $q$\,\% of the pixels; Mean: over $q$. Bold/underline: best/second among perturbation-based methods; grey: references. $^{r}$\,Map stored at 256 to 330\,px. $^{a}$\,Not run on the largest image sides because of its cost (\cref{tab:supp_cost}); not ranked.}
\label{tab:supp_topq}
\footnotesize
\setlength{\tabcolsep}{3.2pt}
\newcommand{\topqgroup}[1]{\addlinespace[3pt]\multicolumn{8}{@{}l}{\tabheadfont #1}\\[1pt]}%
\begin{tabular}{@{}l c@{\hspace{8pt}}ccccc@{\hspace{8pt}}c@{}}
\toprule
 & & \multicolumn{5}{c}{\tabhead{top $q$ (\%) of the pixels}} & \\
\cmidrule(lr){3-7}
\tabhead{Method} & \tabhead{Full} & \tabhead{0.1} & \tabhead{0.5} & \tabhead{1} & \tabhead{5} & \tabhead{10} & \tabhead{Mean} \\
\midrule
\multicolumn{8}{@{}l}{\tabheadfont VisA}\\[1pt]
Occlusion & .237 & .248 & .248 & .248 & .241 & .238 & .244 \\
RISE & .021 & \second{.585} & \second{.480} & .405 & .195 & .122 & .358 \\
LIME & .056 & .413 & .319 & .257 & .126 & .092 & .242 \\
KernelSHAP & .116 & .159 & .159 & .159 & .138 & .125 & .148 \\
PartitionSHAP\sa & .377 & .514 & .476 & .448 & .395 & .384 & .443 \\
h-Shap & \second{.429} & .428 & .428 & \second{.428} & \second{.429} & \second{.429} & \second{.428} \\
ShapBPT\smash{$^{r}$} & .181 & .369 & .348 & .324 & .253 & .227 & .304 \\
Sobol\smash{$^{r}$} & .281 & .450 & .407 & .370 & .292 & .284 & .361 \\
HSIC\smash{$^{r}$} & .194 & .379 & .355 & .326 & .226 & .205 & .298 \\
LIMA & .039 & .205 & .205 & .205 & .146 & .105 & .173 \\
MoXI & .019 & .247 & .164 & .127 & .046 & .031 & .123 \\
WCAM & .076 & .257 & .226 & .200 & .142 & .117 & .188 \\
\ours{\method} & \best{.648} & \best{.658} & \best{.649} & \best{.649} & \best{.648} & \best{.648} & \best{.650} \\
\ctrl{Random map} & \ctrl{.019} & \ctrl{.019} & \ctrl{.019} & \ctrl{.019} & \ctrl{.019} & \ctrl{.019} & \ctrl{.019} \\
\ctrl{Centre map} & \ctrl{.034} & \ctrl{.085} & \ctrl{.084} & \ctrl{.077} & \ctrl{.068} & \ctrl{.064} & \ctrl{.076} \\
\topqgroup{Real-IAD}
Occlusion & .298 & .327 & .327 & .327 & .306 & .299 & .317 \\
RISE & .031 & \second{.769} & \second{.617} & .515 & .277 & .191 & .474 \\
LIME & .093 & .586 & .412 & .336 & .196 & .148 & .336 \\
KernelSHAP & .243 & .305 & .305 & .305 & .271 & .253 & .287 \\
PartitionSHAP\sa & .745 & .853 & .828 & .803 & .755 & .747 & .797 \\
h-Shap & .357 & .359 & .358 & .357 & .357 & .357 & .357 \\
ShapBPT\smash{$^{r}$}\sa & .212 & .452 & .455 & .432 & .333 & .292 & .393 \\
Sobol\smash{$^{r}$} & \second{.390} & .621 & .578 & \second{.527} & \second{.411} & \second{.395} & \second{.507} \\
HSIC\smash{$^{r}$} & .268 & .453 & .441 & .423 & .320 & .288 & .385 \\
LIMA & .036 & .109 & .127 & .137 & .125 & .094 & .118 \\
MoXI & .033 & .324 & .251 & .200 & .090 & .058 & .184 \\
WCAM & .097 & .155 & .153 & .153 & .151 & .139 & .150 \\
\ours{\method} & \best{.895} & \best{.905} & \best{.897} & \best{.895} & \best{.895} & \best{.895} & \best{.898} \\
\ctrl{Random map} & \ctrl{.029} & \ctrl{.029} & \ctrl{.028} & \ctrl{.028} & \ctrl{.029} & \ctrl{.029} & \ctrl{.029} \\
\ctrl{Centre map} & \ctrl{.047} & \ctrl{.074} & \ctrl{.078} & \ctrl{.080} & \ctrl{.080} & \ctrl{.078} & \ctrl{.078} \\
\topqgroup{CAMELYON16 crops}
Occlusion & .120 & .139 & .139 & .139 & .134 & .128 & .136 \\
RISE & .049 & .182 & .172 & .162 & .125 & .105 & .149 \\
LIME & .064 & .125 & .108 & .101 & .084 & .077 & .099 \\
KernelSHAP & .148 & .169 & .169 & .169 & .164 & .157 & .166 \\
PartitionSHAP\sa & .244 & .267 & .262 & .260 & .254 & .248 & .259 \\
h-Shap & \second{.185} & .199 & .195 & .189 & \second{.185} & \second{.185} & .190 \\
ShapBPT\smash{$^{r}$}\sa & .181 & .262 & .264 & .255 & .220 & .205 & .241 \\
Sobol\smash{$^{r}$} & .160 & \second{.220} & \second{.208} & \second{.198} & .176 & .169 & \second{.194} \\
HSIC\smash{$^{r}$} & .122 & .168 & .166 & .161 & .143 & .136 & .155 \\
LIMA\sa & .096 & .121 & .121 & .121 & .114 & .107 & .117 \\
MoXI & .039 & .059 & .042 & .034 & .028 & .029 & .038 \\
WCAM & .079 & .143 & .137 & .131 & .106 & .098 & .123 \\
\ours{\method} & \best{.220} & \best{.230} & \best{.224} & \best{.221} & \best{.220} & \best{.220} & \best{.223} \\
\ctrl{Random map} & \ctrl{.047} & \ctrl{.046} & \ctrl{.046} & \ctrl{.047} & \ctrl{.047} & \ctrl{.047} & \ctrl{.047} \\
\ctrl{Centre map} & \ctrl{.083} & \ctrl{.174} & \ctrl{.170} & \ctrl{.168} & \ctrl{.160} & \ctrl{.151} & \ctrl{.165} \\
\bottomrule
\end{tabular}
\end{suppcol}

%% file: tables/T1_cost.tex
\begin{suppcol}{table}
\centering
\tabstyle\renewcommand{\arraystretch}{1}%
\caption{\textbf{Cost across resolutions.} Forward passes (backward for gradient methods) at 224\,px, 2k\,px, 8k\,px and on one slide. Fixed patch: order in $L$ (none: independent of patches; $k$: nodes per depth). $^{\P}$: cached patch features. \textit{Italics}: projected. $^{\S}$: timed image only. n.r.$^{d,m,t}$: design, memory, or time limit.}
\label{tab:supp_cost}
\footnotesize
\setlength{\tabcolsep}{1.8pt}
\begin{tabular}{@{}l cccc c cc@{}}
\toprule
 & \multicolumn{4}{c}{\tabhead{Forward passes}} & \tabhead{Fixed} & \multicolumn{2}{c}{\tabhead{Time (s)}} \\
\cmidrule(lr){2-5}\cmidrule(lr){6-6}\cmidrule(l){7-8}
\tabhead{Method} & \tabhead{224} & \tabhead{2k} & \tabhead{8k} & \tabhead{Slide} & \tabhead{patch} & \tabhead{2k} & \tabhead{Slide} \\
\midrule
\tabgroup{8}{Gradient- and activation-based}
Saliency & 1 & 1 & 1 & 1 & 1 & 0.003 & 5.8 \\
Int.\ Gradients & 51 & 51 & 51 & 51 & 1 & 0.13 & 207 \\
SmoothGrad & 25 & 25 & 25 & 25 & 1 & 3.5 & 741 \\
Grad-CAM & 1 & 1 & 1 & 1 & 1 & 0.002 & 4.0 \\
Grad-CAM++ & 1 & 1 & 1 & 1 & 1 & 0.10 & 23.5 \\
\addlinespace[2pt]
\tabgroup{8}{Perturbation-based}
Occlusion & 50 & 50 & 50 & 50 & $L^2$ & 0.21 & 0.6 \\
RISE & 8,000 & 4,096 & 4,096 & \textit{4,096} & none & 136 & \textit{38,691} \\
LIME & 1,000 & 1,000 & 1,000 & \textit{1,000} & none & 171 & \textit{6,390} \\
KernelSHAP & 1,002 & 1,002 & 1,002 & 1,002 & $L^2$ & 4.9 & 80.1 \\
PartitionSHAP & 5,000 & 5,000 & \nrm & 500 & $k\log L$ & 846 & 2.4$^{\P}$ \\
h-Shap & 64 & 64 & 64 & 27,968 & $k\log L$ & 0.05 & 255 \\
ShapBPT$^{1}$ & 1,000 & 1,000 & 1,000$^{\S}$ & \nrm & $k\log L$ & 41.0 & \nrm \\
Sobol$^{2}$ & 3,936 & 3,936 & 3,936 & \nrd & $L^2$ & 4.2 & \textit{16,361} \\
HSIC & 1,536 & 1,536 & 1,536 & 4,096 & none & 1.6 & 766 \\
LIMA & 1,766 & 1,742 & \nrt & 1,702 & $L^4$ & 367 & 982 \\
MoXI & 19,306 & 19,306 & 19,306 & 19,307 & $L^4$ & 13.6 & 1,707 \\
WCAM$^{3}$ & 6,288 & 6,288 & 6,288 & \nrd & $L^2$ & 14.3 & \textit{45,939} \\
\addlinespace[2pt]
\tabgroup{8}{Ours}
\ours{\method} & 1,024 & 200$^{\S}$ & 176$^{\S}$ & 458 & $k\log L$ & 0.38 & 6.3$^{\P}$ \\
\bottomrule
\end{tabular}
\par\vspace{2pt}\raggedright\footnotesize
$^{1}$\,ShapBPT (slide): 364\,s, 5.3\,GiB at 16\,Mpx; proj.\ $\ge$34\,h, $\approx$1.8\,TB.\\
$^{2}$\,Sobol (slide, design): 7.7\,s/pass; proj.\ 2,112 passes ($\approx$4.5\,h).\\
$^{3}$\,WCAM (slide, design): 7.3\,s/pass; proj.\ 6,288 passes ($\approx$13\,h).
\end{suppcol}

%% file: supplementary/supp_D_transfer.tex
\section{Generalisation across models and datasets}
\label{sec:supp_more}
This section covers a second slide model (\cref{sec:supp_abmil}), other hospitals (\cref{sec:supp_c17}), the cached interface (\cref{sec:supp_cached}) and a public object detector on aerial images (\cref{sec:supp_yolo}).

\subsection{Second slide model}
\label{sec:supp_abmil}
The second slide model is a gated attention-based multiple-instance network (ABMIL)~\cite{ilse2018attention} that replaces the aggregation head of DSMIL on the same patch features. Trained on 216 CAMELYON16 training slides and selected on 54 validation slides (epoch 36, validation AUC .952), ABMIL attains a slide-level test AUC of .972 [.946, .992] across the 129 test slides, comparable to DSMIL (.970). \method and the perturbation-based baselines explain it on the 10 tumour slides of the 20-slide cohort (\cref{sec:supp_setup}) using the same settings as for DSMIL, together with single-patch leave-one-out perturbation and ABMIL's internal attention weights as references (\cref{tab:supp_abmil}). Slide Deletion uses the 9 slides classified as tumour (RISE and HSIC: 6; h-Shap: 4).

On ABMIL, \method has a Pointing Game of 1.000 with a median of 200 forward passes. It exceeds Occlusion in patch-level AUROC on all 10 slides (Holm $p=.008$) and in slide Deletion across the 9 slides classified as tumour (by .155 [.079, .233], Holm $p=.016$), has higher mean AUROC than RISE and HSIC on the 7 slides they completed (no significant difference, Holm $p=.094$), and surpasses exhaustive single-patch leave-one-out perturbation by $+.377$ AUROC while requiring roughly 90 times fewer forward passes. Within the 40,000-pass model limit, h-Shap completes only 4 of the 10 slides (a run stops at the first forward pass above the limit, the 40,001st; its count of 42,667 adds 2,666 repeated identical masks answered from a cache) with lower Pointing Game accuracy and no statistically significant AUROC difference from \method ($p=.25$). As an internal-access reference, ABMIL's native attention achieves higher AUROC ($p=.002$).

\begin{suppcol}{table}
\centering
\tabstyle
\caption{\textbf{Whole-slide localisation on ABMIL.} $n$: slides for AUROC; PG: Pointing Game; Del.: slide Deletion; forward passes: median per slide. AUROC, PG: $\uparrow$. $\Delta$: \method $-$ method on common slides ($+$: \method is better), 95\,\% bootstrap CI; $p$: Holm-corrected over perturbation-based methods. Grey: references.}
\label{tab:supp_abmil}
\footnotesize
\setlength{\tabcolsep}{1.3pt}
\begin{suppfit}
\begin{tabular}{@{}l c c r@{}l c c c r@{}}
\toprule
\tabhead{Method} & \tabhead{$n$} & \tabhead{AUROC} & \multicolumn{2}{c}{\tabhead{$\Delta$ AUROC}} & \tabhead{$p$} & \tabhead{PG} & \tabhead{Del.\,$\downarrow$} & \tabhead{\shortstack{Forward\\passes}} \\
\midrule
\method & 10 & .939 & \multicolumn{2}{c}{\na} & \na & 1.000 & .562 & 200 \\
Occlusion & 10 & .696 & $+.243$ & \dci{$+.140$}{$+.354$} & .008 & .215 & .717 & 50 \\
RISE & 7 & .766 & $+.178$ & \dci{$+.060$}{$+.305$} & .094 & .357 & .650 & 256 \\
HSIC & 7 & .808 & $+.136$ & \dci{$+.048$}{$+.232$} & .094 & .286 & .636 & 1,024 \\
h-Shap & 4 & .980 & $-.039$ & \dci{$-.071$}{$-.007$} & .25 & .500 & .410 & 42,667 \\
\ctrl{Leave-one-out} & \ctrl{10} & \ctrl{.561} & \ctrl{$+.377$} & \dci{$+.227$}{$+.519$} & \na & \ctrl{1.000} & \ctrl{.539} & \ctrl{18,299} \\
\ctrl{Attention} & \ctrl{10} & \ctrl{.995} & \ctrl{$-.056$} & \dci{$-.094$}{$-.026$} & \na & \ctrl{1.000} & \ctrl{.279} & \ctrl{1} \\
\bottomrule
\end{tabular}
\end{suppfit}
\end{suppcol}

\subsection{Other hospitals}
\label{sec:supp_c17}
The 50 lesion-annotated lymph-node slides of CAMELYON17~\cite{bandi2019camelyon17} come from five hospitals, with other staining protocols and slide scanners; the CAMELYON16 DSMIL explains them without retraining. Because all 50 slides contain tumour, we evaluate patch-level AUROC and Pointing Game on the 35 slides containing at least one patch with at least 50\,\% tumour, and slide Deletion on the 30 slides classified as tumour by DSMIL (3 to 7 of 10 per hospital). Paired differences against Occlusion, RISE, and h-Shap are evaluated on the slides explained by both methods using 95\,\% bootstrap confidence intervals and two-sided paired tests with Holm correction~\cite{holm1979simple} (\cref{tab:supp_c17}).

On these slides, \method has a mean patch-level AUROC of .948 with a median of 182 forward passes per slide, exceeding h-Shap (mean AUROC .905 on 33 slides), RISE, and Occlusion in both patch-level AUROC (Holm $p<.001$ for each) and slide Deletion, with over 100 times fewer forward passes than h-Shap. In slide Deletion, \method also outperforms a random patch ordering by $+.043$ [.024, .064]. Across hospitals, \method's patch-level AUROC remains between .892 and .981, reaching .980 on the 16 slides with macro-metastases (largest lesion above 2\,mm) and .918 on the 18 slides with micro-metastases (0.2 to 2\,mm)~\cite{bandi2019camelyon17}; the remaining slide has only isolated tumour cells.

\begin{suppwide}{table}
\centering
\tabstyle
\caption{\textbf{Cross-hospital generalisation on CAMELYON17.} \method row: means; other rows: paired difference in AUROC and Del.\ ($+$: \method is better). [\,]: 95\,\% CI. $n$: slides for AUROC / Deletion; PG: Pointing Game; Del.: slide Deletion; Fwd.: median forward passes.}
\label{tab:supp_c17}
\footnotesize
\setlength{\tabcolsep}{5pt}
\begin{tabular}{@{}l c r@{}l c r@{}l r@{}}
\toprule
\tabhead{Method} & \tabhead{$n$} & \multicolumn{2}{c}{\tabhead{AUROC / $\Delta$ $\uparrow$}} & \tabhead{PG $\uparrow$} & \multicolumn{2}{c}{\tabhead{Del.\ / $\Delta$ $\downarrow$}} & \tabhead{Fwd.} \\
\midrule
\method & 35/30 & .948 & \dci{.918}{.970} & .714 & .538 & & 182 \\
h-Shap & 33/29 & $+.046$ & \dci{$+.022$}{$+.070$} & .030 & $+.032$ & \dci{$+.019$}{$+.048$} & 19,328 \\
RISE & 24/20 & $+.329$ & \dci{$+.251$}{$+.410$} & .042 & $+.039$ & \dci{$+.017$}{$+.064$} & 256 \\
Occlusion & 35/30 & $+.359$ & \dci{$+.279$}{$+.436$} & .074 & $+.027$ & \dci{$+.014$}{$+.040$} & 50 \\
\bottomrule
\end{tabular}
\end{suppwide}

\subsection{Cached-feature interface}
\label{sec:supp_cached}
\paragraph{Evaluation of a masked slide.}
Before the explanation, the original patches and their low-pass versions are processed by the DSMIL tissue filter. The patch encoder extracts features only from patches that pass the filter, and these features are stored. For a masked slide, DSMIL builds a bag from the stored features and passes it to the multiple-instance aggregator. Unchanged patches contribute their original features. Masked patches contribute their low-pass features only if they pass the tissue filter. Since the low-pass replacement is the mean colour of a square of side $L/16$, it is nearly uniform and usually fails the filter. Such patches are therefore excluded from the bag. \method thus treats the multiple-instance aggregator as a black box on the patch embeddings that the DSMIL pipeline extracts and filters once. Because every mask is a union of whole patches, the stored embeddings give exactly the forward pass of DSMIL on the masked slide, and \method evaluates different masks without running the patch encoder again. In contrast, Occlusion and h-Shap may mask only part of a patch. These patches mix original and low-pass pixels and are re-encoded if they still pass the filter.

\paragraph{Masked patches in the bag.}
\label{sec:supp_corrupt}
\label{sec:supp_fixedbag}
Keeping every masked patch in the bag with the features of its low-pass pixels, on the 20 slides of the cohort, leaves \method nearly unchanged (\cref{fig:supp_fixed_bag}): on the 10 tumour slides, patch-level AUROC differs by $-.002$ [$-.004$, $-.000$], Pointing Game remains 1.000, and map Spearman correlations between the two protocols range from .925 to 1.000, while slide Deletion differs by less than .001 across all 20 slides. For Occlusion, the map correlations fall to $-.594$ and the slide Deletion is higher by .008 (Holm $p=.001$).

\begin{suppwide}{figure}
\includegraphics[width=\linewidth]{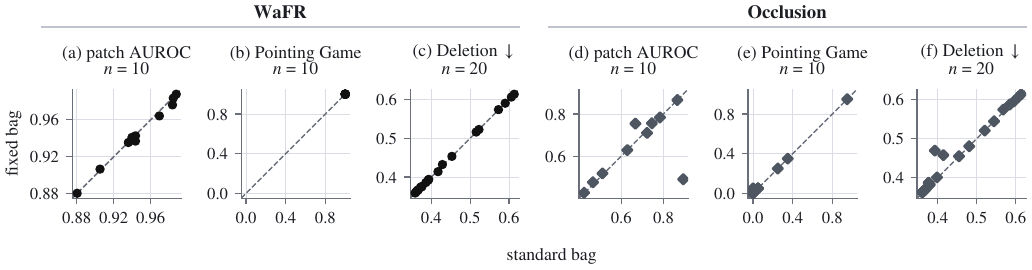}
\caption{\textbf{Whole-slide sensitivity to retaining masked patches in the bag.} Standard bag: masked patches excluded by the tissue filter; fixed bag: masked patches retained. One point per slide; dashed: identity.}
\label{fig:supp_fixed_bag}
\end{suppwide}

\paragraph{All methods on the interface.}
In pixel space, the execution time of a whole-slide method also counts the encoding of partially masked patches. We therefore run all whole-slide baselines on the same patch-aligned cached-feature interface, with whole-patch masks, so that no method runs the patch encoder during the explanation (\cref{tab:supp_cached_wsi}). On the 45 tumour slides of the patch-level AUROC (\cref{sec:supp_setup}), \method has the highest patch-level AUROC and Pointing Game, above each of the ten baseline configurations in paired AUROC by .03 to .32 (two-sided paired test, Holm $p\le.049$); the comparison with h-Shap uses the 31 slides that it completes within 40,000 forward passes. Forward passes and times in \cref{tab:supp_cached_wsi} are medians over all 49 tumour slides. The one-time patch feature extraction (median 37\,s per slide on 10 slides) is shared by all methods and therefore does not affect the comparison between them.

\input{tables/S_cached_wsi}

Without the cache, \method takes 8 to 32 minutes per slide (median 16), within the range of the methods run with few forward passes, Occlusion with 50 and RISE and LIME with 256 (3 to 35 minutes, \cref{tab:supp_wsi_nocache}). Their localisation is worse (Pointing Game .02 to .18 against .89 for \method, \cref{tab:supp_cached_wsi}). With its default 1,000 samples, LIME takes 2 to 8 times as long as \method and barely changes the localisation on this model (AUROC .815, Pointing Game .19). At their default budgets, the other methods take from about one hour to several days per slide.

\input{tables/S_wsi_nocache}

\subsection{Object detection on aerial images}
\label{sec:supp_yolo}
The experiments of the main paper explain classifiers. An object detector also predicts where each object lies, but the question that an attribution map answers stays the same: which regions of the image produce the prediction score of a class. We therefore apply \method to a detector, on a task and a domain that the main paper does not cover, without changing its search, its baseline or its parameters.

DOTA~\cite{xia2018dota} contains aerial and satellite images annotated with oriented boxes of 15 object classes. We filter 20 images of the DOTA v1.0 validation split with a longest side of at least 4,000\,px that contain 1 to 3 objects of a class, the largest first and at most two per class, before computing any output of the detector. Their longest side ranges from 4,654 to 7,297\,px, whereas the target objects have a median longest side of 124\,px, and four of them are smaller than 30\,px.

The detector is YOLO11x-OBB~\cite{khanam2024yolov11}, with its public weights trained on DOTA v1.0, which processes an image in overlapping patches of 1,024\,px; we explain $f_c$, the largest pre-sigmoid score (logit) of class $c$ over all patches and anchors before non-maximum suppression. With $s_{\text{in}}=1{,}024$\,px, the rule $r=L/s_{\text{in}}>1$ of \cref{sec:rule} of the main paper selects the hierarchical search on every image ($r$ from 4.5 to 7.1). The methods are those of \cref{tab:hr} of the main paper, compared among perturbation-based methods, with Integrated Gradients and Grad-CAM as separately evaluated internal-access references, and the metrics are those of \cref{sec:supp_metrics}, with AUROC and the Pointing Game computed over all objects of the explained class. PartitionSHAP, ShapBPT and LIMA did not run at these image sizes because of memory or time constraints.

\method has the highest AUROC (.975) and EPG (.533), with a median of 310 forward passes per image, and its paired differences are positive against every other method on both metrics (\cref{tab:supp_yolo}; paired sign-flip tests, Holm $p\le.03$). Integrated Gradients has a higher Pointing Game (.800) but an AUROC of .494: since $f_c$ is a maximum over patches, its gradient is zero outside the winning patch, so Integrated Gradients marks one object and leaves the rest of the image, including the other objects of class $c$, tied at zero. \Cref{fig:supp_gal_yolo} shows the maps on four of the images.

\input{tables/S_yolo}

%% file: tables/S_cached_wsi.tex
\begin{suppcol}{table}
\centering
\tabstyle
\caption{\textbf{Whole-slide comparison on a shared cached-feature interface.} PG: Pointing Game; Fwd.: median forward passes; Time: median warm time. Values separated by /: one per number of forward passes in Fwd.\ (HSIC: 256, 1,024 and 4,096). [\,]: 95\,\% CI. \sa\,33 of 49 slides (31 of 45 for AUROC and PG) completed within the limit of 40,000 forward passes.}
\label{tab:supp_cached_wsi}
\footnotesize
\setlength{\tabcolsep}{3pt}
\begin{tabular}{@{}l cc rr@{}}
\toprule
\tabhead{Method} & \tabhead{AUROC $\uparrow$} & \tabhead{PG $\uparrow$} & \tabhead{Fwd.} & \tabhead{Time (s)} \\
\midrule
Occlusion & .624 & .12 & 50 & 0.05 \\
RISE & .638/.784 & .02/.18 & 256/4,096 & 0.30/4.73 \\
LIME & .818/.815 & .18/.19 & 256/1,000 & 1.37/1.83 \\
HSIC & .793/.796/.796 & .16/.16/.18 & 256--4,096 & 0.22--3.71 \\
PartitionSHAP & .736 & .02 & 500 & 1.97 \\
h-Shap\sa & .917 & .02 & 24,151 & 21.91 \\
\addlinespace[2pt]
\ours{\method} & \best{.949}\,[.929,.965] & \best{.89} & 178 & 1.06 \\
\bottomrule
\end{tabular}
\end{suppcol}

%% file: tables/S_wsi_nocache.tex
\begin{suppcol}{table}
\centering
\tabstyle
\caption{\textbf{Whole-slide time without the feature cache.} Five CAMELYON16 training slides, three with tumour; Fwd.: forward passes; time per slide: median [range]. \method: measured end to end; other methods: setup plus Fwd.\ times the median of 12 timed forward passes; HiPe: median Fwd.\ of \cref{sec:supp_hipe}.}
\label{tab:supp_wsi_nocache}
\footnotesize
\setlength{\tabcolsep}{3pt}
\begin{tabular}{@{}l r@{\hspace{10pt}}r@{\hspace{6pt}}r@{,\ }r@{}}
\toprule
\tabhead{Method} & \tabhead{Fwd.} & \multicolumn{3}{r}{\tabhead{Time per slide}} \\
\midrule
Occlusion & 50 & 5.2\,min & [3.1 & 6.3\,min] \\
RISE & 256 & 17\,min & [11 & 29\,min] \\
LIME & 256 & 19\,min & [15 & 35\,min] \\
PartitionSHAP & 500 & 53\,min & [29\,min & 1.1\,h] \\
LIME & 1,000 & 1.1\,h & [0.9 & 2.1\,h] \\
KernelSHAP & 1,002 & 1.3\,h & [0.7 & 1.6\,h] \\
LIMA & 1,523--1,828 & 1.7\,h & [1.1 & 3.5\,h] \\
HSIC & 4,096 & 4.3\,h & [3.2 & 7.2\,h] \\
RISE & 4,096 & 4.4\,h & [2.9 & 7.6\,h] \\
h-Shap & 12,916--34,636 & 19\,h & [8\,h & 2.1\,d] \\
MoXI & 19,307 & 20\,h & [14 & 24\,h] \\
HiPe & 88,997 & 3.9\,d & [3.4 & 8.9\,d] \\
\midrule
\ours{\method} & 164--456 & 16\,min & [8.2 & 32\,min] \\
\bottomrule
\end{tabular}
\end{suppcol}

%% file: tables/S_yolo.tex
\begin{suppwide}{table}
\centering
\tabstyle
\caption{\textbf{Object detection.} $n$: images with a map; $\pm$: 95\,\% CI; $\Delta$: \method $-$ method ($+$: \method is better) with 95\,\% CI; forward passes: median per image; bold\,/\,underline: best\,/\,second among perturbation-based methods; n.r.$^{m,t}$: memory or time limit; grey: controls.}
\label{tab:supp_yolo}
\footnotesize
\setlength{\tabcolsep}{3.2pt}
\begin{tabular}{@{}l r c r@{}l c r@{}l c r@{}l r@{}}
\toprule
 & & \multicolumn{3}{c}{\tabhead{AUROC $\uparrow$}} & \multicolumn{3}{c}{\tabhead{Pointing Game $\uparrow$}} & \multicolumn{3}{c}{\tabhead{EPG $\uparrow$}} & \\
\cmidrule(lr){3-5}\cmidrule(lr){6-8}\cmidrule(lr){9-11}
\tabhead{Method} & \tabhead{$n$} & \tabhead{mean} & \multicolumn{2}{c}{\tabhead{$\Delta$}} & \tabhead{mean} & \multicolumn{2}{c}{\tabhead{$\Delta$}} & \tabhead{mean} & \multicolumn{2}{c}{\tabhead{$\Delta$}} & \tabhead{\shortstack{Forward\\passes}} \\
\midrule
\tabgroup{11}{perturbation-based}
Occlusion & 20 & .857\ci{.069} & $+.118$ & \dci{$+.065$}{$+.177$} & .055\ci{.041} & $+.665$ & \dci{$+.473$}{$+.840$} & .037\ci{.036} & $+.495$ & \dci{$+.319$}{$+.670$} & 50 \\
RISE & 20 & .927\ci{.072} & $+.047$ & \dci{$+.005$}{$+.113$} & .250\ci{.175} & $+.470$ & \dci{$+.265$}{$+.682$} & .005\ci{.007} & $+.527$ & \dci{$+.338$}{$+.710$} & 4,096 \\
LIME & 20 & .800\ci{.066} & $+.175$ & \dci{$+.118$}{$+.233$} & \second{.467}\ci{.181} & $+.252$ & \dci{$+.060$}{$+.447$} & .009\ci{.007} & $+.524$ & \dci{$+.343$}{$+.706$} & 1,000 \\
KernelSHAP & 20 & .938\ci{.050} & $+.036$ & \dci{$+.013$}{$+.074$} & .020\ci{.024} & $+.700$ & \dci{$+.510$}{$+.868$} & .031\ci{.027} & $+.501$ & \dci{$+.317$}{$+.677$} & 1,002 \\
PartitionSHAP & \na & \multicolumn{9}{c}{\ctrl{n.r.$^{m,t}$}} & \ctrl{\na} \\
ShapBPT & \na & \multicolumn{9}{c}{\ctrl{n.r.$^{m,t}$}} & \ctrl{\na} \\
h-Shap & 20 & .759\ci{.080} & $+.216$ & \dci{$+.146$}{$+.285$} & .105\ci{.090} & $+.615$ & \dci{$+.425$}{$+.792$} & .027\ci{.020} & $+.506$ & \dci{$+.323$}{$+.680$} & 64 \\
Sobol & 20 & .891\ci{.109} & $+.083$ & \dci{$+.007$}{$+.191$} & .100\ci{.125} & $+.620$ & \dci{$+.408$}{$+.815$} & \second{.046}\ci{.039} & $+.487$ & \dci{$+.306$}{$+.665$} & 3,936 \\
HSIC & 20 & \second{.940}\ci{.044} & $+.035$ & \dci{$+.014$}{$+.063$} & .050\ci{.075} & $+.670$ & \dci{$+.470$}{$+.857$} & .022\ci{.022} & $+.510$ & \dci{$+.326$}{$+.689$} & 1,536 \\
LIMA & \na & \multicolumn{9}{c}{\ctrl{n.r.$^{m,t}$}} & \ctrl{\na} \\
MoXI & 20 & .507\ci{.116} & $+.467$ & \dci{$+.353$}{$+.581$} & .100\ci{.125} & $+.620$ & \dci{$+.420$}{$+.818$} & .006\ci{.008} & $+.527$ & \dci{$+.340$}{$+.706$} & 19,306 \\
WCAM & 20 & .900\ci{.056} & $+.075$ & \dci{$+.040$}{$+.116$} & .000\ci{.000} & $+.720$ & \dci{$+.525$}{$+.890$} & .006\ci{.005} & $+.526$ & \dci{$+.338$}{$+.712$} & 6,288 \\
\tabgroup{11}{gradient- and activation-based}
Integrated Gradients & 20 & .494\ci{.012} & $+.481$ & \dci{$+.452$}{$+.505$} & .800\ci{.175} & $-.080$ & \dci{$-.250$}{$+.082$} & .195\ci{.080} & $+.337$ & \dci{$+.194$}{$+.475$} & 51 \\
Grad-CAM & 20 & .720\ci{.128} & $+.254$ & \dci{$+.135$}{$+.386$} & .050\ci{.075} & $+.670$ & \dci{$+.443$}{$+.875$} & .008\ci{.007} & $+.525$ & \dci{$+.334$}{$+.705$} & 1 \\
\midrule
\ctrl{Random map} & \ctrl{20} & \ctrl{.501}\ci{.004} & \ctrl{$+.474$} & \dci{$+.447$}{$+.493$} & \ctrl{.000}\ci{.000} & \ctrl{$+.720$} & \dci{$+.522$}{$+.893$} & \ctrl{.005}\ci{.007} & \ctrl{$+.528$} & \dci{$+.342$}{$+.710$} & \ctrl{0} \\
\ctrl{Centre map} & \ctrl{20} & \ctrl{.513}\ci{.113} & \ctrl{$+.462$} & \dci{$+.346$}{$+.582$} & \ctrl{.000}\ci{.000} & \ctrl{$+.720$} & \dci{$+.528$}{$+.890$} & \ctrl{.009}\ci{.012} & \ctrl{$+.524$} & \dci{$+.333$}{$+.707$} & \ctrl{0} \\
\midrule
\method & 20 & \best{.975}\ci{.023} & \multicolumn{2}{c}{\na} & \best{.720}\ci{.177} & \multicolumn{2}{c}{\na} & \best{.533}\ci{.188} & \multicolumn{2}{c}{\na} & 310 \\
\bottomrule
\end{tabular}
\end{suppwide}

%% file: supplementary/supp_E_limits.tex
\section{Limitations and failure cases}
\label{sec:supp_limits}
\label{sec:supp_failure}
The hierarchical search refines a region only when masking it changes the prediction score by more than the relative threshold (\cref{eq:open} of the main paper). Evidence whose individual removal produces little or no change in the output therefore escapes refinement, whereas structures sensitive to partial masking are still resolved.

\paragraph{Redundant evidence.}
When a model aggregates multiple copies of the target through an OR rule, removing a redundant copy leaves the remaining copies active and produces little or no change in the prediction score, so its node falls below the threshold of \cref{eq:open} and is not expanded. In \cref{fig:supp_failure}(a, b), only one copy triggers refinement while the remaining copies stay in coarse nodes, so \method recovers one of two and one of four copies. Redundancy affects every removal-based method: on Synthetic Dataset scenes at 2,048\,px with two or four copies of the target, explained for the 8 trained CNNs, \method, h-Shap, Occlusion and RISE each recover .29 to .70 of the copies, and \method keeps the highest mean AUROC (.974 and .942).

\begin{suppwide}{figure}
\centering
\includegraphics[width=\linewidth]{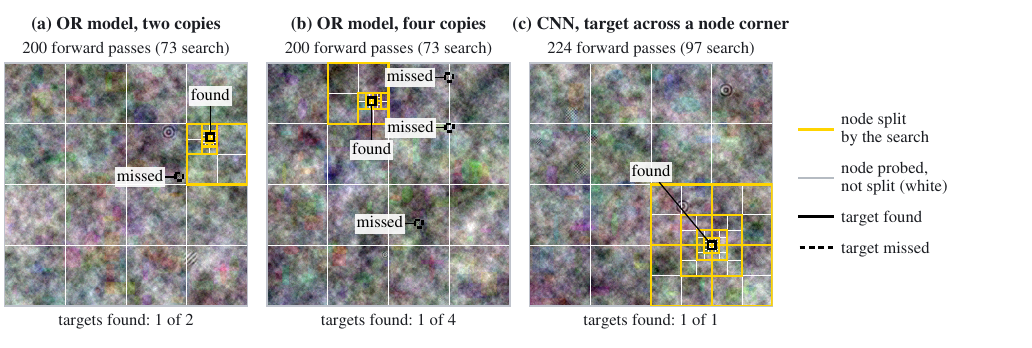}
\caption{\textbf{Redundant targets and a target across a node corner ($r>1$).} Synthetic Dataset, 1,024\,px. (a, b) OR model with two and four copies of the target; (c) trained CNN with a target centred on a node corner. Yellow: expanded nodes; white: evaluated nodes not expanded; solid: target found; dashed: target missed.}
\label{fig:supp_failure}
\end{suppwide}

\paragraph{Target across a node corner.}
Because the quadtree partitions along fixed dyadic boundaries, a target centred on the corner of four nodes is divided across four regions at every depth, which could reduce the score difference of each individual node. In \cref{fig:supp_failure}(c), however, masking part of the target still lowers the output of the trained CNN enough to exceed the relative threshold, so the search refines the nodes around the corner and resolves the target. Across the controlled scenes of \cref{sec:supp_search}, \method attains an object recall of 1.000 both when the target aligns with the quadtree grid and when it is centred on a four-node corner. A split target thus still changes the score when partly masked; a redundant copy does not, yet every removal-based search depends on such a change in the score.

%% file: supplementary/supp_F_tables.tex
\section{Complete tables}
\label{sec:supp_results}
\label{sec:supp_maps}
This section completes the tables of the main paper with 95\,\% bootstrap confidence intervals, Holm-corrected paired tests against the random map (\cref{sec:supp_stats}), and the five gradient- and activation-based methods as references.

\subsection{High-resolution images}
\label{sec:supp_all}
In \cref{tab:hr} of the main paper, log-scale Deletion for ShapBPT, Sobol, and HSIC on VisA and the CAMELYON16 crops is evaluated from attribution maps stored at a side of 256 to 330\,px (omitted from the per-side tables); on the same images, this storage agrees with native-resolution Deletion within $\pm$.001 for Occlusion and within $-.018$ (VisA) and $+.006$ (crops) for RISE, whereas it distorts smooth maps on Real-IAD ($+.16$ for RISE), so those Real-IAD cells are omitted. \Cref{tab:supp_hr_full} gives an interval for every entry of \cref{tab:hr} (slide times of RISE, LIME, Sobol and WCAM projected; slide AUROC of RISE and LIME with 256 masks).

\input{tables/S_hr_full}

\paragraph{The same image sides.}
Slower methods stop at a time or memory limit at the largest sides, so the methods of \cref{tab:hr} do not all explain the same image sides. \Cref{tab:supp_same_sides} compares \method with each perturbation-based method on the pairs of image and side that both explained: \method has the higher EPG in all 36 comparisons (Holm $p\le.0012$ in 35) and the higher Pointing Game in 34 (Holm $p\le.014$).

\begin{suppwide}{table}
\centering
\tabstyle
\caption{\textbf{\method against each perturbation-based method on the same image sides.} Paired difference \method $-$ method ($+$: \method is better) with 95\,\% bootstrap CI over pairs of image and side; $n$: pairs with EPG; $^{*}$: not significant (two-sided paired test, Holm correction over the 12 methods).}
\label{tab:supp_same_sides}
\footnotesize
\setlength{\tabcolsep}{1.8pt}
\begin{suppfit}
\begin{tabular}{@{}l rll rll rll@{}}
\toprule
 & \multicolumn{3}{c}{\tabhead{VisA}} & \multicolumn{3}{c}{\tabhead{Real-IAD}} & \multicolumn{3}{c}{\tabhead{CAMELYON16 crops}} \\
\cmidrule(lr){2-4}\cmidrule(lr){5-7}\cmidrule(l){8-10}
\tabhead{Method} & \tabhead{$n$} & \tabhead{EPG $\uparrow$} & \tabhead{Pointing Game $\uparrow$} & \tabhead{$n$} & \tabhead{EPG $\uparrow$} & \tabhead{Pointing Game $\uparrow$} & \tabhead{$n$} & \tabhead{EPG $\uparrow$} & \tabhead{Pointing Game $\uparrow$} \\
\midrule
\csname @@input\endcsname tables/S_same_sides.tex %
\bottomrule
\end{tabular}
\end{suppfit}
\end{suppwide}

\subsection{ImageNet and GridPG}
\label{sec:supp_inet}
\paragraph{ImageNet across four architectures.}
The intervals for both parts of \cref{tab:std} of the main paper are in \cref{tab:supp_imagenet_cnn,tab:supp_gridpg,tab:supp_ablation_full}. \Cref{tab:supp_imagenet_cnn,tab:supp_imagenet_vit} extend \cref{tab:std} of the main paper to ViT-B/16 and DeiT-B/16 and add Pointing Game and EPG localisation on ImageNet-S50 (247 of the 250 images, excluding two masks that cover the full image and one that is empty after the centre crop), alongside $\mu$Fidelity (200 samples per map, \cref{sec:supp_metrics}). \method explains all 1,000 ImageNet images and all 247 ImageNet-S50 images with all four models. On ViT-B/16, it has the lowest mean Deletion and the highest mean $\mu$Fidelity among the perturbation-based methods, and on DeiT-B/16 the highest mean Insertion. Methods marked \sa\ use reduced subsets: 61 ImageNet-S50 images for localisation, and the first 250 of the 1,000 images for Deletion, Insertion and $\mu$Fidelity, whose values on all 1,000 images are in the lower group of \cref{tab:supp_imagenet_cnn} for ResNet-50 and VGG16; where baseline runs reached memory limits on individual images, averages are over the completed images (550 to 1,000 of 1,000, and 127 to 250 on reduced subsets).

\begin{suppwide}{table}
\centering
\tabstyle\renewcommand{\arraystretch}{.95}
\caption{\textbf{ImageNet, ResNet-50 and VGG16.} Deletion, Insertion and $\mu$Fidelity on ImageNet; Pointing Game and EPG on ImageNet-S50. \sa: reduced subset; lower group: these methods on all 1,000 images. $\pm$: 95\,\% CI; $^{\dagger}$\,better than the random map (one-sided paired test, Holm correction); bold/underline: best/second among perturbation-based methods; --: not computed; n.r.$^{m,t}$: memory or time limit.}
\label{tab:supp_imagenet_cnn}
\footnotesize
\setlength{\tabcolsep}{2pt}
\begin{tabular}{@{}l cccccccccc@{}}
\toprule
 & \multicolumn{5}{c}{\tabhead{ResNet-50}} & \multicolumn{5}{c}{\tabhead{VGG16}} \\
\cmidrule(lr){2-6}\cmidrule(lr){7-11}
\tabhead{Method} & \tabhead{Deletion $\downarrow$} & \tabhead{Insertion $\uparrow$} & \tabhead{$\mu$Fidelity $\uparrow$} & \tabhead{\shortstack{Pointing\\Game $\uparrow$}} & \tabhead{EPG $\uparrow$} & \tabhead{Deletion $\downarrow$} & \tabhead{Insertion $\uparrow$} & \tabhead{$\mu$Fidelity $\uparrow$} & \tabhead{\shortstack{Pointing\\Game $\uparrow$}} & \tabhead{EPG $\uparrow$} \\
\midrule
\csname @@input\endcsname tables/A5a_imagenet_cnn_body.tex %
\csname @@input\endcsname tables/A5a_full_rows_487.tex %
\bottomrule
\end{tabular}
\end{suppwide}

\begin{suppwide}{table}
\centering
\tabstyle\renewcommand{\arraystretch}{.95}
\caption{\textbf{ImageNet, ViT-B/16 and DeiT-B/16.} Notation as in \cref{tab:supp_imagenet_cnn}.}
\label{tab:supp_imagenet_vit}
\footnotesize
\setlength{\tabcolsep}{2pt}
\begin{tabular}{@{}l cccccccccc@{}}
\toprule
 & \multicolumn{5}{c}{\tabhead{ViT-B/16}} & \multicolumn{5}{c}{\tabhead{DeiT-B/16}} \\
\cmidrule(lr){2-6}\cmidrule(lr){7-11}
\tabhead{Method} & \tabhead{Deletion $\downarrow$} & \tabhead{Insertion $\uparrow$} & \tabhead{$\mu$Fidelity $\uparrow$} & \tabhead{\shortstack{Pointing\\Game $\uparrow$}} & \tabhead{EPG $\uparrow$} & \tabhead{Deletion $\downarrow$} & \tabhead{Insertion $\uparrow$} & \tabhead{$\mu$Fidelity $\uparrow$} & \tabhead{\shortstack{Pointing\\Game $\uparrow$}} & \tabhead{EPG $\uparrow$} \\
\midrule
\csname @@input\endcsname tables/A5b_imagenet_vit_body.tex %
\bottomrule
\end{tabular}
\end{suppwide}

\paragraph{GridPG per model.}
\Cref{tab:supp_gridpg} gives the pooled GridPG share of \cref{tab:std} for ResNet-50 and VGG16 separately. Methods marked \sa\ use 100 of the 500 grids per model.

\begin{suppwide}{table}
\centering
\tabstyle\renewcommand{\arraystretch}{.98}
\caption{\textbf{GridPG per model.} Share of positive attribution in the target cell ($\uparrow$). Pooled: ResNet-50 and VGG16; notation as in \cref{tab:supp_imagenet_cnn}.}
\label{tab:supp_gridpg}
\footnotesize
\setlength{\tabcolsep}{3pt}
\begin{tabular}{@{}l ccc @{\hspace{18pt}} c @{\hspace{0pt}} l ccc@{}}
\toprule
\tabhead{Method} & \tabhead{ResNet-50} & \tabhead{VGG16} & \tabhead{Pooled} & & \tabhead{Method} & \tabhead{ResNet-50} & \tabhead{VGG16} & \tabhead{Pooled} \\
\midrule
\csname @@input\endcsname tables/G507_gridpg_2up.tex %
\bottomrule
\end{tabular}
\end{suppwide}

\subsection{High-resolution images per side}
\label{sec:supp_side}
\Cref{tab:supp_side} gives VisA and Real-IAD for each image side $L$.

\begin{suppwide}{table}
\centering
\tabstyle\renewcommand{\arraystretch}{.98}
\caption{\textbf{VisA and Real-IAD per image side.} $L$: image side (px); AUROC: window AUROC. $\pm$: 95\,\% CI; $^{\dagger}$\,better than the random map (one-sided paired test, Holm correction); bold/underline: best/second; \sa: pre-selected subset; n.r.$^{m,t}$: memory or time limit; --: not computed (map stored at a reduced side).}
\label{tab:supp_side}
\footnotesize
\setlength{\tabcolsep}{1pt}
\begin{tabular}{@{}l cccccccccccc@{}}
\toprule
\multicolumn{13}{@{}l}{\tabhead{\textbf{VisA, 512 to 1,024\,px}}} \\
 & \multicolumn{3}{c}{\tabhead{$L=512$}} & \multicolumn{3}{c}{\tabhead{$L=960$}} & \multicolumn{3}{c}{\tabhead{$L=992$}} & \multicolumn{3}{c}{\tabhead{$L=1024$}} \\
\cmidrule(lr){2-4}\cmidrule(lr){5-7}\cmidrule(lr){8-10}\cmidrule(lr){11-13}
\tabhead{Method} & \tabhead{AUROC $\uparrow$} & \tabhead{EPG $\uparrow$} & \tabhead{\shortstack{log-scale\\Deletion $\downarrow$}} & \tabhead{AUROC $\uparrow$} & \tabhead{EPG $\uparrow$} & \tabhead{\shortstack{log-scale\\Deletion $\downarrow$}} & \tabhead{AUROC $\uparrow$} & \tabhead{EPG $\uparrow$} & \tabhead{\shortstack{log-scale\\Deletion $\downarrow$}} & \tabhead{AUROC $\uparrow$} & \tabhead{EPG $\uparrow$} & \tabhead{\shortstack{log-scale\\Deletion $\downarrow$}} \\
\midrule
\csname @@input\endcsname tables/G503_visa_a.tex %
\midrule
\multicolumn{13}{@{}l}{\tabhead{\textbf{VisA, 1,056 to 1,168\,px}}} \\
 & \multicolumn{3}{c}{\tabhead{$L=1056$}} & \multicolumn{3}{c}{\tabhead{$L=1104$}} & \multicolumn{3}{c}{\tabhead{$L=1152$}} & \multicolumn{3}{c}{\tabhead{$L=1168$}} \\
\cmidrule(lr){2-4}\cmidrule(lr){5-7}\cmidrule(lr){8-10}\cmidrule(lr){11-13}
\tabhead{Method} & \tabhead{AUROC $\uparrow$} & \tabhead{EPG $\uparrow$} & \tabhead{\shortstack{log-scale\\Deletion $\downarrow$}} & \tabhead{AUROC $\uparrow$} & \tabhead{EPG $\uparrow$} & \tabhead{\shortstack{log-scale\\Deletion $\downarrow$}} & \tabhead{AUROC $\uparrow$} & \tabhead{EPG $\uparrow$} & \tabhead{\shortstack{log-scale\\Deletion $\downarrow$}} & \tabhead{AUROC $\uparrow$} & \tabhead{EPG $\uparrow$} & \tabhead{\shortstack{log-scale\\Deletion $\downarrow$}} \\
\midrule
\csname @@input\endcsname tables/G503_visa_b.tex %
\midrule
\multicolumn{13}{@{}l}{\tabhead{\textbf{Real-IAD}}} \\
 & \multicolumn{3}{c}{\tabhead{$L=512$}} & \multicolumn{3}{c}{\tabhead{$L=1024$}} & \multicolumn{3}{c}{\tabhead{$L=2048$}} & \multicolumn{3}{c}{\tabhead{$L=2992$}} \\
\cmidrule(lr){2-4}\cmidrule(lr){5-7}\cmidrule(lr){8-10}\cmidrule(lr){11-13}
\tabhead{Method} & \tabhead{AUROC $\uparrow$} & \tabhead{EPG $\uparrow$} & \tabhead{\shortstack{log-scale\\Deletion $\downarrow$}} & \tabhead{AUROC $\uparrow$} & \tabhead{EPG $\uparrow$} & \tabhead{\shortstack{log-scale\\Deletion $\downarrow$}} & \tabhead{AUROC $\uparrow$} & \tabhead{EPG $\uparrow$} & \tabhead{\shortstack{log-scale\\Deletion $\downarrow$}} & \tabhead{AUROC $\uparrow$} & \tabhead{EPG $\uparrow$} & \tabhead{\shortstack{log-scale\\Deletion $\downarrow$}} \\
\midrule
\csname @@input\endcsname tables/G503_realiad.tex %
\bottomrule
\end{tabular}
\end{suppwide}

\noindent \Cref{fig:supp_side} details \cref{fig:side} of the main paper: it gives the Pointing Game, window AUROC and EPG of every method per image side on the CAMELYON16 crops and Real-IAD.

\begin{suppwide}{figure}
  \centering
  \includegraphics[width=\linewidth]{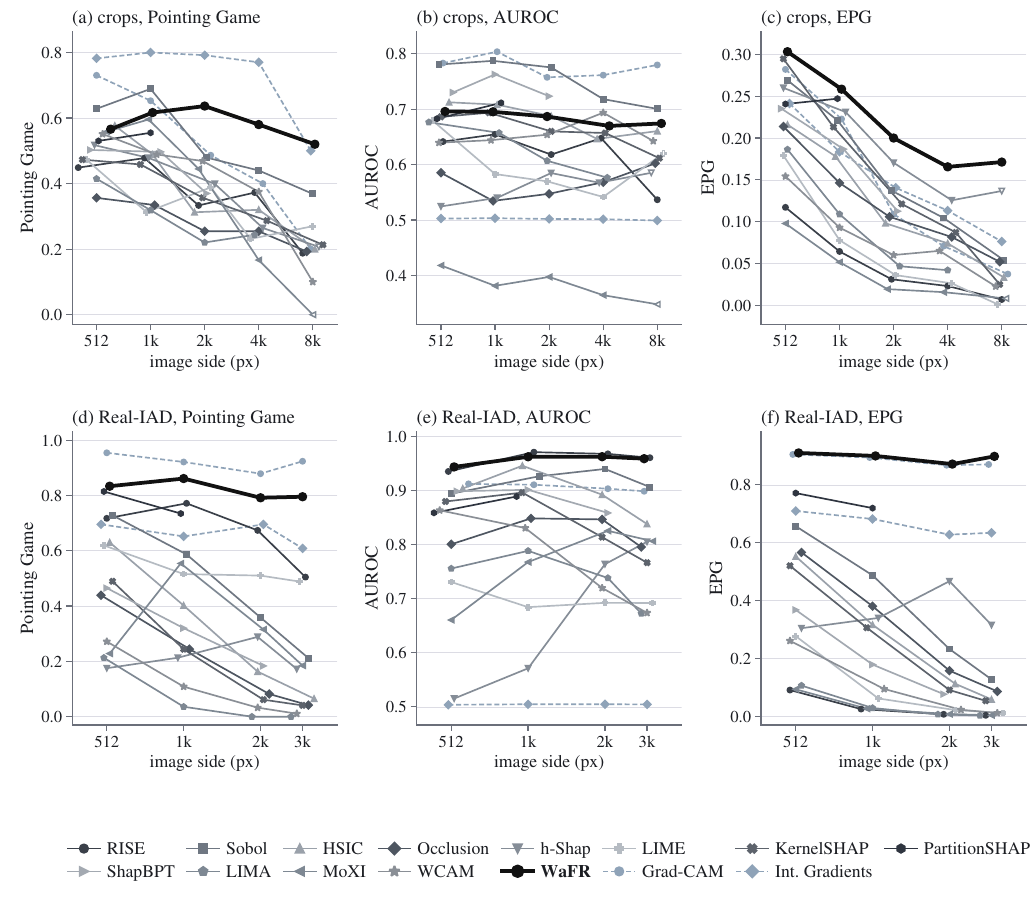}
  \caption{\textbf{Localisation metrics across image sides for individual methods.} Dashed: gradient- and activation-based references; a curve ends at a time or memory limit.}
  \label{fig:supp_side}
\end{suppwide}

\subsection{Synthetic Dataset}
\label{sec:supp_synth}
On the Synthetic Dataset (\cref{tab:supp_synth}), RISE uses 4,096 masks at every side, 21 to 24 times the median number of forward passes of \method (168 to 196 per scene), and takes about 360 times longer than \method on the timed 2,048\,px scene (\cref{tab:supp_cost}). \Cref{tab:equal} of the main paper compares the methods at equal numbers of forward passes.

\begin{suppwide}{table}
\centering
\tabstyle\renewcommand{\arraystretch}{1.02}
\caption{\textbf{Synthetic Dataset, window AUROC ($\uparrow$).} One group per model and target side, one row per image side $L$ (px). LSE: log-sum-exp pooling; receptive field 128\,px unless large; $s$: target side. Bold/underline: best/second per row; other notation as in \cref{tab:supp_side}.}
\label{tab:supp_synth}
\footnotesize
\setlength{\tabcolsep}{1.5pt}
\begin{tabular}{@{}r *{13}{c}@{}}
\toprule
\tabhead{$L$} & \tabhead{Occlusion} & \tabhead{RISE\sa} & \tabhead{LIME\sa} & \tabhead{\shortstack{Kernel-\\SHAP}} & \tabhead{\shortstack{Partition-\\SHAP\sa}} & \tabhead{h-Shap} & \tabhead{ShapBPT} & \tabhead{Sobol\sa} & \tabhead{HSIC} & \tabhead{LIMA\sa} & \tabhead{MoXI\sa} & \tabhead{WCAM\sa} & \tabhead{\method} \\
\midrule
\csname @@input\endcsname tables/G507_synth_t.tex %
\bottomrule
\end{tabular}
\end{suppwide}

\subsection{Whole slides}
\label{sec:supp_slides}
On the 45 CAMELYON16 tumour slides with at least one patch of at least 50\,\% tumour, \method uses a median of 179 forward passes per slide (range 160 to 667) and has a Pointing Game of .889 (40 of 45 slides). Its patch-level AUROC is higher than that of Occlusion, RISE, LIME, KernelSHAP, PartitionSHAP, HSIC, LIMA and MoXI on at least 39 of the 45 slides (KernelSHAP: 36 of 38; Holm $p<.001$ for each) and than that of h-Shap on 24 of the 31 slides that h-Shap completes within 40,000 forward passes. Its mean slide Deletion, .522, is lower than that of each of the nine perturbation-based methods (Holm $p<.001$ for eight) and than a random patch order. \Cref{tab:supp_slides} gives the patch-level AUROC on these 45 slides and the slide Deletion on the 40 slides classified as tumour by DSMIL; each method is paired with \method on their common slides.

\input{tables/S_slides}

\section{Attribution galleries}
\label{sec:supp_gallery}
\Cref{fig:supp_gal_a,fig:supp_gal_b,fig:supp_gal_crops,fig:supp_gal_synth,fig:supp_gal_yolo} show examples of the attribution maps of \method and of every method of the tables, from the same runs as the tables. Each map shows the rank of its pixels, and only the top 10\,\% of pixels are coloured; a dashed outline marks the ground truth, a filled dot (open circle) the top-1 pixel inside (outside) it, a black frame \method, and the number below a map the forward passes of that run; n.r.\ denotes a method with no stored map for that instance.

\begin{suppwide}{figure}
\centering
\includegraphics[width=\linewidth]{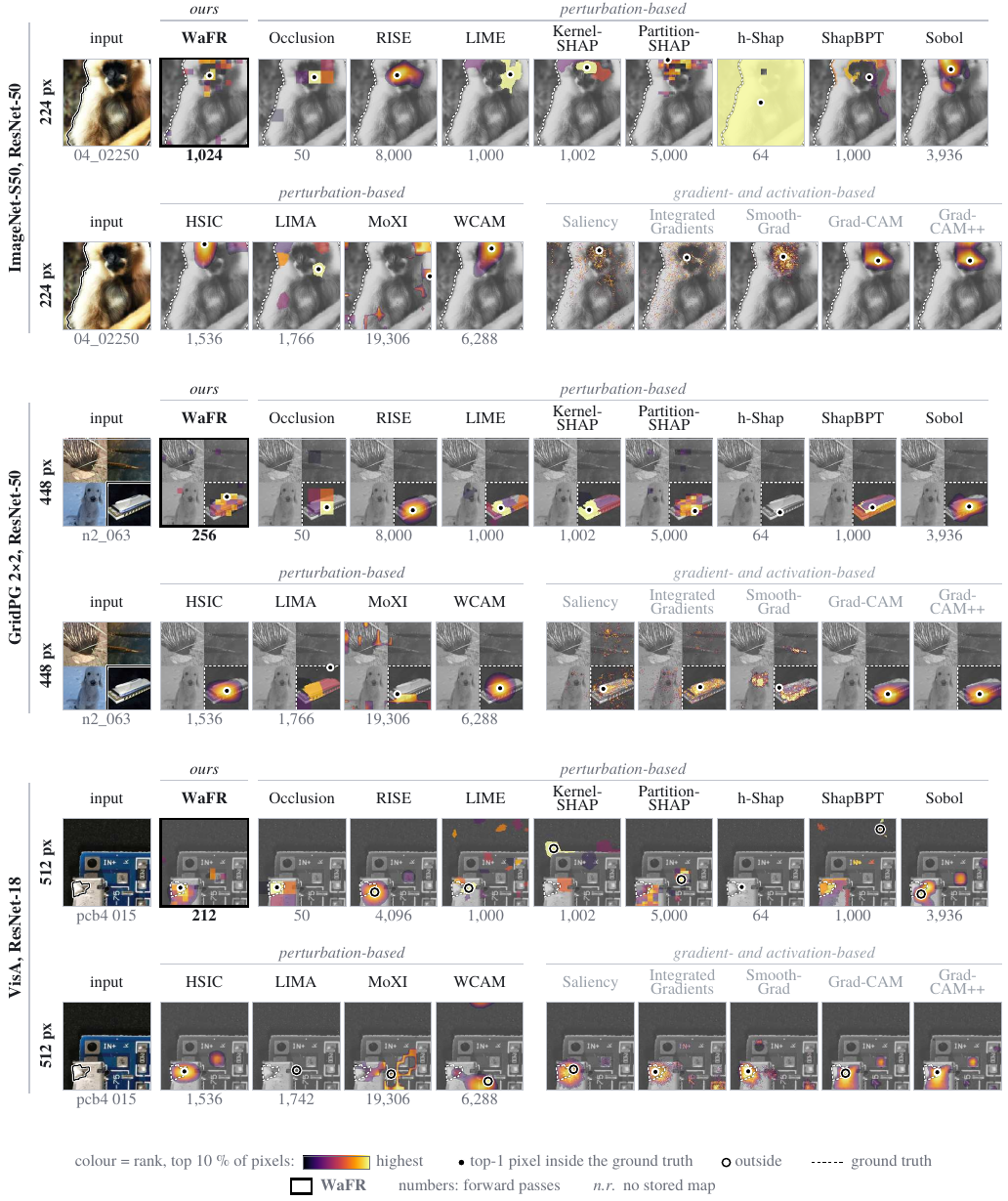}
\caption{\textbf{Attribution maps: ImageNet-S50 and GridPG $2\times2$ with ResNet-50, VisA with ResNet-18.} GridPG outline: target cell. Symbols in \cref{sec:supp_gallery}.}
\label{fig:supp_gal_a}
\end{suppwide}

\begin{suppwide}{figure}
\centering
\includegraphics[width=\linewidth]{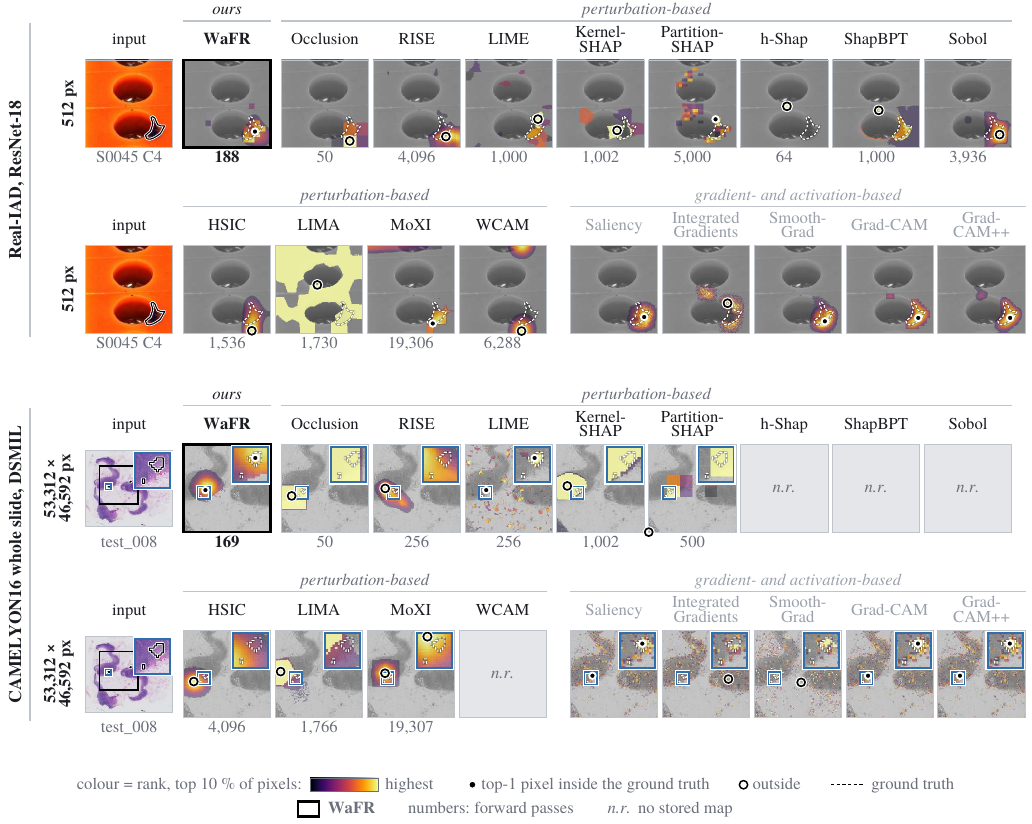}
\caption{\textbf{Attribution maps: Real-IAD with ResNet-18, CAMELYON16 whole slide with DSMIL.} Blue boxes: zoom on the target region. Symbols in \cref{sec:supp_gallery}.}
\label{fig:supp_gal_b}
\end{suppwide}

\begin{suppwide}{figure}
\centering
\includegraphics[width=\linewidth]{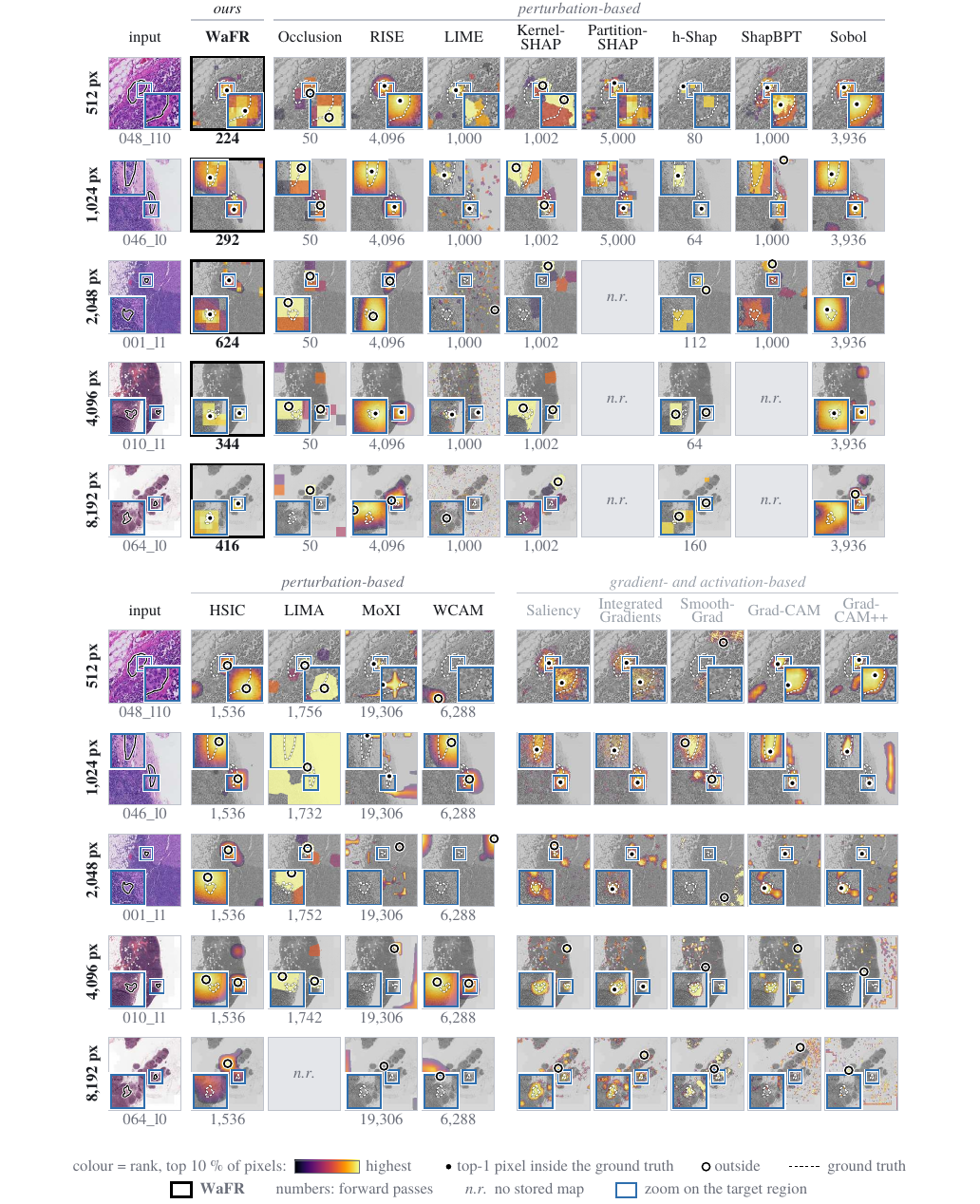}
\caption{\textbf{Attribution maps: CAMELYON16 crops, ResNet-18 with attention-based MIL.} Blue boxes: zoom on the target region. Symbols in \cref{sec:supp_gallery}.}
\label{fig:supp_gal_crops}
\end{suppwide}

\begin{suppwide}{figure}
\centering
\includegraphics[width=\linewidth]{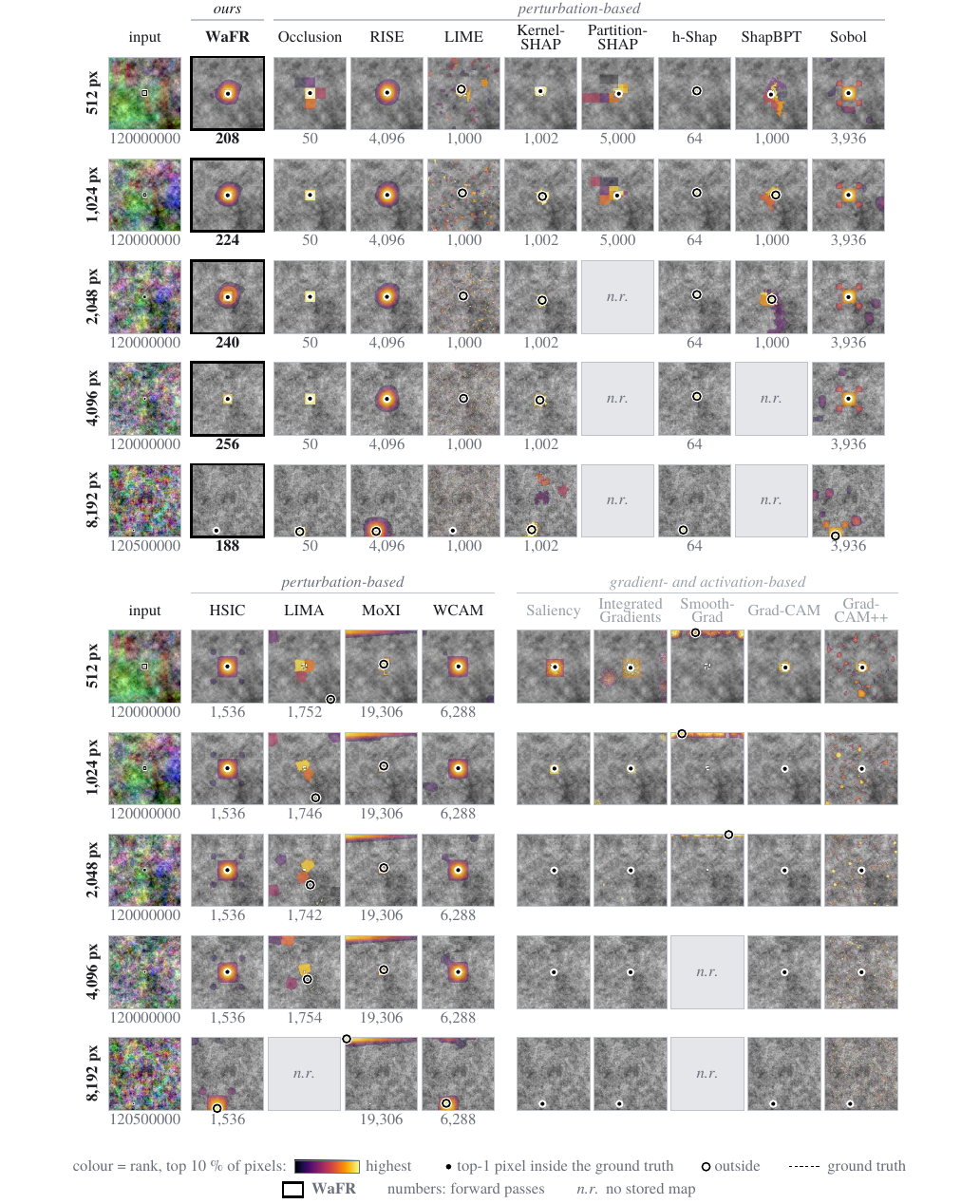}
\caption{\textbf{Attribution maps: Synthetic Dataset, max-pooling CNN (receptive field 128\,px, target 32\,px).} Symbols in \cref{sec:supp_gallery}.}
\label{fig:supp_gal_synth}
\end{suppwide}

\begin{suppwide}{figure}
\centering
\includegraphics[width=\linewidth]{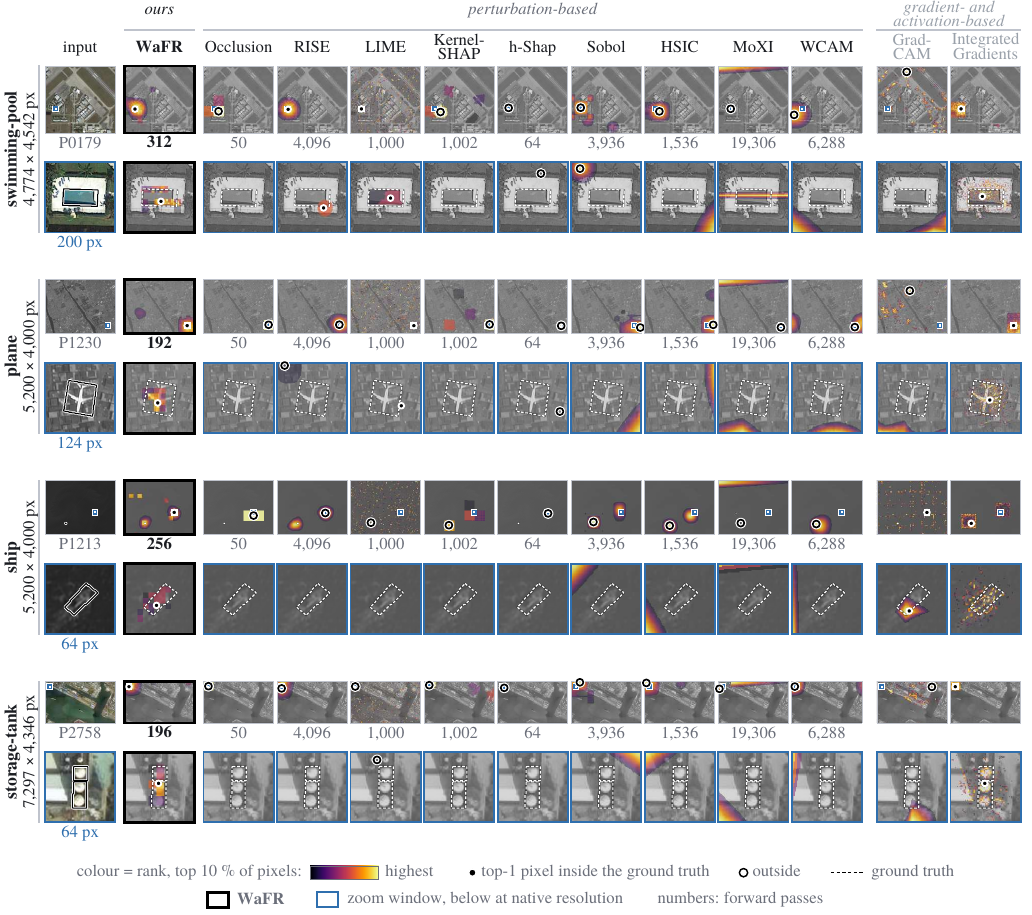}
\caption{\textbf{Attribution maps: DOTA with YOLO11x-OBB.} Blue boxes: zoom on the target object, coloured by the rank of each pixel within the zoom window. Symbols in \cref{sec:supp_gallery}.}
\label{fig:supp_gal_yolo}
\end{suppwide}

%% file: tables/S_hr_full.tex
\begin{suppwide}{table}
\centering
\tabstyle\renewcommand{\arraystretch}{0.94}%
\caption{\textbf{Real high-resolution images and whole slides, with all intervals.} VisA, Real-IAD, crops: mean over image sides. AUROC: in a window around the target; EPG: energy-based Pointing Game; slides: patch-level AUROC and time on one slide; last column: forward passes for patches of a predefined size (none: no such rule; $k$: relevant nodes per depth, up to $L^2$). $\pm$: 95\,\% CI; $^{*}$: not statistically different from \method (paired test, Holm correction). Bold: best; underline: second; grey: random and centre maps; --: not computed.}
\label{tab:supp_hr_full}
\footnotesize
\setlength{\tabcolsep}{1.2pt}
\begin{tabular}{@{}l ccc ccc ccc cc c@{}}
\toprule
 & \multicolumn{3}{c}{\tabhead{VisA}} & \multicolumn{3}{c}{\tabhead{Real-IAD}} & \multicolumn{3}{c}{\tabhead{CAMELYON16 crops}} & \multicolumn{2}{c}{\tabhead{CAMELYON16 slides}} & \\
\cmidrule(lr){2-4}\cmidrule(lr){5-7}\cmidrule(lr){8-10}\cmidrule(lr){11-12}
\tabhead{Method} & \tabhead{AUROC $\uparrow$} & \tabhead{EPG $\uparrow$} & \tabhead{\shortstack{log-scale\\Deletion $\downarrow$}} & \tabhead{AUROC $\uparrow$} & \tabhead{EPG $\uparrow$} & \tabhead{\shortstack{log-scale\\Deletion $\downarrow$}} & \tabhead{AUROC $\uparrow$} & \tabhead{EPG $\uparrow$} & \tabhead{\shortstack{log-scale\\Deletion $\downarrow$}} & \tabhead{AUROC $\uparrow$} & \tabhead{\shortstack{Execution\\time (s) $\downarrow$}} & \tabhead{\shortstack{Forward\\passes at\\predefined\\patch size}} \\
\midrule
\csname @@input\endcsname tables/T2_hr_body.tex %
\bottomrule
\end{tabular}
\end{suppwide}

%% file: tables/S_slides.tex
\begin{suppwide}{table}
\centering
\tabstyle
\caption{\textbf{Whole slides, per method.} Left: patch-level AUROC ($\uparrow$); right: slide Deletion ($\downarrow$). $n$: slides; $\Delta$: paired difference ($+$: \method is better) with 95\,\% bootstrap CI over slides; $p$: two-sided paired test, Holm correction over the nine perturbation-based methods; gradient- and activation-based methods: uncorrected (left), Holm correction among them (right); references: uncorrected; wins: slides on which \method is higher. n.r.: no map. Random map\,/\,order: random map (left), random patch order (right). Slide Deletion: slides classified as tumour by DSMIL. Fewer slides: KernelSHAP, design limit on 7 slides; h-Shap, time or forward-pass limit on 14; SmoothGrad and Grad-CAM++, memory limit on 1; Integrated Gradients, the 10 tumour slides of the 20-slide cohort (\cref{sec:supp_setup}).}%

\label{tab:supp_slides}
\footnotesize
\setlength{\tabcolsep}{3.2pt}
\begin{tabular}{@{}l r c r@{}l c c @{\hspace{10pt}} r r@{}l c@{}}
\toprule
 & \multicolumn{6}{c}{\tabhead{Patch-level AUROC $\uparrow$}} & \multicolumn{4}{c}{\tabhead{Slide Deletion $\downarrow$}} \\
\cmidrule(lr){2-7}\cmidrule(l){8-11}
\tabhead{Method} & \tabhead{$n$} & \tabhead{mean} & \multicolumn{2}{c}{\tabhead{$\Delta$}} & \tabhead{$p$} & \tabhead{wins} & \tabhead{$n$} & \multicolumn{2}{c}{\tabhead{$\Delta$}} & \tabhead{$p$} \\
\midrule
\method & 45 & .949 & \multicolumn{2}{c}{\na} & \na & \na & 40 & \multicolumn{2}{c}{\na} & \na \\
\midrule
Occlusion & 45 & .625 & $+.324$ & \dci{$+.269$}{$+.382$} & $<$.001 & 45/45 & 40 & $+.033$ & \dci{$+.021$}{$+.047$} & $<$.001 \\
RISE & 45 & .608 & $+.340$ & \dci{$+.295$}{$+.386$} & $<$.001 & 45/45 & 40 & $+.051$ & \dci{$+.034$}{$+.070$} & $<$.001 \\
LIME & 45 & .744 & $+.204$ & \dci{$+.149$}{$+.261$} & $<$.001 & 39/45 & 40 & $+.011$ & \dci{$+.001$}{$+.022$} & .062 \\
KernelSHAP & 38 & .778 & $+.167$ & \dci{$+.120$}{$+.219$} & $<$.001 & 36/38 & 34 & $+.040$ & \dci{$+.025$}{$+.056$} & $<$.001 \\
PartitionSHAP & 45 & .736 & $+.213$ & \dci{$+.181$}{$+.248$} & $<$.001 & 45/45 & 40 & $+.046$ & \dci{$+.033$}{$+.061$} & $<$.001 \\
h-Shap & 31 & .922 & $+.025$ & \dci{$-.003$}{$+.049$} & .071 & 24/31 & 27 & $+.057$ & \dci{$+.039$}{$+.077$} & $<$.001 \\
ShapBPT & \multicolumn{6}{c}{\ctrl{n.r.}} & \multicolumn{4}{c}{\ctrl{n.r.}} \\
Sobol & \multicolumn{6}{c}{\ctrl{n.r.}} & \multicolumn{4}{c}{\ctrl{n.r.}} \\
HSIC & 45 & .797 & $+.152$ & \dci{$+.110$}{$+.197$} & $<$.001 & 40/45 & 40 & $+.026$ & \dci{$+.015$}{$+.039$} & $<$.001 \\
LIMA & 45 & .725 & $+.224$ & \dci{$+.172$}{$+.278$} & $<$.001 & 42/45 & 40 & $+.045$ & \dci{$+.030$}{$+.062$} & $<$.001 \\
MoXI & 45 & .685 & $+.264$ & \dci{$+.218$}{$+.313$} & $<$.001 & 44/45 & 40 & $+.019$ & \dci{$+.012$}{$+.028$} & $<$.001 \\
WCAM & \multicolumn{6}{c}{\ctrl{n.r.}} & \multicolumn{4}{c}{\ctrl{n.r.}} \\
\midrule
Saliency & 45 & .994 & $-.045$ & \dci{$-.063$}{$-.030$} & $<$.001 & 1/45 & 40 & $-.084$ & \dci{$-.103$}{$-.065$} & $<$.001 \\
Integrated Gradients & 10 & .392 & $+.556$ & \dci{$+.404$}{$+.709$} & .002 & 10/10 & 9 & $+.065$ & \dci{$+.016$}{$+.119$} & .10 \\
SmoothGrad & 44 & .943 & $+.008$ & \dci{$-.014$}{$+.032$} & .53 & 22/44 & 39 & $+.009$ & \dci{$-.003$}{$+.023$} & .15 \\
Grad-CAM & 45 & .993 & $-.045$ & \dci{$-.062$}{$-.030$} & $<$.001 & 3/45 & 40 & $-.091$ & \dci{$-.109$}{$-.072$} & $<$.001 \\
Grad-CAM++ & 44 & .994 & $-.043$ & \dci{$-.061$}{$-.028$} & $<$.001 & 3/44 & 39 & $-.094$ & \dci{$-.113$}{$-.075$} & $<$.001 \\
\midrule
\ctrl{Random map\,/\,order} & \ctrl{45} & \ctrl{.509} & \ctrl{$+.440$} & \dci{$+.403$}{$+.477$} & \ctrl{$<$.001} & \ctrl{45/45} & \ctrl{40} & \ctrl{$+.060$} & \dci{$+.042$}{$+.080$} & \ctrl{$<$.001} \\
\ctrl{Centre map} & \ctrl{45} & \ctrl{.564} & \ctrl{$+.384$} & \dci{$+.335$}{$+.432$} & \ctrl{$<$.001} & \ctrl{43/45} & \ctrl{40} & \ctrl{$+.060$} & \dci{$+.041$}{$+.079$} & \ctrl{$<$.001} \\
\bottomrule
\end{tabular}
\end{suppwide}

%% file: main.bbl
\begin{thebibliography}{61}
\providecommand{\natexlab}[1]{#1}
\providecommand{\url}[1]{\texttt{#1}}
\expandafter\ifx\csname urlstyle\endcsname\relax
  \providecommand{\doi}[1]{doi: #1}\else
  \providecommand{\doi}{doi: \begingroup \urlstyle{rm}\Url}\fi

\bibitem[Achanta et~al.(2012)Achanta, Shaji, Smith, Lucchi, Fua, and S{\"{u}}sstrunk]{achanta2012slic}
Radhakrishna Achanta, Appu Shaji, Kevin Smith, Aur{\'{e}}lien Lucchi, Pascal Fua, and Sabine S{\"{u}}sstrunk.
\newblock {SLIC} superpixels compared to state-of-the-art superpixel methods.
\newblock \emph{{IEEE} Trans. Pattern Anal. Mach. Intell.}, 34\penalty0 (11):\penalty0 2274--2282, 2012.

\bibitem[Bergmann et~al.(2019)Bergmann, Fauser, Sattlegger, and Steger]{bergmann2019mvtec}
Paul Bergmann, Michael Fauser, David Sattlegger, and Carsten Steger.
\newblock {MVTec AD} -- {A} comprehensive real-world dataset for unsupervised anomaly detection.
\newblock In \emph{{CVPR}}, pages 9592--9600. Computer Vision Foundation / {IEEE}, 2019.

\bibitem[Bhatt et~al.(2020)Bhatt, Weller, and Moura]{bhatt2020evaluating}
Umang Bhatt, Adrian Weller, and José M.~F. Moura.
\newblock Evaluating and aggregating feature-based model explanations.
\newblock In \emph{{IJCAI}}, pages 3016--3022. ijcai.org, 2020.

\bibitem[Bándi et~al.(2019)Bándi, Geessink, Manson, Dijk, Balkenhol, Hermsen, Bejnordi, Lee, Paeng, Zhong, Li, Zanjani, Zinger, Fukuta, Komura, Ovtcharov, Cheng, Zeng, Thagaard, Dahl, Lin, Chen, Jacobsson, Hedlund, Çetin, Halici, Jackson, Chen, Both, Franke, Küsters-Vandevelde, Vreuls, Bult, van Ginneken, van~der Laak, and Litjens]{bandi2019camelyon17}
Péter Bándi, Oscar Geessink, Quirine Manson, Marcory~Van Dijk, Maschenka Balkenhol, Meyke Hermsen, Babak~Ehteshami Bejnordi, Byungjae Lee, Kyunghyun Paeng, Aoxiao Zhong, Quanzheng Li, Farhad~Ghazvinian Zanjani, Svitlana Zinger, Keisuke Fukuta, Daisuke Komura, Vlado Ovtcharov, Shenghua Cheng, Shaoqun Zeng, Jeppe Thagaard, Anders~B. Dahl, Huangjing Lin, Hao Chen, Ludwig Jacobsson, Martin Hedlund, Melih Çetin, Eren Halici, Hunter Jackson, Richard Chen, Fabian Both, Jörg Franke, Heidi Küsters-Vandevelde, Willem Vreuls, Peter Bult, Bram van Ginneken, Jeroen van~der Laak, and Geert Litjens.
\newblock From detection of individual metastases to classification of lymph node status at the patient level: The {CAMELYON17} challenge.
\newblock \emph{IEEE Trans. Medical Imaging}, 38\penalty0 (2):\penalty0 550--560, 2019.

\bibitem[Böhle et~al.(2021)Böhle, Fritz, and Schiele]{bohle2021coda}
Moritz Böhle, Mario Fritz, and Bernt Schiele.
\newblock Convolutional dynamic alignment networks for interpretable classifications.
\newblock In \emph{{CVPR}}, pages 10029--10038. Computer Vision Foundation / IEEE, 2021.

\bibitem[Campanella et~al.(2019)Campanella, Hanna, Geneslaw, Miraflor, Silva, Busam, Brogi, Reuter, Klimstra, and Fuchs]{campanella2019clinical}
Gabriele Campanella, Matthew~G. Hanna, Luke Geneslaw, Allen Miraflor, Vitor Werneck~Krauss Silva, Klaus~J. Busam, Edi Brogi, Victor~E. Reuter, David~S. Klimstra, and Thomas~J. Fuchs.
\newblock Clinical-grade computational pathology using weakly supervised deep learning on whole slide images.
\newblock \emph{Nature Medicine}, 25\penalty0 (8):\penalty0 1301--1309, 2019.

\bibitem[Chattopadhyay et~al.(2018)Chattopadhyay, Sarkar, Howlader, and Balasubramanian]{chattopadhay2018gradcampp}
Aditya Chattopadhyay, Anirban Sarkar, Prantik Howlader, and Vineeth~N. Balasubramanian.
\newblock Grad-cam++: Generalized gradient-based visual explanations for deep convolutional networks.
\newblock In \emph{{WACV}}, pages 839--847. IEEE Computer Society, 2018.

\bibitem[Chen et~al.(2025)Chen, Liang, Li, Liu, Liu, Zhang, and Cao]{chen2025lima}
Ruoyu Chen, Siyuan Liang, Jingzhi Li, Shiming Liu, Li Liu, Hua Zhang, and Xiaochun Cao.
\newblock Less is more: Efficient black-box attribution via minimal interpretable subset selection.
\newblock \emph{CoRR}, abs/2504.00470, 2025.

\bibitem[Chen et~al.(2022)Chen, Chen, Li, Chen, Trister, Krishnan, and Mahmood]{chen2022hipt}
Richard~J. Chen, Chengkuan Chen, Yicong Li, Tiffany~Y. Chen, Andrew~D. Trister, Rahul~G. Krishnan, and Faisal Mahmood.
\newblock Scaling vision transformers to gigapixel images via hierarchical self-supervised learning.
\newblock In \emph{{CVPR}}, pages 16123--16134. IEEE, 2022.

\bibitem[Cooper et~al.(2022)Cooper, Arandjelovic, and Harrison]{cooper2022hipe}
Jessica Cooper, Ognjen Arandjelovic, and David~J. Harrison.
\newblock Believe the {HiPe}: Hierarchical perturbation for fast, robust, and model-agnostic saliency mapping.
\newblock \emph{Pattern Recognit.}, 129:\penalty0 108743, 2022.

\bibitem[Darcet et~al.(2024)Darcet, Oquab, Mairal, and Bojanowski]{darcet2024registers}
Timothée Darcet, Maxime Oquab, Julien Mairal, and Piotr Bojanowski.
\newblock Vision transformers need registers.
\newblock In \emph{{ICLR}}. OpenReview.net, 2024.

\bibitem[Deng et~al.(2009)Deng, Dong, Socher, Li, Li, and Fei-Fei]{deng2009imagenet}
Jia Deng, Wei Dong, Richard Socher, Li-Jia Li, Kai Li, and Li Fei-Fei.
\newblock {ImageNet}: A large-scale hierarchical image database.
\newblock In \emph{{CVPR}}, pages 248--255. IEEE Computer Society, 2009.

\bibitem[Dosovitskiy et~al.(2021)Dosovitskiy, Beyer, Kolesnikov, Weissenborn, Zhai, Unterthiner, Dehghani, Minderer, Heigold, Gelly, Uszkoreit, and Houlsby]{dosovitskiy2021image}
Alexey Dosovitskiy, Lucas Beyer, Alexander Kolesnikov, Dirk Weissenborn, Xiaohua Zhai, Thomas Unterthiner, Mostafa Dehghani, Matthias Minderer, Georg Heigold, Sylvain Gelly, Jakob Uszkoreit, and Neil Houlsby.
\newblock An image is worth 16x16 words: Transformers for image recognition at scale.
\newblock In \emph{{ICLR}}. OpenReview.net, 2021.

\bibitem[Efron and Tibshirani(1993)]{efron1993introduction}
Bradley Efron and Robert Tibshirani.
\newblock \emph{An Introduction to the Bootstrap}.
\newblock Springer, 1993.

\bibitem[{Ehteshami Bejnordi} et~al.(2017){Ehteshami Bejnordi}, Veta, {van Diest}, {van Ginneken}, Karssemeijer, Litjens, {van der Laak}, et~al.]{bejnordi2017camelyon}
Babak {Ehteshami Bejnordi}, Mitko Veta, Paul~Johannes {van Diest}, Bram {van Ginneken}, Nico Karssemeijer, Geert Litjens, Jeroen A. W.~M. {van der Laak}, et~al.
\newblock Diagnostic assessment of deep learning algorithms for detection of lymph node metastases in women with breast cancer.
\newblock \emph{JAMA}, 318\penalty0 (22):\penalty0 2199--2210, 2017.

\bibitem[Esteva et~al.(2017)Esteva, Kuprel, Novoa, Ko, Swetter, Blau, and Thrun]{esteva2017dermatologist}
Andre Esteva, Brett Kuprel, Roberto~A. Novoa, Justin Ko, Susan~M. Swetter, Helen~M. Blau, and Sebastian Thrun.
\newblock Dermatologist-level classification of skin cancer with deep neural networks.
\newblock \emph{Nature}, 542\penalty0 (7639):\penalty0 115--118, 2017.

\bibitem[Fel et~al.(2021)Fel, Cadène, Chalvidal, Cord, Vigouroux, and Serre]{fel2021sobol}
Thomas Fel, Rémi Cadène, Mathieu Chalvidal, Matthieu Cord, David Vigouroux, and Thomas Serre.
\newblock Look at the variance! efficient black-box explanations with {Sobol}-based sensitivity analysis.
\newblock In \emph{{NeurIPS}}, pages 26005--26014, 2021.

\bibitem[Fel et~al.(2022)Fel, Hervier, Vigouroux, Poché, Plakoo, Cadène, Chalvidal, Colin, Boissin, Béthune, Picard, Nicodeme, Gardes, Flandin, and Serre]{fel2022xplique}
Thomas Fel, Lucas Hervier, David Vigouroux, Antonin Poché, Justin Plakoo, Rémi Cadène, Mathieu Chalvidal, Julien Colin, Thibaut Boissin, Louis Béthune, Agustin~Martin Picard, Claire Nicodeme, Laurent Gardes, Grégory Flandin, and Thomas Serre.
\newblock Xplique: A deep learning explainability toolbox.
\newblock \emph{CoRR}, abs/2206.04394, 2022.

\bibitem[Gao et~al.(2023)Gao, Li, Yang, Cheng, Han, and Torr]{gao2022imagenets}
Shanghua Gao, Zhong-Yu Li, Ming-Hsuan Yang, Ming-Ming Cheng, Junwei Han, and Philip H.~S. Torr.
\newblock Large-scale unsupervised semantic segmentation.
\newblock \emph{IEEE Trans. Pattern Anal. Mach. Intell.}, 45\penalty0 (6):\penalty0 7457--7476, 2023.

\bibitem[Gu et~al.(2026)Gu, Zhang, Liu, Cao, and Zhang]{gu2026phasewin}
Zihan Gu, Junchi Zhang, Li Liu, Xiaochun Cao, and Hua Zhang.
\newblock Phasewin: An efficient search algorithm for faithful visual attribution.
\newblock \emph{CoRR}, abs/2606.18008, 2026.

\bibitem[He et~al.(2016)He, Zhang, Ren, and Sun]{he2016deep}
Kaiming He, Xiangyu Zhang, Shaoqing Ren, and Jian Sun.
\newblock Deep residual learning for image recognition.
\newblock In \emph{{CVPR}}, pages 770--778. IEEE Computer Society, 2016.

\bibitem[Hense et~al.(2024)Hense, Idaji, Eberle, Schnake, Dippel, Ciernik, Buchstab, Mock, Klauschen, and Müller]{hense2024xmil}
Julius Hense, Mina~Jamshidi Idaji, Oliver Eberle, Thomas Schnake, Jonas Dippel, Laure Ciernik, Oliver Buchstab, Andreas Mock, Frederick Klauschen, and Klaus-Robert Müller.
\newblock xmil: Insightful explanations for multiple instance learning in histopathology.
\newblock In \emph{{NeurIPS}}, 2024.

\bibitem[Holm(1979)]{holm1979simple}
Sture Holm.
\newblock A simple sequentially rejective multiple test procedure.
\newblock \emph{Scandinavian Journal of Statistics}, 6\penalty0 (2):\penalty0 65--70, 1979.

\bibitem[Ilse et~al.(2018)Ilse, Tomczak, and Welling]{ilse2018attention}
Maximilian Ilse, Jakub~M. Tomczak, and Max Welling.
\newblock Attention-based deep multiple instance learning.
\newblock In \emph{{ICML}}, pages 2132--2141. PMLR, 2018.

\bibitem[Jumper et~al.(2021)Jumper, Evans, Pritzel, Green, Figurnov, et~al.]{jumper2021alphafold}
John Jumper, Richard Evans, Alexander Pritzel, Tim Green, Michael Figurnov, et~al.
\newblock Highly accurate protein structure prediction with {AlphaFold}.
\newblock \emph{Nature}, 596\penalty0 (7873):\penalty0 583--589, 2021.

\bibitem[Kapishnikov et~al.(2019)Kapishnikov, Bolukbasi, Viégas, and Terry]{kapishnikov2019xrai}
Andrei Kapishnikov, Tolga Bolukbasi, Fernanda~B. Viégas, and Michael Terry.
\newblock Xrai: Better attributions through regions.
\newblock In \emph{{ICCV}}, pages 4947--4956. IEEE, 2019.

\bibitem[Kasmi et~al.(2023)Kasmi, Dubus, Saint~Drenan, and Blanc]{kasmi2023wcam}
Gabriel Kasmi, Laurent Dubus, Yves-Marie Saint~Drenan, and Philippe Blanc.
\newblock Assessment of the reliablity of a model's decision by generalizing attribution to the wavelet domain.
\newblock In \emph{NeurIPS Workshop XAI in Action: Past, Present, and Future Applications}, 2023.

\bibitem[Kasmi et~al.(2025)Kasmi, Brunetto, Fel, and Parekh]{kasmi2025wam}
Gabriel Kasmi, Amandine Brunetto, Thomas Fel, and Jayneel Parekh.
\newblock One wave to explain them all: A unifying perspective on feature attribution.
\newblock In \emph{{ICML}}. PMLR / OpenReview.net, 2025.

\bibitem[Khanam and Hussain(2024)]{khanam2024yolov11}
Rahima Khanam and Muhammad Hussain.
\newblock {YOLOv11}: An overview of the key architectural enhancements.
\newblock \emph{CoRR}, abs/2410.17725, 2024.

\bibitem[Kolek et~al.(2022)Kolek, Nguyen, Levie, Bruna, and Kutyniok]{kolek2022cartoonx}
Stefan Kolek, Duc~Anh Nguyen, Ron Levie, Joan Bruna, and Gitta Kutyniok.
\newblock Cartoon explanations of image classifiers.
\newblock In \emph{{ECCV (12)}}, pages 443--458. Springer, 2022.

\bibitem[Kolek et~al.(2023)Kolek, Windesheim, Andrade-Loarca, Kutyniok, and Levie]{kolek2023shearletx}
Stefan Kolek, Robert Windesheim, Héctor Andrade-Loarca, Gitta Kutyniok, and Ron Levie.
\newblock Explaining image classifiers with multiscale directional image representation.
\newblock In \emph{{CVPR}}, pages 18600--18609. IEEE, 2023.

\bibitem[Li et~al.(2021)Li, Li, and Eliceiri]{li2021dsmil}
Bin Li, Yin Li, and Kevin~W. Eliceiri.
\newblock Dual-stream multiple instance learning network for whole slide image classification with self-supervised contrastive learning.
\newblock In \emph{{CVPR}}, pages 14318--14328. Computer Vision Foundation / IEEE, 2021.

\bibitem[Litjens et~al.(2017)Litjens, Kooi, Bejnordi, Setio, Ciompi, Ghafoorian, van~der Laak, van Ginneken, and S{\'{a}}nchez]{litjens2017survey}
Geert Litjens, Thijs Kooi, Babak~Ehteshami Bejnordi, Arnaud Arindra~Adiyoso Setio, Francesco Ciompi, Mohsen Ghafoorian, Jeroen A. W.~M. van~der Laak, Bram van Ginneken, and Clara~I. S{\'{a}}nchez.
\newblock A survey on deep learning in medical image analysis.
\newblock \emph{Medical Image Anal.}, 42:\penalty0 60--88, 2017.

\bibitem[Loshchilov and Hutter(2017)]{loshchilov2017sgdr}
Ilya Loshchilov and Frank Hutter.
\newblock {SGDR:} stochastic gradient descent with warm restarts.
\newblock In \emph{{ICLR}}, 2017.

\bibitem[Loshchilov and Hutter(2019)]{loshchilov2019decoupled}
Ilya Loshchilov and Frank Hutter.
\newblock Decoupled weight decay regularization.
\newblock In \emph{{ICLR}}, 2019.

\bibitem[Lundberg(2020)]{lundberg2020partition}
Scott Lundberg.
\newblock The {SHAP} partition explainer.
\newblock Documentation of the SHAP library, 2020.

\bibitem[Lundberg and Lee(2017)]{lundberg2017unified}
Scott~M. Lundberg and Su-In Lee.
\newblock A unified approach to interpreting model predictions.
\newblock In \emph{{NIPS}}, pages 4765--4774, 2017.

\bibitem[Mallat(1989)]{mallat1989theory}
St{\'{e}}phane Mallat.
\newblock A theory for multiresolution signal decomposition: The wavelet representation.
\newblock \emph{{IEEE} Trans. Pattern Anal. Mach. Intell.}, 11\penalty0 (7):\penalty0 674--693, 1989.

\bibitem[Novello et~al.(2022)Novello, Fel, and Vigouroux]{novello2022hsic}
Paul Novello, Thomas Fel, and David Vigouroux.
\newblock Making sense of dependence: Efficient black-box explanations using dependence measure.
\newblock In \emph{{NeurIPS}}, 2022.

\bibitem[Petsiuk et~al.(2018)Petsiuk, Das, and Saenko]{petsiuk2018rise}
Vitali Petsiuk, Abir Das, and Kate Saenko.
\newblock Rise: Randomized input sampling for explanation of black-box models.
\newblock In \emph{{BMVC}}, page 151. BMVA Press, 2018.

\bibitem[Pinheiro and Collobert(2015)]{pinheiro2015image}
Pedro H.~O. Pinheiro and Ronan Collobert.
\newblock From image-level to pixel-level labeling with convolutional networks.
\newblock In \emph{{CVPR}}, pages 1713--1721, 2015.

\bibitem[Rao et~al.(2022)Rao, Böhle, and Schiele]{rao2022towards}
Sukrut Rao, Moritz Böhle, and Bernt Schiele.
\newblock Towards better understanding attribution methods.
\newblock In \emph{{CVPR}}, pages 10213--10222. IEEE, 2022.

\bibitem[Rashid et~al.(2026)Rashid, Amparore, Ferrari, and Verda]{rashid2026shapbpt}
Muhammad Rashid, Elvio~G. Amparore, Enrico Ferrari, and Damiano Verda.
\newblock Shapbpt: Image feature attributions using data-aware binary partition trees.
\newblock In \emph{{AAAI}}, pages 25099--25107. AAAI Press, 2026.

\bibitem[Ribeiro et~al.(2016)Ribeiro, Singh, and Guestrin]{ribeiro2016lime}
Marco~Túlio Ribeiro, Sameer Singh, and Carlos Guestrin.
\newblock "why should i trust you?": Explaining the predictions of any classifier.
\newblock In \emph{{KDD}}, pages 1135--1144. ACM, 2016.

\bibitem[Rong et~al.(2022)Rong, Leemann, Borisov, Kasneci, and Kasneci]{rong2022road}
Yao Rong, Tobias Leemann, Vadim Borisov, Gjergji Kasneci, and Enkelejda Kasneci.
\newblock A consistent and efficient evaluation strategy for attribution methods.
\newblock In \emph{{ICML}}, pages 18770--18795. PMLR, 2022.

\bibitem[Selvaraju et~al.(2017)Selvaraju, Cogswell, Das, Vedantam, Parikh, and Batra]{selvaraju2017gradcam}
Ramprasaath~R. Selvaraju, Michael Cogswell, Abhishek Das, Ramakrishna Vedantam, Devi Parikh, and Dhruv Batra.
\newblock Grad-cam: Visual explanations from deep networks via gradient-based localization.
\newblock In \emph{{ICCV}}, pages 618--626. IEEE Computer Society, 2017.

\bibitem[Simonyan and Zisserman(2015)]{simonyan2015very}
Karen Simonyan and Andrew Zisserman.
\newblock Very deep convolutional networks for large-scale image recognition.
\newblock In \emph{{ICLR}}, 2015.

\bibitem[Simonyan et~al.(2014)Simonyan, Vedaldi, and Zisserman]{simonyan2014deep}
Karen Simonyan, Andrea Vedaldi, and Andrew Zisserman.
\newblock Deep inside convolutional networks: Visualising image classification models and saliency maps.
\newblock In \emph{{ICLR (Workshop Poster)}}, 2014.

\bibitem[Smilkov et~al.(2017)Smilkov, Thorat, Kim, Viégas, and Wattenberg]{smilkov2017smoothgrad}
Daniel Smilkov, Nikhil Thorat, Been Kim, Fernanda~B. Viégas, and Martin Wattenberg.
\newblock Smoothgrad: removing noise by adding noise.
\newblock \emph{CoRR}, abs/1706.03825, 2017.

\bibitem[Smith and Topin(2017)]{smith2017superconvergence}
Leslie~N. Smith and Nicholay Topin.
\newblock Super-convergence: Very fast training of residual networks using large learning rates.
\newblock \emph{CoRR}, abs/1708.07120, 2017.

\bibitem[Sumiyasu et~al.(2024)Sumiyasu, Kawamoto, and Kera]{sumiyasu2024moxi}
Kosuke Sumiyasu, Kazuhiko Kawamoto, and Hiroshi Kera.
\newblock Identifying important group of pixels using interactions.
\newblock In \emph{{CVPR}}, pages 6017--6026. IEEE, 2024.

\bibitem[Sundararajan et~al.(2017)Sundararajan, Taly, and Yan]{sundararajan2017axiomatic}
Mukund Sundararajan, Ankur Taly, and Qiqi Yan.
\newblock Axiomatic attribution for deep networks.
\newblock In \emph{{ICML}}, pages 3319--3328. PMLR, 2017.

\bibitem[Teneggi et~al.(2023)Teneggi, Luster, and Sulam]{teneggi2022hshap}
Jacopo Teneggi, Alexandre Luster, and Jeremias Sulam.
\newblock Fast hierarchical games for image explanations.
\newblock \emph{IEEE Trans. Pattern Anal. Mach. Intell.}, 45\penalty0 (4):\penalty0 4494--4503, 2023.

\bibitem[Tibshirani(1996)]{tibshirani1996regression}
Robert Tibshirani.
\newblock Regression shrinkage and selection via the lasso.
\newblock \emph{Journal of the Royal Statistical Society: Series B (Methodological)}, 58\penalty0 (1):\penalty0 267--288, 1996.

\bibitem[Touvron et~al.(2021)Touvron, Cord, Douze, Massa, Sablayrolles, and Jégou]{touvron2021training}
Hugo Touvron, Matthieu Cord, Matthijs Douze, Francisco Massa, Alexandre Sablayrolles, and Hervé Jégou.
\newblock Training data-efficient image transformers \& distillation through attention.
\newblock In \emph{{ICML}}, pages 10347--10357. PMLR, 2021.

\bibitem[Wang et~al.(2024)Wang, Zhu, Gao, Gan, Zhang, Gu, Qian, Chen, and Ma]{wang2024realiad}
Chengjie Wang, Wenbing Zhu, Bin-Bin Gao, Zhenye Gan, Jiangning Zhang, Zhihao Gu, Shuguang Qian, Mingang Chen, and Lizhuang Ma.
\newblock Real-iad: A real-world multi-view dataset for benchmarking versatile industrial anomaly detection.
\newblock In \emph{{CVPR}}, pages 22883--22892. IEEE, 2024.

\bibitem[Wang et~al.(2020)Wang, Wang, Du, Yang, Zhang, Ding, Mardziel, and Hu]{wang2020scorecam}
Haofan Wang, Zifan Wang, Mengnan Du, Fan Yang, Zijian Zhang, Sirui Ding, Piotr Mardziel, and Xia Hu.
\newblock Score-cam: Score-weighted visual explanations for convolutional neural networks.
\newblock In \emph{{CVPR Workshops}}, pages 111--119. Computer Vision Foundation / IEEE, 2020.

\bibitem[Xia et~al.(2018)Xia, Bai, Ding, Zhu, Belongie, Luo, Datcu, Pelillo, and Zhang]{xia2018dota}
Gui{-}Song Xia, Xiang Bai, Jian Ding, Zhen Zhu, Serge~J. Belongie, Jiebo Luo, Mihai Datcu, Marcello Pelillo, and Liangpei Zhang.
\newblock {DOTA:} {A} large-scale dataset for object detection in aerial images.
\newblock In \emph{{CVPR}}, pages 3974--3983. Computer Vision Foundation / IEEE Computer Society, 2018.

\bibitem[Zeiler and Fergus(2014)]{zeiler2014visualizing}
Matthew~D. Zeiler and Rob Fergus.
\newblock Visualizing and understanding convolutional networks.
\newblock In \emph{{ECCV (1)}}, pages 818--833. Springer, 2014.

\bibitem[Zhang et~al.(2016)Zhang, Lin, Brandt, Shen, and Sclaroff]{zhang2016excitation}
Jianming Zhang, Zhe Lin, Jonathan Brandt, Xiaohui Shen, and Stan Sclaroff.
\newblock Top-down neural attention by excitation backprop.
\newblock In \emph{{ECCV (4)}}, pages 543--559. Springer, 2016.

\bibitem[Zou et~al.(2022)Zou, Jeong, Pemula, Zhang, and Dabeer]{zou2022spot}
Yang Zou, Jongheon Jeong, Latha Pemula, Dongqing Zhang, and Onkar Dabeer.
\newblock Spot-the-difference self-supervised pre-training for anomaly detection and segmentation.
\newblock In \emph{{ECCV (30)}}, pages 392--408. Springer, 2022.

\end{thebibliography}
